\documentclass[review]{elsarticle}
\usepackage{lineno,hyperref}

\journal{Pattern Recognition}

\usepackage{dcolumn}
\usepackage{bm}
\usepackage{color}
\usepackage{tabularx}
\usepackage{amsmath}
\usepackage{graphicx}
\usepackage{epstopdf}
\usepackage{epsfig}
\usepackage{makecell}
\usepackage{multirow}
\usepackage{caption}
\usepackage{float}
\usepackage{subfigure}
\usepackage{cleveref}
\usepackage{booktabs}
\usepackage{amsfonts}
\usepackage{mathrsfs}
\usepackage{threeparttable}
\usepackage{upgreek}
\biboptions{numbers,sort&compress}
\usepackage{setspace}
\usepackage{chngcntr}
\usepackage{amsthm}
\usepackage{url}
\newtheorem{theorem}{Theorem}[section]

\newtheorem{remark}{Remark}[section]

\def\tr{\mbox{tr}}

\usepackage{algorithm,algorithmic}

\makeatletter
\newenvironment{breakablealgorithm}
{
	\begin{center}
		\refstepcounter{algorithm}
		\hrule height.8pt depth0pt \kern2pt
		\renewcommand{\caption}[2][\relax]{
			{\raggedright\textbf{\ALG@name~\thealgorithm} ##2\par}%
			\ifx\relax##1\relax 
			\addcontentsline{loa}{algorithm}{\protect\numberline{\thealgorithm}##2}%
			\else 
			\addcontentsline{loa}{algorithm}{\protect\numberline{\thealgorithm}##1}%
			\fi
			\kern2pt\hrule\kern2pt
		}
	}{
		\kern2pt\hrule\relax
	\end{center}
}
\begin{document}
\captionsetup[figure]{labelfont={bf},labelformat={default},labelsep=period,name={Fig.}} 
\begin{frontmatter}

\title{A Restarted Large-Scale Spectral Clustering with Self-Guiding and Block Diagonal Representation}


\author[mymainaddress]{Yongyan Guo}
\ead{guoyy@cumt.edu.cn}

\author[mymainaddress]{Gang Wu\corref{mycorrespondingauthor}}
\cortext[mycorrespondingauthor]{Corresponding author.}
\ead{gangwu@cumt.edu.cn}

\address[mymainaddress]{School of Mathematics, China University of Mining and Technology, Xuzhou 221116, P.R. China.}

\begin{abstract}
Spectral clustering is one of the most popular unsupervised machine learning methods. Constructing similarity matrix is crucial to this type of method. In most existing works, the similarity matrix is computed {\it once for all} or is updated alternatively. However, the former is difficult to reflect comprehensive relationships among data points, and the latter is time-consuming and is even infeasible for large-scale problems. In this work, we propose a restarted clustering framework with self-guiding and block diagonal representation. An advantage of the framework is that some useful clustering information obtained from previous cycles could
be preserved as much as possible. To the best of our knowledge, this is the first work that applies this strategy to spectral
clustering. The key difference is that we reclassify the samples in each cycle of our method, while they are classified only once in existing methods.
 To further release the overhead, we introduce a block diagonal representation with Nystr\"{o}m approximation for constructing the similarity matrix. Theoretical results are established to show the rationality of inexact computations in spectral clustering.
Comprehensive experiments are performed on some benchmark databases, which show the superiority of our proposed algorithms over many state-of-the-art algorithms for large-scale problems. Specifically, our framework has a potential boost for clustering algorithms and works well even using an initial guess chosen randomly.
\end{abstract}

\begin{keyword}
Spectral clustering, Restarting, Self-guiding, Block diagonal representation, Kernel trick, Nystr\"{o}m approximation.
\end{keyword}

\end{frontmatter}


\section{Introduction}\label{sec1}

Clustering is an important task in many applications such as pattern recognition \cite{20PR}, machine learning \cite{19LRKC}, community detection \cite{18CCCD}, gene sequence analysis \cite{20GSA}, and so on. During past decades, a large number of clustering methods have been proposed, such as $K$-means clustering \cite{79ACKA}, subspace clustering \cite{19BDRSC}, kernel-based clustering \cite{22MKKM}, multi-view clustering \cite{22UMVSC}, spectral clustering \cite{22FINC,22SGEC}, and so on.

Of particular interest, spectral clustering is one of the most used clustering methods, which is based on spectral graph theory and can dexterously capture the underlying data structure information \cite{07TSC}. Spectral clustering method regards similarity between samples as an undirected graph, and the samples are clustered into $c$ classes through the graph cut point of view. There are two most commonly used cut criterion, called Ratio cut (Rcut) \cite{92RCPC} and Normalized cut (Ncut) \cite{02SCAA,00NCIS}, respectively. Both of them correspond to optimization problems on trace minimization. In general, spectral clustering method is composed of three stages \cite{07TSC}: learning pairwise similarity, computing eigenvalue problem, and discretizing the spectral embedding into a clustering indicator matrix.

A good similarity matrix can contribute to spectral clustering. Inspired by this observation, the adaptive neighbors graph learns a graph by assigning adaptive and optimal neighbors for each data point \cite{14CAN}. In \cite{22NAKNN}, Cai {\it et al.} built an adaptive $k$-nearest neighbors similarity graph by using the density information. Moreover, a new sparse and subspace-preserving coefficient matrix was constructed by seeking good neighbors \cite{20FGNSC}. In \cite{22EBDR}, Qin {\it et al.} presented an enforced block diagonal subspace clustering with closed-form solution. Due to the fact that exact prior information is helpful to improve clustering performance, a self-supervised spectral clustering was proposed in \cite{22SESC}. Including some classical spectral clustering methods such as Ncut, Rcut, the similarity matrix is computed {\it once for all} in the above mentioned methods, and we call them {\it non-updating methods} in this paper. However, the construction of similarity matrix and the eigenvalue problem involved may be time-consuming for large-scale problem. To alleviate the burden of constructing the similarity matrix, bipartite graph based methods were introduced for handling large-scale databases \cite{20USPEC,22DNSC}. In this method, the similarity with respect to anchor graph is calculated between the original data and a small number of representative points \cite{21FOER,22FGCSS,21FMVPG}. To accelerate the eigendecomposition process, Fowlkes {\it et al.} \cite{04NYST} introduced the Nystr\"{o}m method to approximate the spectral embedding, and some approximate clustering methods were proposed \cite{23FSAB,21CoALa}.

Recently, a series of methods have been proposed by integrating the spectral embedding or clustering indicator matrix into a unified framework \cite{20SWAN,22SGEC,22SCGLD,19MVCMK,22UMVSC}. In these methods, the similarity matrix is {\it updated alternatively}, by using some valid information from spectral embedding, clustering indicator matrix, and other relevant factors, and we call these methods {\it updating methods} in this paper. For instance, based on the idea that block diagonal property may achieve perfect clustering by using spectral clustering in an ideal case, Lu {\it et al.} \cite{19BDRSC} proposed a subspace clustering via block diagonal representation (BDR) to pursue a block diagonal matrix directly. Bai {\it et al.} \cite{22SCSC} proposed a self-constrained spectral clustering by learning pairwise and label self-constrained terms simultaneously. Zhong {\it et al.} \cite{23STSC} proposed a self-taught multi-view spectral clustering based on convex combination and centroid graph fusion schemes. By explicitly pursuing the block diagonality of the similarity matrix, Lin {\it et al.} \cite{22CSBDR} investigated an adaptive BDR method in which the rows and columns of the similarity matrix are fused to ensure block-diagonal pattern. Moreover, some self-expressiveness-based subspace clustering
methods were studied recently by learning the representation matrix with soft block-diagonal regularizer \cite{22BDLSR,21LRMKSC,23PBDR}.

However, in these methods, updating similarity matrix and the corresponding matrix computation problems are very time-consuming in practical calculations, especially for large-scale databases. In order to overcome these difficulties, we propose a restarted spectral clustering framework via self-guiding and block diagonal representation. The main contributions of our work are as follows:
\begin{itemize}
\item{We propose a new restarted clustering framework with self-guiding, and apply it to two classical spectral clustering methods including $K$-means and spectral rotation. The idea stems from keeping the last clustering information to improve the clustering performance gradually during cycles. Our framework has a potential boost for clustering algorithms, and works well even using a initial guess chosen randomly.}

\item{To release the computation overhead and storage requirement for large-scale problems, we introduce a block diagonal representation for constructing the similarity matrix. More precisely, we make use of a block-diagonal similarity kernel matrix $A$, and the between cluster affinities are all zeros in the proposed method. Nystr\"{o}m method are utilized to approximate the kernel matrix instead of forming and storing it explicitly. Theoretical results are established to show the rationality of using low-rank approximations in spectral clustering.}

\item{A common disadvantage of the spectral embedding with spectral rotation method is the heavy workload for computing the clustering indicator matrix. Based on the correlation between vectors, we propose a new scheme for updating the clustering indicator matrix efficiently during cycles. }
\end{itemize}

This paper is organized as follows. In Section \ref{sec2}, we briefly introduce some preliminaries including the Ncut model and spectral rotation for spectral clustering. We elaborate on the idea of this work in Section \ref{sec3}. In Section \ref{sec4}, we propose two restarted spectral clustering algorithms via self-guiding mechanism. Comprehensive numerical experiments are performed in Section \ref{sec5}, to show the superiority of the proposed methods over many state-of-the-art methods for large-scale clustering. Concluding remarks are given in Section \ref{sec6}.

\section{Preliminaries}\label{sec2}


The intuitive goal of clustering is to partition the objects into several groups such that objects in the same group are similar to each other, while those in different groups are dissimilar.
The problem of clustering can also be approached from a graph theory point of view. Let $X=[\boldsymbol{x}_1,\ldots,\boldsymbol{x}_n]^{T}\in \mathbb{R}^{n \times d}$ be the data set, with $n$ and $d$ being the number of samples and features respectively. Consider an undirected similarity graph $\mathbb{G}=(\mathbb{V},\mathbb{E})$ with vertex set $\mathbb{V}=(\boldsymbol{v}_1,\ldots,\boldsymbol{v}_n)$, where every vertex $\boldsymbol{v}_i$ represents the object $\boldsymbol{x}_i$, and the edge between vertices $\boldsymbol{v}_i$ and $\boldsymbol{v}_j$ is weighted by a similarity $A=[{a}_{ij}]\in \mathbb{R}^{n\times n}$ \cite{07TSC}.

In this work, we pay special attention to the famous Ncut for graph clustering, whose model is expressed as \cite{00NCIS,02SCAA,07TSC}
\begin{equation}\label{2.1.2}
\makecell[c]{
Ncut(c)=\sum_{j=1}^c\frac{cut(\mathbb{C}_j,\bar{\mathbb{C}}_j)}{vol(\mathbb{C}_j)},\\
}
\end{equation}
where $\mathbb{C}_j~(j=1,2,\ldots,c)$ denotes the set of vertices in the $j$-th class, and $\bar{\mathbb{C}}_j$, $vol(\mathbb{C}_j)$ represent the complement part of $\mathbb{C}_j$ and the sum of degree of the $j$-th class, respectively.
Indeed, the objective function of \eqref{2.1.2} can be formulated as the following minimization problem \cite{07TSC},
\begin{equation}\label{2.1.4}
\begin{split}
\underset{M=D^{\frac{1}{2}}Y(Y^{T}DY)^{-\frac{1}{2}}}{\min}& \ \tr(M^{T}D^{-\frac{1}{2}}(D-A)D^{-\frac{1}{2}}M),\\
& s.t.\ Y \in \{0,1\}^{n \times c}, \boldsymbol{y}^{i}\mathbf{1}=1,
\end{split}
\end{equation}
where $D$ is the diagonal matrix whose diagonal elements are the sum of the rows of the similarity matrix $A$, and
\begin{equation}\label{3}
 Y=[\boldsymbol{y}^{1},\boldsymbol{y}^{2},\ldots,\boldsymbol{y}^{n}]^{T} \in \mathbb{R}^{n \times c}
 \end{equation}
 is a binary indicator matrix in which one and only one element is equal to $1$ in each row. More precisely, the $j$-th entry of $\boldsymbol{y}^{i}$ is 1 if the sample $\boldsymbol{x}_i$ is assigned to the $j$-th cluster, and 0 otherwise.

Unfortunately, the above optimization problem is NP-hard \cite{07TSC}. The spectral clustering provides a computationally tractable solution to the Ncut problem.
The computation of this problem consists of two steps. First, the most commonly used strategy for addressing this issue is to relax the embedding matrix $M$ from the discrete values to continuous ones \cite{00NCIS,02SCAA,07TSC}
\begin{equation}\label{2.1.6}
 \underset{\substack{M \in \mathbb{R}^{n \times c}\\M^{T}M=I}}{\min} \ \tr\big(M^{T}(I-D^{-\frac{1}{2}}AD^{-\frac{1}{2}})M\big),
\end{equation}
where the columns of $M=[{\bm m}_1,{\bm m}_2,\ldots,{\bm m}_c]$ are composed of the eigenvectors corresponding to the smallest $c$ eigenvalues of $I-D^{-\frac{1}{2}}AD^{-\frac{1}{2}}$ \cite{07TSC}.
It is worth noting that $I-D^{-\frac{1}{2}}AD^{-\frac{1}{2}}$ is called the \emph{Normalized Laplacian matrix}, and if $D=I$, then Ncut reduces to Rcut \cite{07TSC}.

Second, to obtain the indicator matrix $Y$ for the final clustering results, one needs to perform additional processing on the spectral embedding matrix $M$ in spectral clustering. There are two commonly-used post-processing techniques, namely, $K$-means \cite{15kmeans} and spectral rotation \cite{13SRSC}.

Given the spectral embedding matrix $M$, the mathematical model of $K$-means can be described as follows \cite{15kmeans}
\begin{align}\label{2.1.8}
 \underset{Y}{\min} \ \sum_{i=1}^{n}\sum_{j=1}^{c}y_{ij}\|\bm{m}_i-\bm{\mu}_j\|_{2}^{2}\nonumber\\
 s.t.\ Y \in \{0,1\}^{n \times c}, \bm{y}^{i}\mathbf{1}=1,
\end{align}
where $\bm{m}_i$ and $\bm{\mu}_j$ are the $i$-th column and $j$-th cluster centroid of the spectral embedding matrix $M$, respectively. Indeed, \eqref{2.1.8} can be rewritten into the following matrix factorization form \cite{15kmeans}
\begin{align}\label{2.1.9}
 \underset{Y}{\max} \ \tr((Y^{T}Y)^{-\frac{1}{2}}Y^{T}MM^{T}Y(Y^{T}Y)^{-\frac{1}{2}})\nonumber\\
 s.t. \ Y \in \{0,1\}^{n \times c}, \bm{y}^{i}\mathbf{1}=1.
\end{align}

However, the solution of the optimization problem \eqref{2.1.6} is not unique. Indeed, for any orthonormal matrix $Q\in\mathbb{R}^{c\times c}$, $MQ$ is also a solution to \eqref{2.1.6}. The goal of spectral rotation is to make $MQ$ be closer to the matrix $D^{\frac{1}{2}}Y(Y^{T}DY)^{-\frac{1}{2}}$ by finding a proper orthogonal matrix $Q$.
To this end, the mathematical model of the spectral rotation is \cite{13SRSC}
\begin{align}\label{2.1.10}
 \underset{Y,Q}{\min} \ \|MQ-D^{\frac{1}{2}}Y(Y^{T}DY)^{-\frac{1}{2}}\|_{F}^{2}\nonumber\\
 s.t. \ Y \in \{0,1\}^{n \times c}, \boldsymbol{y}^{i}\mathbf{1}=1, Q^{T}Q=I.
\end{align}
For more details, refer to \cite{22MKKM,21FOER,22FGCSS}.


\section{Motivation: Restarting with self-guiding and Block-diagonal representation}\label{sec3}

Despite the progress on spectral clustering, two thorny issues remain unresolved. On one hand, a fixed similarity matrix is difficult to reflect comprehensive relationships among data points and may not be optimal for learning sample embedding. On the other hand, the existing methods for updating similarity matrix alternatively
are time-consuming and even infeasible for large-scale problems. To deal with these problems, we propose a framework for {\it restarting} the spectral clustering methods, and
 the key ideas are two-folds: {\it Restarting with self-guiding} and {\it Block-diagonal representation}. Let us discuss them in detail.

$\bullet$ {\bf Restarting with self-guiding}: In this work, we propose a restarted framework for large-scale spectral clustering. More precisely, we {\it reclassify} the samples to construct a new similarity matrix according to the clustering information just obtained, and run the spectral clustering method from scratch. We call this strategy {\it restarting}, and refer to this process as a {\it cycle}. An advantage of this strategy is that some useful clustering information from previous cycles could be preserved as much as possible. To the best of our knowledge, this is the first work that applies the restarting strategy to spectral clustering.

\begin{figure}[htbp]
\centering
\includegraphics[width=1\linewidth]{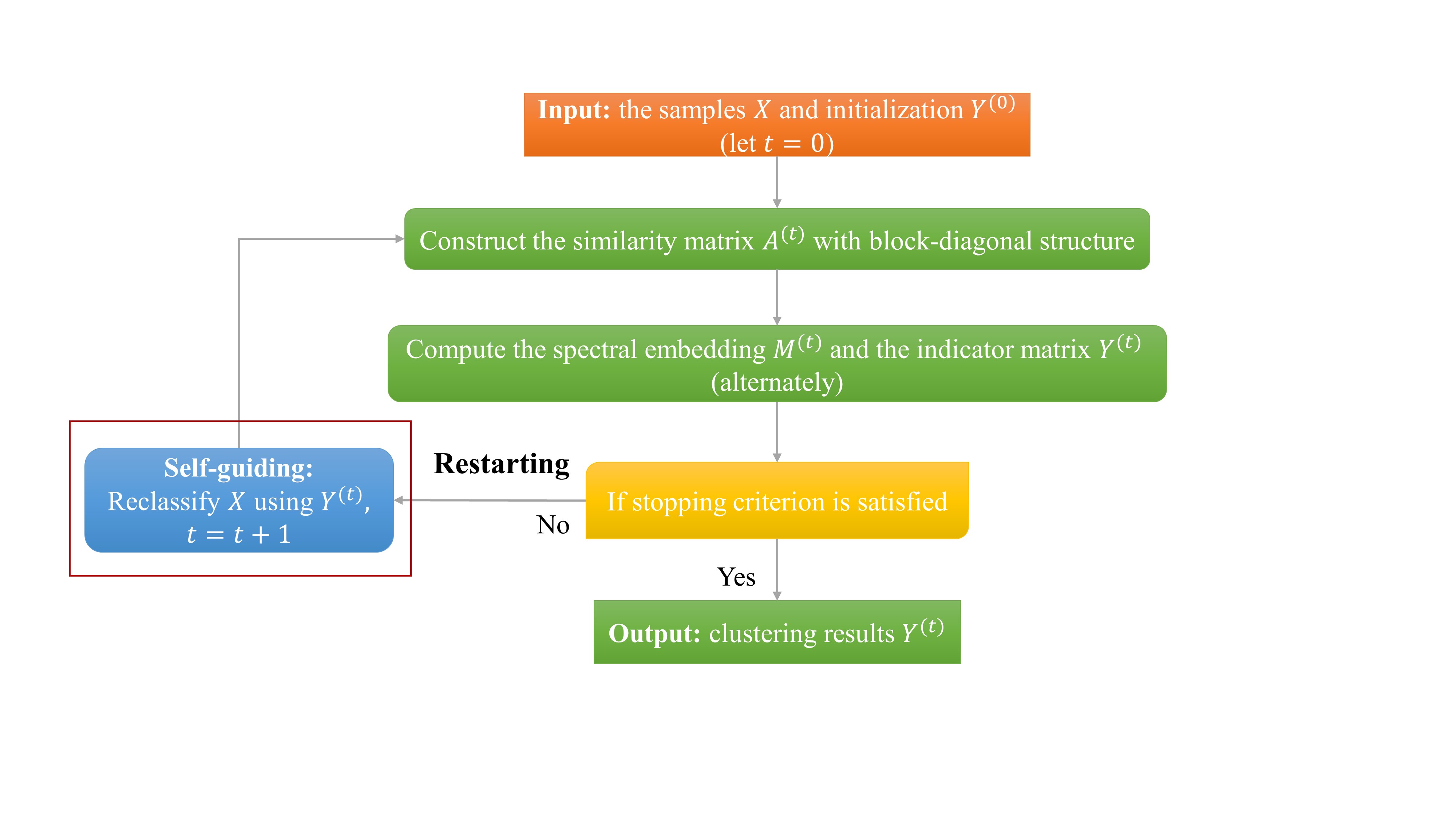}
\caption{Framework of the large-scale spectral clustering with restarting and block-diagonal representation.
}
\label{view}
\end{figure}

The motivation of our approach is that we exploit the clustering information from the current cycle as a {\it self-guide} for classifying in the next cycle.
Notice that the similarity matrices are recalculated in both the {\it restarting} method and the {\it updating} method.
The key difference is that we reclassify the samples in each cycle of the {\it restarting} method, while the samples are classified {\it only once} in the {\it updating} method. One refers to Fig. \ref{view} for the framework of the proposed methods. Notice that the initial guess $Y^{(0)}$ can be from {\it any} clustering methods.

$\bullet$ {\bf Block-diagonal representation:} The following theorem shows that if we express the similarity matrix as a block diagonal one, the graph has at least $c$ connected components corresponding to $c$ categories.
\begin{theorem}\cite[Proposition 4]{07TSC}\label{th1}
The multiplicity $c$ of the eigenvalue 0 of the graph Laplacian $L$ is equal to the number of connected components in the graph $A$.
\end{theorem}
For large-scale problems, the overhead of computing similarity matrix in each cycle will be prohibitive or even infeasible. Inspired by \cite{19BDRSC} and Theorem \ref{th1}, we suggest using the block-diagonalization scheme. More precisely, we make use of a block-diagonal similarity matrix $A$ {\it directly}, and the between-cluster affinities are all zeros, which introduces a block diagonal Laplacian matrix correspondingly. Therefore, one only needs to perform eigendecomposition on block matrices with much smaller size. This scheme can reduce the workload for eigenvector computations significantly, and is suitable to parallel computing.

\section{Restarted Spectral Clustering Methods with Self-guiding}\label{sec4}

In this section, we propose two restarted spectral clustering methods with self-guiding,
one is based on the classical spectral clustering method, and the other is based on spectral embedding with spectral rotation.

\subsection{Approximate similarity matrix by using kernel trick and block diagonal representation}\label{Sec4.1}

The construction and computation to similarity matrix are key problems to spectral clustering. In the proposed framework, we have to update the similarity matrix in each cycle, and the computational overhead may be heavy and the proposed methods can be infeasible for large-scale data problems. In order to release the overhead, we assume that the similarities between the samples in different clusters are zero. Under this {\it ideal} case, the similarity matrix constructed will share block diagonal structure. Combining kernel tricks \cite{Kernel} and block diagonal representation, we propose to approximate the similarity matrix $A$ in the following way:
Suppose that the data set $X$ can be divided into $c$ subsets $X=[X_1;X_2;\ldots;X_c]$, where $X_j$ are composed of the (normalized) samples in the $j$-th class. Let $X_j=[\bm{x}_{j_{1}},\bm{x}_{j_{2}},\ldots,\bm{x}_{j_{n_j}}]^{T} \in \mathbb{R}^{n_j \times d}$, where $n_j$ is the number of the samples in the $j$-th class.
In this paper, we take advantage of the popular used Radial Basis Function (RBF) (or
Gaussian kernel function) \cite{Kernel}, and compute the $(p,q)$-th element in the kernel similarity matrix with respect to the $j$-th block $A_j\in \mathbb{R}^{n_j \times n_j}$ in the following way
\begin{equation}\label{3.1.2}
A_j(p,q)=\exp\left\{-\frac{\|\boldsymbol{x}_{j_p}-\boldsymbol{x}_{j_q}\|^{2}}{2\tau^2}\right\},~j=1,2,\ldots,c,
\end{equation}
where $\tau>0$ is the bandwidth of Gaussian kernel, $p,q=1,2,\ldots,n_j$, and $\sum_{j=1}^{c}n_j=n$.

We approximate $A$ by using the following similarity matrix

\begin{equation}\label{3.1.3}
\widetilde{A}=diag(A_1,A_2,\ldots,A_c) \in \mathbb{R}^{n \times n},
\end{equation}
which is a block diagonal and symmetric positive semi-definite matrix. Define the diagonal matrix
\begin{equation}\label{3.1.4}
\widetilde{D}=diag(D_1,D_2,\ldots,D_c) \in \mathbb{R}^{n \times n},	
\end{equation}
where $D_j\in \mathbb{R}^{n_j \times n_j}$ is the degree matrix of the $j$-th block $A_j,~j=1,2,\ldots,c$.
{Thus, we approximate $D^{-\frac{1}{2}}AD^{-\frac{1}{2}}$ by using
\begin{equation}\label{3.1.5}
\begin{split}
\widetilde{L}&=\widetilde{D}^{-\frac{1}{2}}\widetilde{A}\widetilde{D}^{-\frac{1}{2}}= diag(L_1,L_2,\ldots,L_c) \in \mathbb{R}^{n \times n},
\end{split}
\end{equation}
where $L_j=D_j^{-\frac{1}{2}}A_jD_j^{-\frac{1}{2}}\in \mathbb{R}^{n_j \times n_j}$, $j=1,2,\ldots,c$, and $I-\widetilde{D}^{-\frac{1}{2}}\widetilde{A}\widetilde{D}^{-\frac{1}{2}}$ is an approximation to the Laplacian matrix}.

However, if the number of the samples in the $j$-th class $n_j$
is large, the computational cost and storage requirement for the kernel matrix $A_j$ will be prohibitive. A commonly used technique is to compute some low-rank approximations to the kernel matrix, rather than form it directly. As $A_j$ is a symmetric positive semi-definite matrix, we make use of the famous Nystr\"{o}m method \cite{04NYST} to approximate it.

Consider the Nystr\"{o}m approximation
\begin{equation}\label{3.1.8}
A_j\approx T_j=C_jW_j^{\dagger}C_j^{T},
\end{equation}
where $C_j \in \mathbb{R}^{n_j \times m_j}$ is composed of $m_j(\ll n_j)$ columns sampled randomly from $A_j$, and $W_j$ is an $m_j \times m_j$ matrix consisting of the intersection of these $m_j$ columns with the corresponding $m_j$ rows of $C_j$.
Let $C_j=Q_jR_j$ be the economized QR factorization, then
$
T_j=Q_j(R_jW_j^{\dagger}R_j^{T})Q_j^{T}.
$
Let $r_j(\leq m_j)$ be a user-prescribed parameter, $R_jW_j^{\dagger}R_j^{T}=P_j\Lambda_jP_j^{T}$ be the SVD of $R_jW_j^{\dagger}R_j^{T}$, $U_j=Q_jP_j(:,1:r_j)\in \mathbb{R}^{n_j \times r_j}$, $\Sigma_j=\Lambda_j(1:r_j,1:r_j)\in \mathbb{R}^{r_j \times r_j}$, where $P_j(:,1:r_j)$ denotes the matrix composed of the first $r_j$ columns of $P_j$, and $\Lambda_j(1:r_j,1:r_j)$ stands for the matrix formed by using the first $r_j$ rows and $r_j$ columns of $\Lambda_j$.
Then
\begin{align}
T_j^{[r_j]}&= Q_jP_j(:,1:r_j)\Lambda_j(1:r_j,1:r_j)P_j(:,1:r_j)^{T}Q_j^{T}= U_j\Sigma_jU_j^{T} \label{3.1.9}
\end{align}
is an optimal rank $r_j(\ll n_j)$ approximation to $T_j$ \cite{90MPT}, which can be used as a low-rank approximation to $A_j,~j=1,2,\ldots,c$, and $\widehat{A}=diag(T_1^{[r_1]},T_2^{[r_2]},\ldots,T_c^{[r_c]})$ is an approximation to $\widetilde{A}$.

\begin{remark}
Different from the ``soft" block diagonal regularizer \cite{19BDRSC}, our self-guiding mechanism enforces $\widetilde{A}$ to be a {\tt (strictly)} block-diagonal structure with just $c$ blocks. It only requires to approximate $A_j$ by using some low-rank matrices $T_j^{[r_j]},j=1,2,\ldots,c$, instead of updating the $n$-by-$n$ similarity matrix $A$ in each cycle.
\end{remark}

\subsection{A restarted spectral clustering algorithm based on $K$-means}\label{subsec4.2}
In this subsection, we propose a restarted algorithm based on the classical spectral clustering algorithm. In each cycle of this algorithm, we apply the $K$-means algorithm on $M^{(t)}$ to get the clustering indicator matrix $Y^{(t+1)}$. Some theoretical analysis is given.


\subsubsection{The proposed algorithm}\label{subsec3.2.1}

To make use of the block diagonal structure of the matrix $\widehat{A}$ sufficiently, an initial classification label $Y^{(0)}$ is needed in our restarted algorithm.
We can do this by running some clustering algorithms such as $K$-means, or separate the samples randomly into $c$ classes. Notice that the main overhead in the classical spectral clustering algorithm is to solve $c$ smallest eigenpairs of the Laplacian matrix of size $n$-by-$n$.
As a comparison, we only need to calculate the largest eigenpair of the $n_j$-by-$n_j$ matrix $L_j,~j=1,2,\ldots,c$. So we can reduce the computational overhead of the conventional methods significantly.

Let us discuss it in more detail. We consider how to approximate the $\{L_j\}$'s in practice, rather than form them explicitly. With $T_j^{[r_j]}$ at hand, we first introduce the diagonal matrix $\widehat{D}_j$, whose $i$-th diagonal element\footnote{\it Unfortunately, there is no guarantee that the diagonal matrix
$\widehat{D}_j$ is positive and invertible theoretically. To deal with this problem, we take absolute value to the diagonal elements, and set them to 1 if they are smaller than a tolerance, e.g., $10^{-12}$.} is the sum of the $i$-th row of $T_j^{[r_j]},~i=1,2,\ldots,n_j$.
In detail,
\begin{equation}\label{3.2.3.0}
\widehat{D}=diag(\widehat{D}_1,\widehat{D}_2,\ldots,\widehat{D}_c).
\end{equation}
Consider
\begin{align}
\widehat{L}_j\!=\!\widehat{D}_j^{-\frac{1}{2}}T_j^{[r_j]}\widehat{D}_j^{-\frac{1}{2}}\!=\!\widehat{D}_j^{-\frac{1}{2}}U_j\Sigma_jU_j^{T}\widehat{D}_j^{-\frac{1}{2}}\!=\!\widehat{U}_j\Sigma_j\widehat{U}_j^{T}\in \mathbb{R}^{n_j\times n_j},\label{3.2.3}
\end{align}
where $\widehat{U}_j=\widehat{D}_j^{-\frac{1}{2}}U_j \in \mathbb{R}^{n_j\times r_j}$, $j=1,2,\ldots,c$, and
\begin{equation}\label{3.2.3.1}
\widehat{L}=diag(\widehat{L}_1,\widehat{L}_2,\ldots,\widehat{L}_c)\in \mathbb{R}^{n\times n}
\end{equation}
can be taken as an approximation to $\widetilde{L}$, and the eigenpairs of the former can be utilized as some approximations to those of the latter.

We are ready to compute the largest eigenpair of $\widehat{L}_j$, which is an approximation to that of $L_j$, $j=1,2,\ldots,c$.
Recall that $\widehat{U}_j$ is not an orthogonal matrix any more, let $\widehat{U}_j=\widehat{Q}_j\widehat{R}_j$ be the economized QR factorization, with $\widehat{R}_j \in \mathbb{R}^{r_j\times r_j}$ being an upper-triangular matrix. Thus,
\begin{equation}\label{3.2.4}
\widehat{L}_j=\widehat{Q}_j\big(\widehat{R}_j\Sigma_j\widehat{R}_j^{T}\big)\widehat{Q}_j^{T}=\!\widehat{Q}_j\Delta_j\widehat{Q}_j^{T},
\end{equation}
where $\Delta_j=\widehat{R}_j\Sigma_j\widehat{R}_j^{T} \in \mathbb{R}^{r_j\times r_j}$. If $(\hat{\lambda}_j,\bm{p}_j)$ is a largest eigenpair of $\Delta_j$, then it is seen from \eqref{3.2.4} that $(\hat{\lambda}_j,\widehat{Q}_j\bm{p}_j)$ is a largest eigenpair of $\widehat{L}_j$, $j=1,2,\ldots,c$.
Denote by $\bm{m}_j=\widehat{Q}_j\bm{p}_j$, and by
$
M=diag(\bm{m}_1,\bm{m}_2,\ldots,\bm{m}_c),~\widehat{\Lambda}=diag(\hat{\lambda}_1,\hat{\lambda}_2,\ldots,\hat{\lambda}_c),
$
then we have that
$\widehat{L}M=M\widehat{\Lambda}$,
and the columns of $M$ are the eigenvectors corresponding to the largest $c$ eigenvalues of $\widehat{L}$.
Specifically, we divide $X^{(t)}$ into $c$ subsets by $Y^{(t)}$ beforehand and directly.
In summary, we propose the first restarted algorithm based on $K$-means.
\begin{breakablealgorithm}
\caption{A restarted spectral clustering algorithm with self-guiding based on $K$-means}\label{Alg1}
\begin{algorithmic}
\STATE  {{\bf Input:} Given the sample matrix $X=[\bm{x}_1,\bm{x}_2,\ldots,\bm{x}_n]^{T}\in \mathbb{R}^{n \times d}$, the cluster number $c$, the maximum number of iterations $itermax_1$, and the error tolerance $tol$;} \\
\STATE {{\bf 1.}Initialization: Given an initial classification label $Y^{(0)}$ and an orthonormal matrix $M_0\in\mathbb{R}^{n\times c}$, let $X^{(0)}=X=[X^{(0)}_1;X^{(0)}_2;\ldots;X^{(0)}_c]$ and $t=0$;}\\
\STATE {{\bf 2.}\textbf{while} $t\leq itermax_1$}\\
\STATE {{\bf 3.}~\textbf{for} $j=1,2,\ldots,c$ \textbf{do}$~~~\%$~Compute~ $M$}\\
\STATE {{\bf 4.}~~Sample $C_j$ from the rows of $X^{(t)}_j$ without forming $A_j$, and compute $W_j$;}\\
\STATE {{\bf 5.}~~Compute the economized QR factorization of $C_j$, and the SVD of $R_j(W_j)^{\dagger}(R_j)^{T}$, for  $U_j$ and $\Sigma_j$;}\\
\STATE {{\bf 6.}~~Form $\widehat{U}_j=\widehat{D}_j^{-\frac{1}{2}}U_j$, and perform the economized QR factorization of $\widehat{U}_j$, for $\widehat{Q}_j$ and $\widehat{R}_j$;}\\
\STATE {{\bf 7.}~~Let $\Delta_j=\widehat{R}_j\Sigma_j(\widehat{R}_j)^{T}$ and compute its largest eigenpair $(\hat{\lambda}_j,\bm{p}_j)$, and let $\bm{m}_j=\widehat{Q}_j\bm{p}_j$;}\\
\STATE {{\bf 8.}~\textbf{end for}}\\
\STATE {{\bf 9.}~~Let $M=diag(\bm{m}_1,\bm{m}_2,\ldots,\bm{m}_c)$, and run $K$-means algorithm on $M$ for the clustering indicator matrix $Y$;}\\
\STATE {{\bf 10.} ~~If \eqref{stopsin} is satisfied, then break, else continue};\\
\STATE {{\bf 12.} {\bf Self-guiding}: Reclassify $X^{(t)}$ by using the indicator matrix $Y$, and set $X^{(t+1)}=[X^{(t+1)}_1;X^{(t+1)}_2;\ldots;X^{(t+1)}_c]$;}\\
\STATE {{\bf 13.} Let $t=t+1$ and $M_0=M$;}\\
\STATE {{\bf 14.} \textbf{end while}}\\
\STATE {{\bf Output:}  {The indicator matrix $Y\in \mathbb{R}^{n \times c}$}.}
\end{algorithmic}
\end{breakablealgorithm}

\begin{remark}\label{re3}
Two remarks are given.
First, there is no need to store $X^{(t+1)}$ in each cycle. One only needs to rearrange the rows of $X$ without any additional storage requirements.
Second, we use sine of the angle between the subspaces spanned by two consecutive approximations to $M$ as the stopping criterion
\begin{equation}\label{stopsin}
\begin{split}
\|\sin\theta\big[\mathcal{R}(M_0),\mathcal{R}(M)\big]\|_F
=\!\|M_0-M(M^{T}M_0)\|_F\!<\!tol,
\end{split}
\end{equation}
where $tol$ is a user determined parameter.
\end{remark}

\subsubsection{Theoretical analysis}\label{subsec3.2.2}
In Algorithm \ref{Alg1}, we make use of a low-rank approximation $\widehat{A}$ to approximate the similarity matrix $\widetilde{A}$.
An interesting question is: Given the error between $\widehat{A}$ and $\widetilde{A}$, how large is the distance between the {\it approximate solution} $M$ and the
{\it exact} solution $M^{*}$? Our analysis is composed of two steps. First, we give insight into the derivation between  $\widehat{L}$ and $\widetilde{L}$. Second, we consider
the distance $\|\sin\theta[\mathcal{R}(M),\mathcal{R}(M^*)]\|_2$, i.e., the distance between the two subspaces spanned by $M$ and $M^{*}$.


As was discussed in Subsection \ref{Sec4.1}, let $T_j^{[r_j]}$ be a low-rank approximation to $A_j$, we have that \cite{11RS,15RNLS}:
$
\|A_j-T_j^{[r_j]}\|_2= \beta_j\sigma_{r_j+1},~j=1,2,\ldots,c,
$
where $\sigma_{r_j+1}$ is the $(r_j+1)$-th largest singular value of $R_jW_j^{\dagger}R_j^{T}$ and $\beta_j>0$ is a constant related to the matrix.
Let $A_j=[\bm{\alpha}_1,\bm{\alpha}_2,\ldots,\bm{\alpha}_{n_j}]$ and $T_j^{[r_j]}=[\hat{\bm{\alpha}}_1,\hat{\bm{\alpha}}_2,\ldots,\hat{\bm{\alpha}}_{n_j}]$, then
\begin{align*}
\|D_j-\widehat{D}_j\|_2\!&=\!\max_{1\leq l\leq n_j}\Big\{\Big|\big|\sum_{i=1}^{n_j}\alpha_{il}\big|-\big|\sum_{i=1}^{n_j}\hat{\alpha}_{il}\big|\Big|\Big\}
\!\leq\! \max_{1\leq l\leq n_j}\Big\{\big|\sum_{i=1}^{n_j}\alpha_{il}-\sum_{i=1}^{n_j}\hat{\alpha}_{il}\big|\Big\}\nonumber\\
&\!\leq\!\max_{1\leq l\leq n_j}\Big\{\sum_{i=1}^{n_j}\big|\alpha_{il}-\hat{\alpha}_{il}\big|\Big\}\!=\!\|A_j-T_j^{[r_j]}\|_1\nonumber\\ &\!\leq\!\sqrt{n_j}\|A_j-T_j^{[r_j]}\|_2\!=\!\sqrt{n_j}\beta_j\sigma_{r_j+1}:=\rho_j,\quad j=1,2,\ldots,c.
\end{align*}
As a result,
\begin{align}\label{eqna}
\|\widetilde{A}\!-\!\widehat{A}\|_{2}\!=\!\max_{1\leq j\leq c}\big\{\|A_j\!-\!T_j^{[r_j]}\|_2\big\}\!=\!\max_{1\leq j\leq c}\big\{\beta_j\sigma_{r_j+1}\big\}:=\epsilon_A
\end{align}
and
\begin{align}\label{eqnd}
\|\widetilde{D}\!-\!\widehat{D}\|_{2}\!=\!\max_{1\leq j\leq c }\{\|D_j\!-\!\widehat{D}_j\|_2\}\!\leq\!\max_{1\leq j\leq c}\{\sqrt{n_j}\beta_j\sigma_{r_j+1}\}:=\epsilon_D.
\end{align}

Let $\hat{d}_k$ be the smallest diagonal element of  $\widehat{D}=diag(\hat{d}_1,\hat{d}_2,\ldots,\hat{d}_n)$ in magnitude. Suppose that it is in the $i$-th block, and $|d_k-\hat{d}_k|\leq\|D_i-\widehat{D}_i\|_2\leq \rho_i$.
From \eqref{3.1.2} and \eqref{3.1.4}, we see that the elements of $A_i$ are in $[e^{-\frac{2}{\tau^{2}}},1]$, where the samples are normalized
using Euclidean norm. Without loss of generality, we assume that the number of samples in each class is larger than 1, and all the diagonal elements of $\widetilde{D}=diag(d_1,d_2,\ldots,d_n)$ are greater than or equal to $1+e^{-\frac{2}{\tau^{2}}}$, and $\hat{d}_k$ will be close to $d_k$ provided that $\rho_i$ is sufficiently small. Indeed, $\hat{d}_k \geq d_k-\rho_i\geq 1+e^{-\frac{2}{\tau^{2}}}-\rho_i.$


\begin{theorem}\label{Thm4.1}
Under the above notations, suppose that $\rho_i <1+e^{-\frac{2}{\tau^{2}}}$, then we have that
\begin{equation}\label{eqn15}
\|\widetilde{L}-\widehat{L}\|_{2}\leq \frac{\epsilon_A}{1\!+\!e^{-\frac{2}{\tau^2}}}\!+\frac{\epsilon_D\|\widehat{A}\|_{2}}{\sqrt{1\!+\!e^{-\frac{2}{\tau^2}}}\!\cdot\!\sqrt{1\!+\!e^{-\frac{2}{\tau^2}}\!-\!\rho_i}}.
\end{equation}
\end{theorem}
\begin{proof}
Notice that $\|\widetilde{D}^{\!-\!\frac{1}{2}}\|_{2}\leq (1\!+\!e^{-\frac{2}{\tau^2}})^{-\frac{1}{2}}$ and $\|\widehat{D}^{\!-\!\frac{1}{2}}\|_{2}\leq (1\!+\!e^{-\frac{2}{\tau^2}}\!-\!\rho_i)^{-\frac{1}{2}}$. Thus,
\begin{align}\label{eqn19}
&\|\widetilde{L}\!-\!\widehat{L}\|_{2}
\!=\!\|\widetilde{D}^{\!-\!\frac{1}{2}}\widetilde{A}\widetilde{D}^{\!-\!\frac{1}{2}}\!-\!\widetilde{D}^{\!-\!\frac{1}{2}}\widehat{A}\widehat{D}^{\!-\!\frac{1}{2}}\!+\!\widetilde{D}^{\!-\!\frac{1}{2}}\widehat{A}\widehat{D}^{\!-\!\frac{1}{2}}\!-\!\widehat{D}^{\!-\!\frac{1}{2}}\widehat{A}\widehat{D}^{\!-\!\frac{1}{2}}\|_{2}\nonumber\\
&\!\leq\!\|\widetilde{D}^{\!-\!\frac{1}{2}}\|_{2}\|\widetilde{A}\widetilde{D}^{\!-\!\frac{1}{2}}\!-\!\widehat{A}\widehat{D}^{\!-\!\frac{1}{2}}\|_{2}\!+\!\|\widetilde{D}^{\!-\!\frac{1}{2}}\!-\!\widehat{D}^{\!-\!\frac{1}{2}}\|_{2}\|\widehat{A}\|_{2}\|\widehat{D}^{\!-\!\frac{1}{2}}\|_{2}\nonumber\\
&\leq\!\frac{1}{\sqrt{1\!+\!e^{-\frac{2}{\tau^2}}}}\|\widetilde{A}\widetilde{D}^{\!-\!\frac{1}{2}}\!-\!\widehat{A}\widetilde{D}^{\!-\!\frac{1}{2}}\!+\!\widehat{A}\widetilde{D}^{\!-\!\frac{1}{2}}\!-\!\widehat{A}\widehat{D}^{\!-\!\frac{1}{2}}\|_{2}\!\!+\!\frac{1}{\sqrt{1\!+\!e^{-\frac{2}{\tau^2}}\!-\!\rho_i}}\|\widetilde{D}^{\!-\!\frac{1}{2}}\!-\!\widehat{D}^{\!-\!\frac{1}{2}}\|_{2}\|\widehat{A}\|_{2}\nonumber\\
&\!\leq\!\frac{1}{1\!+\!e^{-\frac{2}{\tau^2}}}\|\widetilde{A}\!-\!\widehat{A}\|_{2}\!+\!\Big(\frac{1}{\sqrt{1\!+\!e^{-\frac{2}{\tau^2}}}}\!+\!\frac{1}{\sqrt{1\!+\!e^{-\frac{2}{\tau^2}}\!-\!\rho_i}}\Big)\|\widetilde{D}^{\!-\!\frac{1}{2}}\!-\!\widehat{D}^{\!-\!\frac{1}{2}}\|_{2}\|\widehat{A}\|_{2}.
\end{align}

As $\widetilde{D}$ and $\widehat{D}$ are positive definite, we obtain from \cite[Theorem 6.2]{MF} and \eqref{eqnd} that
\begin{align}\label{3.3.27}
\|\widetilde{D}^{\frac{1}{2}}\!-\!\widehat{D}^{\frac{1}{2}}\|_{2}\!\leq\!\frac{\|\widetilde{D}\!-\!\widehat{D}\|_{2}}{\lambda^{\frac{1}{2}}_{\min}(\widetilde{D})\!+\!\lambda^{\frac{1}{2}}_{\min}(\widehat{D})}\leq\!\frac{\epsilon_D}{\sqrt{1\!+\!e^{-\frac{2}{\tau^2}}}\!+\!\sqrt{1\!+\!e^{-\frac{2}{\tau^2}}\!-\!\rho_i}}.
\end{align}
Consequently, we have from \cite[p.118, Theorem 2.5]{90MPT} and \eqref{3.3.27} that
\begin{align}
&\|\widetilde{D}^{-\frac{1}{2}}\!-\!\widehat{D}^{-\frac{1}{2}}\|_{2}\!\leq\!\|\widetilde{D}^{-\frac{1}{2}}\|_{2}\|\widetilde{D}^{\frac{1}{2}}\!-\!\widehat{D}^{\frac{1}{2}}\|_{2}\|\widehat{D}^{-\frac{1}{2}}\|_{2}\!\leq\!\frac{\|\widetilde{D}^{\frac{1}{2}}\!-\!\widehat{D}^{\frac{1}{2}}\|_{2}}{\sqrt{1\!+\!e^{-\frac{2}{\tau^2}}}\!\cdot\!\sqrt{1\!+\!e^{-\frac{2}{\tau^2}}\!-\!\rho_i}}\nonumber\\
&\!\leq\!\frac{1}{\sqrt{1\!+\!e^{-\frac{2}{\tau^2}}}\!\cdot\!\sqrt{1\!+\!e^{-\frac{2}{\tau^2}}\!-\!\rho_i}}\cdot\frac{\epsilon_D}{\sqrt{1\!+\!e^{-\frac{2}{\tau^2}}}\!+\!\sqrt{1\!+\!e^{-\frac{2}{\tau^2}}\!-\!\rho_i}}.\label{3.3.28}
\end{align}
A combination of \eqref{eqn19} and \eqref{3.3.28} yields
\begin{align*}\label{3.3.29}
\|\widetilde{L}\!-\!\widehat{L}\|_{2}\leq \frac{\epsilon_A}{1\!+\!e^{-\frac{2}{\tau^2}}}\!+\frac{\epsilon_D\|\widehat{A}\|_{2}}{\sqrt{1\!+\!e^{-\frac{2}{\tau^2}}}\!\cdot\!\sqrt{1\!+\!e^{-\frac{2}{\tau^2}}\!-\!\rho_i}},
\end{align*}
which completes the proof.
\end{proof}

\begin{remark}
Theorem \ref{Thm4.1} indicates that the distance between $\widetilde{L}$ and $\widehat{L}$ strongly relies on $\epsilon_A$ and $\epsilon_D$, i.e., the quality of the low-rank approximation to $\widetilde{A}$. In practical calculations, we find numerically that both $\hat{d}_k$ and the denominator of \eqref{eqn15} are often in the order of $\mathcal{O}(1)$.
\end{remark}

Theorem \ref{Thm4.1} builds a bridge between $\epsilon_A$ and the distance $\|\sin\theta[\mathcal{R}(M),\mathcal{R}(M^*)]\|_2$, from which we have the main theorem in this subsection.
\begin{theorem}\label{th3}
Under the above notations, let $\lambda^{*}_1\geq\cdots\geq\lambda^{*}_c>\lambda^{*}_{c+1}\geq\cdots\geq\lambda^{*}_n$ be the eigenvalues of $\widetilde{L}$, and suppose that $\lambda^{*}_c\!-\!\lambda^{*}_{c+1}\!>\!\|\widehat{L}\!-\!\widetilde{L}\|_2$, then
\begin{equation*}
\|\sin\theta[\mathcal{R}(M),\mathcal{R}(M^*)]\|_2\leq \frac{2\|\widehat{L}\!-\!\widetilde{L}\|_2}{\lambda^{*}_c\!-\!\lambda^{*}_{c+1}\!-\!\|\widehat{L}\!-\!\widetilde{L}\|_2}.
\end{equation*}
\end{theorem}
\begin{proof}
Let $\hat{\lambda}_1\geq\cdots\geq\hat{\lambda}_c\geq\hat{\lambda}_{c+1}\geq\cdots\geq\hat{\lambda}_n$ be the eigenvalues of $\widehat{L}$.
Let the eigendecompositions of $\widehat{L}$ and $\widetilde{L}$ be
$\widehat{L}[M,M_\perp]=[M,M_\perp]\ diag(
\Lambda,\Lambda_\perp)$
and
$\widetilde{L}[M^*,M^*_\perp]=[M^*,M^*_\perp]\ diag(
\Lambda^{*},\Lambda^{*}_\perp)$,
respectively, where both $[M,M_\perp]$ and $[M^*,M^*_\perp]$ are unitary matrices,
$\Lambda^{*}\!=\!diag(\lambda^{*}_1,\lambda^{*}_2,\ldots,\lambda^{*}_c),
\Lambda_\perp\!=\!diag(\hat{\lambda}_{c+1},\hat{\lambda}_{c+2},\ldots,\hat{\lambda}_n)$,
and $M,M^{*}\in\mathbb{R}^{n\times c}$, $M_\perp,M^*_\perp\in\mathbb{R}^{n\times (n-c)}$ are orthonormal matrices.

Let $\delta=\!\lambda^{*}_c\!-\!\hat{\lambda}_{c+1}$ be the minimum of the separation between the eigenvalues of $\Lambda^{*}$ and $\Lambda_\perp$. We have from {the Weyl's inequality} \cite[p.203, Corollary 4.10]{90MPT} that $\hat{\lambda}_{c+1}\leq\lambda^{*}_{c+1}\!+\!\|\widehat{L}\!-\!\widetilde{L}\|_2.$
Thus,
\begin{align}\label{eq26}
\delta\!\geq\lambda^{*}_c\!-\!\lambda^{*}_{c+1}\!-\!\|\widehat{L}\!-\!\widetilde{L}\|_2>0,
\end{align}
and
$
\hat{\lambda}_{c+1}=\lambda^{*}_c-\delta<\lambda^{*}_c-\frac{\delta}{2}.
$
As a result,
\begin{align}
\{\hat{\lambda}_{c+1},\hat{\lambda}_{c+2},\ldots,\hat{\lambda}_n\}\subset\mathbb{R}\backslash\Big[\lambda^{*}_c-\frac{\delta}{2},\lambda^{*}_1+\frac{\delta}{2}\Big].
\end{align}
So we have from the second $\sin\Theta$ theorem \cite[p.251, Theorem 3.6]{90MPT} that
\begin{align}
\|\sin\theta[\mathcal{R}(M),\mathcal{R}(M^*)]\|_2\!&\leq\!\frac{\|\widehat{L}M^*\!-\!M^*\Lambda^*\|_2}{\delta/2}=\frac{\|(\widehat{L}\!-\!\widetilde{L})M^*\!+\!\widetilde{L}M^*\!-\!M^*\Lambda^*\|_2}{\delta/2}\nonumber\\
&\!=\!\frac{2\|(\widehat{L}\!-\!\widetilde{L})M^*\|_2}{\delta}\leq \frac{2\|\widehat{L}\!-\!\widetilde{L}\|_2}{\delta}.\label{3.2.9}
\end{align}
A combination of \eqref{eq26} and \eqref{3.2.9} gives the result.
\end{proof}
\begin{remark}
Theorem \ref{th3} shows the rationality of inexact computations in spectral clustering. We mention that Theorem \ref{th3} may not be sharp in general. Indeed, it reveals that the distance between $M$ and $M^{*}$ is influenced by two factors. The first one is the approximate quality $\epsilon_A$, and the second one is the gap $\delta$. Therefore, if $\epsilon_A$ is not large and $\delta$ is not too small, then $M$ can be a good approximation to $M^{*}$.
\end{remark}

\subsection{A restarted spectral clustering based on spectral rotation}\label{subsec4.3}


In \cite{13SRSC}, it was pointed out that the solution obtained from the $K$-means algorithm may severely deviate from the true discrete solution, and spectral rotation could achieve better performance. Inspired by spectral clustering with spectral rotation, we attempt to propose
a restarted algorithm based on fusing spectral rotation and spectral embedding.

We notice that the degree matrix $\widetilde{D}$ is not an identity matrix. From the idea of block diagonal representation and
\eqref{2.1.6}, we obtain
\begin{equation}\label{3.3.1}
\underset{\substack{M \in \mathbb{R}^{n \times c}\\M^{T}M=I}}{\min} \|\widetilde{L}-MM^{T}\|_{F}^{2}.
\end{equation}
Thus, we propose the following model via combining spectral embedding and spectral rotation. More precisely, knitting \eqref{2.1.10} and \eqref{3.3.1} yields
\begin{align}
\underset{M,Y,Q}{\min}&f(M,Q,Y)\!=\!\underset{M,Y,Q}{\min}\{\|\widetilde{L}-MM^{T}\|_{F}^{2}\!+\!\lambda \|MQ-\widetilde{D}^{\frac{1}{2}}Y(Y^{T}\widetilde{D}Y)^{-\frac{1}{2}}\|_{F}^{2}\},\nonumber\\
& s.t. \ M^{T}M=I, Y \in \{0,1\}^{n \times c}, \boldsymbol{y}^{i}\mathbf{1}=1, Q^{T}Q=I.\label{3.3.2}
\end{align}

For large-scale problems, we approximate $\widetilde{L}$ and $\widetilde{D}$ by using $\widehat{L}$ and $\widehat{D}$, respectively; refer to \eqref{3.2.3.0}--\eqref{3.2.3.1}.
Hence, we solve the following model instead of \eqref{3.3.2}:
\begin{align}\label{333}
\underset{M,Y,Q}{\min} &f(M,Q,Y)\!=\!\underset{M,Y,Q}{\min}\{\|\widehat{L}-MM^{T}\|_{F}^{2}\!+\!\lambda \|MQ-\widehat{D}^{\frac{1}{2}}Y(Y^{T}\widehat{D}Y)^{-\frac{1}{2}}\|_{F}^{2}\},\nonumber\\
& s.t. \ M^{T}M=I,~ Y \in \{0,1\}^{n \times c}, \boldsymbol{y}^{i}\mathbf{1}=1, Q^{T}Q=I.
\end{align}
in practical calculations.


There are five variables $\widehat{D}$, $\widehat{L}$, $M$, $Q$ and $Y$ involved in \eqref{333}. Different from the known methods, the two matrices $\widehat{D}$ and $\widehat{L}$ are constructed by using the proposed self-guiding technique during cycles, while the other three are updated alternatively for given $\widehat{D}$ and $\widehat{L}$. Let us discuss it in more detail.


$\bullet$ \textbf{Construct the block diagonal matrices $\widehat{D}$ and $\widehat{L}$ during cycles}:
Similar to Subsection \ref{Sec4.1} and Subsection \ref{subsec4.2}, during the $t$-th cycle of the restarted algorithm, we make use of the proposed self-guiding technique to reclassify the data set $X^{(t)}$ into $c$ subsets, for constructing
$\widehat{D}$ and $\widehat{L}$; see \eqref{3.2.3.0}--\eqref{3.2.3.1}.

$\bullet$ \textbf{Update the spectral embedding $M$}:
Fixing the other variables other than $M$, the optimization problem \eqref{333} reduces to:

\begin{equation}\label{3.3.9}
\begin{split}
\underset{M^{T}M=I}{\max}\ \{\tr(M^{T}\widehat{L}M)\!+\!\lambda \tr(M^{T}S)\},
\end{split}
\end{equation}
where
$S=\widehat{D}^{\frac{1}{2}}Y(Y^{T}\widehat{D}Y)^{-\frac{1}{2}}Q^{T}$, and $\widehat{L}$ is defined in \eqref{3.2.3.1}.
Note that $\widehat{L}_j$ can be rewritten as $\widehat{L}_j=\widehat{L}^{h}_j(\widehat{L}^{h}_j)^{T}$, where $\widehat{L}^{h}_j=\widehat{U}_j\Sigma^{\frac{1}{2}}_j,~j=1,2,\ldots,c$.
One can solve \eqref{3.3.9} by using the generalized power iteration (GPI) method \cite[Algorithm 1]{17GPISM}.

$\bullet$ \textbf{Update the orthonormal matrix $Q$}:
Fixing the other variables other than $Q$, the optimization problem \eqref{333} reduces to
\begin{equation*}\label{3.3.12}
\underset{Q^{T}Q=I}{\min} \ \|MQ-\widehat{D}^{\frac{1}{2}}Y(Y^{T}\widehat{D}Y)^{-\frac{1}{2}}\|_{F}^{2},
\end{equation*}
which is mathematically equivalent to
$
\underset{Q^{T}Q=I}{\max} \tr(Q^{T}N)$,
where $N=M^{T}\widehat{D}^{\frac{1}{2}}Y(Y^{T}\widehat{D}Y)^{-\frac{1}{2}}.$
Therefore, the optimal solution is \cite[Theorem 1]{13SRSC}
\begin{equation}\label{3.3.15}
Q = UV^{T},
\end{equation}
where $U$ and $V$ are the matrices composed of the left and right singular vectors of $N$.

$\bullet$ \textbf{Update the clustering indicator matrix $Y$}: By \eqref{333},
with other variables fixed, we compute $Y$ from solving
\begin{align*}
\underset{Y}{\min} \ \|MQ-\widehat{D}^{\frac{1}{2}}Y(Y^{T}\widehat{D}Y)^{-\frac{1}{2}}\|_{F}^{2}\nonumber\\
 s.t. \ Y \in \{0,1\}^{n \times c}, \bm{y}^{i}\mathbf{1}=1,
\end{align*}
which is equivalent to
\begin{align}
\underset{Y}{\max} \ \tr((Y^{T}\widehat{D}Y)^{-\frac{1}{2}}Y^{T}\widehat{D}^{\frac{1}{2}}MQ)\nonumber\\
s.t. \ Y \in \{0,1\}^{n \times c}, {\bm y}^{i}\mathbf{1}=1.\label{3.3.17}
\end{align}

One common way to update ${\bm y}^{i}$ is based on the increment of objective function value \cite{22UMVSC,21FOER,21LGCD}.
However, a disadvantage of this scheme is that some columns in $Y$ could be zero, which is unreasonable. Another disadvantage is that the workload is heavy for large-scale problems. To partially cure these drawbacks, we propose the following strategy to solve \eqref{3.3.17}. Denote by $G=MQ\in \mathbb{R}^{n \times c}$, which is an orthonormal matrix. Then we can rewrite \eqref{3.3.17} as
\begin{align}
&\underset{Y}{\max} \ \tr((Y^{T}\widehat{D}Y)^{-\frac{1}{2}}Y^{T}\widehat{D}^{\frac{1}{2}}G)\nonumber\\
&\!=\!\underset{Y}{\max}\sum_{j=1}^{c}\frac{\bm{e}_j^{T}Y^{T}\widehat{D}^{\frac{1}{2}}G\bm{e}_j}{(\bm{e}_j^{T}Y^{T}\widehat{D}Y\bm{e}_j)^{\frac{1}{2}}}
\!=\!\underset{\bm{y}_j}{\max}\sum_{j=1}^{c}\frac{\bm{y}_j^{T}\widehat{D}^{\frac{1}{2}}\bm{g}_j}{(\bm{y}_j^{T}\widehat{D}\bm{y}_j)^{\frac{1}{2}}},\label{3.3.18}
\end{align}
where $\bm{y}_j$ and $\bm{g}_j$ be the $j$-th column of $Y$ and $G$, respectively.
Let $\bm{b}_j=\widehat{D}^{\frac{1}{2}}\bm{y}_j\in \mathbb{R}^{n}$, then \eqref{3.3.18} can be rewritten as
\begin{equation*}\label{3.3.19}
\begin{split}
\underset{\bm{b}_j}{\max} \ \sum_{j=1}^{c}\frac{\bm{b}_j^{T}\bm{g}_j}{(\bm{b}_j^{T}\bm{b}_j)^{\frac{1}{2}}}=\underset{\bm{b}_j}{\max}\ \sum_{j=1}^{c}\frac{\bm{b}_j^{T}\bm{g}_j}{\|\bm{b}_j\|_{2}\cdot\|\bm{g}_j\|_{2}}=\underset{\bm{b}_j}{\max}\ \sum_{j=1}^{c}\cos\angle(\bm{b}_j,\bm{g}_j)\leq c,
\end{split}
\end{equation*}
where we used $\|\bm{g}_j\|_2=1,j=1,2,\ldots,c$.
It is obvious to see that the above equation holds if and only if $\bm{b}_j=\bm{g}_j$, i.e., $Y=\widehat{D}^{-\frac{1}{2}}G$. However, the rows of $\widehat{D}^{-\frac{1}{2}}G$ may have negative entries. To deal with this problem, we reset the $j$-th element
in the $i$-th row in the following way
\begin{equation}\label{3.3.20}
\boldsymbol{y}^{i}_{j}=< \underset{}{\arg} \underset{j\in[1,c]}{\max}|\boldsymbol{g}_{j}^{i}| >,
\end{equation}
where $<\cdot>$ denotes $1$ if the argument is true or $0$ otherwise.

In summary, we present the second restarted algorithm based on spectral rotation.
\begin{breakablealgorithm}
\caption{A restarted spectral clustering algorithm with self-guiding based on spectral rotation}\label{Alg3}
\begin{algorithmic}
\STATE  {{\bf Input:} Given the sample matrix $X=[\bm{x}_1,\bm{x}_2,\ldots,\bm{x}_n]^{T}\in \mathbb{R}^{n \times d}$, the cluster number $c$, and the maximum number of iterations $itermax_2$;} \\
\STATE {{\bf 1.} Initialization: Given an initial classification label $Y^{(0)}$ and two orthonormal matrices $M_0\in \mathbb{R}^{n \times c}$ and $Q_0\in \mathbb{R}^{c \times c}$, let $X^{(0)}=X=[X^{(0)}_1;X^{(0)}_2;\ldots;X^{(0)}_c]$ and $t=0$;}\\
\STATE {{\bf 2.} \textbf{while} $t\leq itermax_2$}\\
\STATE {{\bf 3.} ~\textbf{for} $j=1,2,\ldots,c$ \textbf{do}} ~$\%$ Compute $\widehat{L}$\\
\STATE {{\bf 4.} ~~Sample $C_j$ from the rows of $X^{(t)}_j$ without forming $A_j$, and compute $W_j$;}\\
\STATE {{\bf 5.} ~~Compute the economized QR factorization of $C_j$, and the SVD of $R_j(W_j)^{\dagger}(R_j)^{T}$, for the computation of $U_j$ and $\Sigma_j$;}\\
\STATE {{\bf 6.}~~Compute $\widehat{U}_j=\widehat{D}_j^{-\frac{1}{2}}U_j$ and $\widehat{L}^{h}_j=\widehat{U}_j\Sigma^{\frac{1}{2}}_j$;}\\
\STATE {{\bf 7.} ~\textbf{end for}}\\
\STATE {{\bf 8.} Solve $M$ by the generalized power iteration (GPI) \cite[Algorithm 1]{17GPISM};}\\
\STATE {{\bf 9.} Update $Q$ by solving \eqref{3.3.15};}\\
\STATE {{\bf 10.} Update $Y$ by solving \eqref{3.3.20};}\\
\STATE {{\bf 11.} ~~If \eqref{stopsin} is satisfied, then break, else continue};\\
\STATE {{\bf 12.} {\bf Self-guiding}: Reclassify $X^{(t)}$ by using the indicator matrix $Y$, and set $X^{(t+1)}=[X^{(t+1)}_1;X^{(t+1)}_2;\ldots;X^{(t+1)}_c]$;}\\
\STATE {{\bf 13.} Let $t=t+1$, $Q_0=Q$, and $M_0=M$;}\\
\STATE {{\bf 14.} \textbf{end while}}\\
\STATE {{\bf Output:}  {The indicator matrix $Y\in \mathbb{R}^{n \times c}$}.}
\end{algorithmic}
\end{breakablealgorithm}

\begin{remark}
First, for the computation of $M$ in Step 8, we point out that there is no need to form the $n$-by-$n$ matrix $\widehat{L}$ in practice.
By virtue of the block-diagonal structure of $\widehat{L}$, we only need to utilize $\widehat{L}^{h}_j$ for the matrix-matrix products.
Second, one merit of our strategy over the ones used in \cite{22UMVSC,21FOER,21LGCD} is the cheapness for updating $Y$.
Although it is hard to prove theoretically, we find numerically that zero column rarely occurs in $Y$ in our method.
\end{remark}

Let us briefly discuss the computation and space complexities of Algorithm \ref{Alg1} and Algorithm \ref{Alg3}. The computational cost (in each cycle) and the (total) storage requirement are summarized in Table \ref{CA}.
\begin{table}[!h]
\begin{center}
\caption{\it Complexity analysis of Algorithm \ref{Alg1} and Algorithm \ref{Alg3}, where $d$, $c$ and $n_j$ denote the dimension of samples, the number of clusters, and the size of the $j$-th block $A_j,~j=1,2,\ldots,c$, respectively.}\label{CA}
\begin{threeparttable}
\resizebox{\linewidth}{!}{
\begin{tabular}{c| c c |c c}
\toprule
Method  &Detail &Computation complexity &Detail &Space complexity \\
\midrule
\multirow{6}{*}{Alg.1}
&Construction of $C_j$ &$\mathcal{O}(d\cdot n_jm_j)$ &For $C_j$ &$\mathcal{O}(n_jm_j)$\\
&QR decomposition for $U_j$   &$\mathcal{O}(n_jm_j^{2})$    &For $R_j(W_j)^{\dagger}(R_j)^{T}$ &$\mathcal{O}(m_j^{2})$\\
&SVD for $\Sigma_j$ &$\mathcal{O}(m^{3}_j)$ &For $\widehat{U}_j$  &$\mathcal{O}(n_{j}r_j)$ \\
&QR decomposition of $\widehat{U}_j$  &$\mathcal{O}(n_{j}r_j^{2})$  &For $M$  &$\mathcal{O}(n)$ \\
&$K$-means   &$\mathcal{O}(nc\ell_1)$\tnote{1}  &For $Y$ &$\mathcal{O}(n)$ \\
&Total &$\approx\mathcal{O}(d\sum_{j=1}^cn_jm_j+nc\ell_1)$ &Total &$\approx\mathcal{O}(n)$\\
\midrule
\multirow{6}{*}{Alg.2}
&Construction of $C_j$ &$\mathcal{O}(d\cdot n_{j}m_j)$ &For $C_j$ &$\mathcal{O}(n_jm_j)$\\
&QR decomposition for $U_j$  &$\mathcal{O}(n_jm_j^{2})$ &For $R_j(W_j)^{\dagger}(R_j)^{T}$ &$\mathcal{O}(m_j^{2})$\\
&SVD for $\Sigma_j$ &$\mathcal{O}(m^{3}_j)$ &For $\widehat{L}_j^{h}$ &$\mathcal{O}(\sum_{j=1}^cn_{j}r_j)$\\
&Update $M$ &$\mathcal{O}(nc^2\ell_2)$\tnote{2} &For $M$ &$\mathcal{O}(nc)$\\
&Update $Q$ &$\mathcal{O}(c^3)$ &For $Q$ &$\mathcal{O}(c^2)$\\
&Update $Y$ &$\mathcal{O}(nc)$  &For $Y$ &$\mathcal{O}(n)$\\
&Total &$\approx\mathcal{O}(d\sum_{j=1}^cn_jm_j+nc^2\ell_2)$ &Total &$\approx\mathcal{O}(\sum_{j=1}^cn_{j}r_j+nc)$\\
\bottomrule
\end{tabular}}
\begin{tablenotes}
\footnotesize
\item[1] $\ell_1$ is the number of iterations of $K$-means.
\item[2] $\ell_2$ represents the number of iterations of \cite[Algorithm 1]{17GPISM}.
\end{tablenotes}
\end{threeparttable}
\end{center}
\end{table}

\section{Experiments}\label{sec5}
In this section, we perform comprehensive experiments on real-world databases to show the numerical behavior of the proposed methods.
All the numerical experiments were run on a Sugon computer with 64 cores double Intel(R) Xeon(R) Platinum E5-2637 processors, with CPU 3.50 GHz and RAM 256 GB. The operation system is 64-bit Windows 10. All the numerical results were obtained from running the MATLAB R2018b software.


\begin{table}[!h]
\renewcommand\arraystretch{1.1}
\begin{center}
\caption{\it Details of the databases}\label{dataset}
\begin{tabular}{c c c c c}
\toprule
Database  &Samples($n$) &Dimensions($d$) &Clusters($c$) \\
\midrule
Yale &165 &1024 &15\\
wine &178 &13 &3\\
FERET &1400 &1024 &200\\
Extended YaleB  &2432    &1024  &38\\
AR &2600 &1200 &100\\
\hline
USPS  &11000   &256 &10\\
CMU-PIE    &11560   &1024  &68\\
20Newsgroups &18846   &26214 &20\\
\hline
MNIST        &70000   &784  &10\\
YoutubeFace  &103390  &16384  &38\\
\bottomrule
\end{tabular}
\end{center}
\end{table}


In the numerical experiments, we use five small databases for clustering visualization and statistical experiment, and five large-scale databases for comparison of numerical performances, including {\tt wine}\footnote{\url{https://archive.ics.uci.edu/ml/index.php}.}, five face image databases {\tt Yale}\footnote{\url{http://cvc.yale.edu/projects/yalefaces/yalefaces.html}.}, {\tt FERET}, {\tt Extended YaleB}\footnote{\url{http://cvc.yale.edu/projects/yalefacesB/yalefacesB.html}.}, {\tt AR}\footnote{\url{http:// rvl1. ecn. purdue. edu/ ~aleix/ aleix_ face_ DB. html}.} and {\tt CMU-PIE}\footnote{\url{http://www.cs.cmu.edu/afs/cs/project/PIE/web/}.}, a text database {\tt 20Newsgroups}\footnote{\url{http://qwone.com/~jason/20Newsgroups/}.}, two handwritten digit databases {\tt USPS} \cite{22SCSC} and {\tt MNIST}\footnote{\url{http://yann.lecun.com/exdb/mnist/}.}, as well as a video data {\tt YoutubeFace}\footnote{\url{https://www.cs.tau.ac.il/~wolf/ytfaces/}.}. The details of the databases are briefly given in Table \ref{dataset}.

In Experiment \ref{Sec5.1} and Experiment \ref{Sec5.2}, we compare Algorithm \ref{Alg1} and Algorithm \ref{Alg3} with nine ``baseline" algorithms including $K$-means \cite{79ACKA}, Rcut \cite{92RCPC}, Ncut \cite{00NCIS}, the self-supervised spectral clustering algorithm (SESC\footnote{\url{https://github.com/zyxforever/Self-SC-Code}. We thank Dr. Zhao Yunxiao for providing us the MATLAB codes of the method.}) \cite{22SESC},
the ultra-scalable spectral clustering and ensemble clustering (USPEC and USENC\footnote{\url{https://www.researchgate.net/publication/330760669}}) \cite{20USPEC},
the divide-and-conquer based large-scale spectral clustering method (DNSC \footnote{\url{https://github.com/Li-Hongmin/MyPaperWithCode}}) \cite{22DNSC}, the subspace clustering by finding good neighbors (FGNSC\footnote{\url{https://github.com/JLiangNKU/FGNSC}}) \cite{20FGNSC}, and the multiple kernel $K$-means clustering with simultaneous spectral rotation (MKKM\footnote{\url{https://github.com/MoetaYuko/MKKM-SR}}) \cite{22MKKM}. Default values are utilized in these algorithms.
Specifically, in the DNSC algorithm, the number of landmarks is taken as 1000 if the samples are greater than 1000, and 100 otherwise. In the MKKM algorithm, we choose Gaussian kernel, polynomial kernel, and linear kernel to construct multi-kernel matrix.

In Experiment \ref{Sec5.3}, we compare our two {\it restarting} algorithms (using initial guesses chosen randomly) with four {\it updating} algorithms, including the subspace clustering by block diagonal representation (BDRSC\footnote{\url{https://github.com/canyilu/Block-Diagonal-Representation-for-Subspace-Clustering}}) \cite{19BDRSC}, the auto-weighted simultaneous consensus graph learning and discretization method (ASCGLD\footnote{\url{https://github.com/Ekin102003/AwSCGLD}}) \cite{22SCGLD}, the unified one-step multi-view spectral clustering (UoMvSC\footnote{\url{https://github.com/guanyuezhen/UOMvSC}}) \cite{22UMVSC}, and the auto-weighted multi-view subspace clustering with multiple kernels (MVCMK\footnote{\url{https://github.com/huangsd/MVC-via-kernelized-graph-learning}}) \cite{19MVCMK}.

Seven commonly adopted clustering evaluation criteria, i.e., Clustering Accuracy (ACC), Normalized Mutual Information (NMI), Purity, Adjusted Rand Index (ARI), precision, Recall, and F-score are utilized for algorithm validation \cite{21FOER}. Moreover, we define {\tt Average} the average value of the seven evaluation metrics.
Notice that a larger value gives a better clustering result for these measures.
In the tables below, {\tt CPU} denotes the running time in seconds. All the numerical results are the mean from five runs.

\subsection{Clustering visualization and statistical experiment}\label{Sec5.1}

In this example, we try to show the effectiveness of our proposed methods for improving clustering performance via clustering visualization and statistical experiment. There are two experiments in this example.
In the first experiment, we visualize the spectral embedding obtained from the Rcut \cite{92RCPC}, DNSC \cite{22DNSC}, MKKM \cite{22MKKM}, as well as the two {\it restarting} algorithms (denoted by Baseline+Alg.1 and Baseline+Alg.2) on {\tt Extended YaleB} database, using the t-Distributed Stochastic Neighbor Embedding (t-SNE) technique \cite{08tsne}. For instance, ``$K$-means+Alg.1" stands for running Algorithm \ref{Alg1} with the clustering result obtained from the $K$-means method as the initial guess.

\begin{figure*}[h!]
\centering
\subfigure[Rcut]{
\centering
\includegraphics[width=3.65cm,height=3cm]{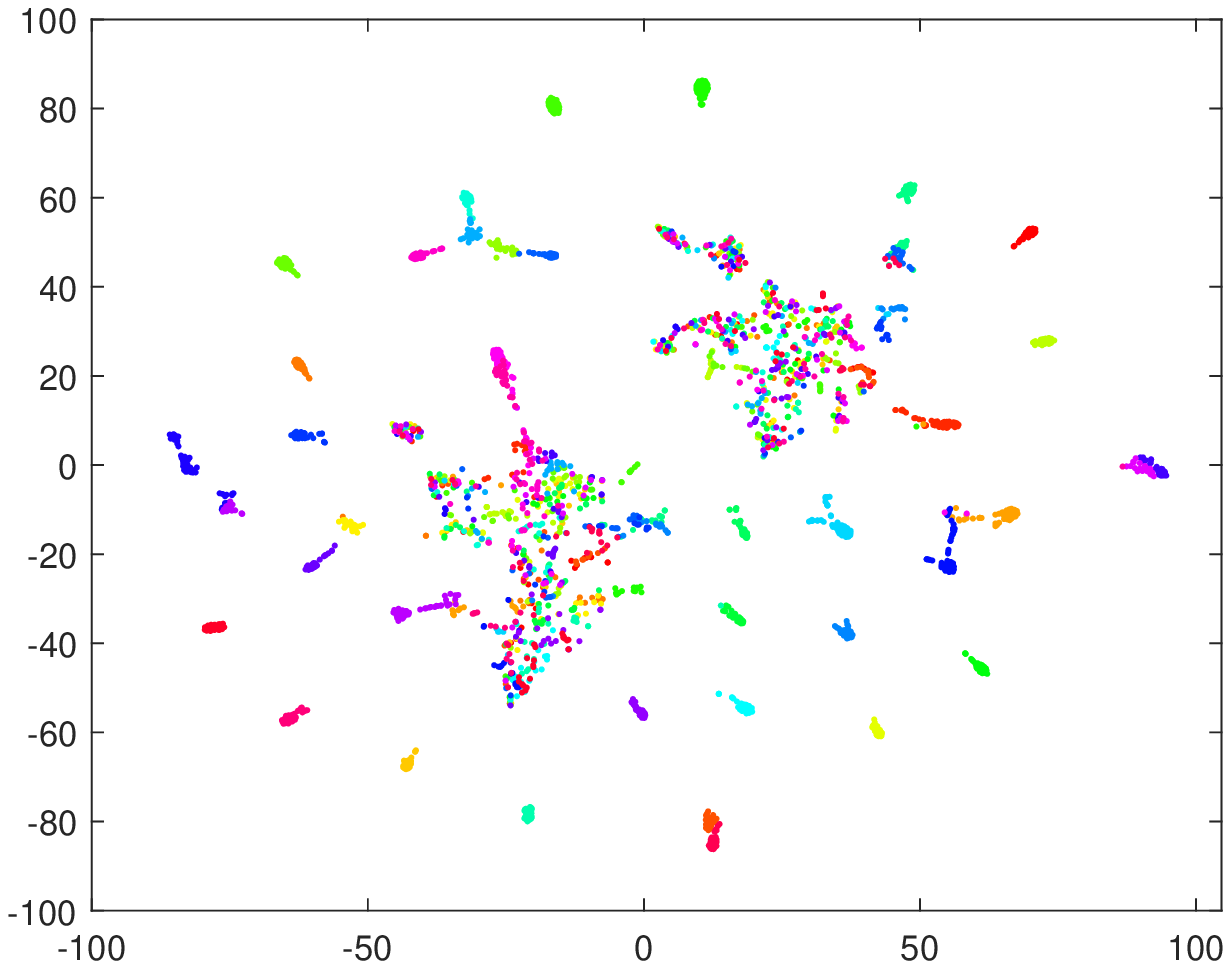}
}
\subfigure[Rcut+Alg.1]{
\centering
\label{tsnec}
\includegraphics[width=3.65cm,height=3cm]{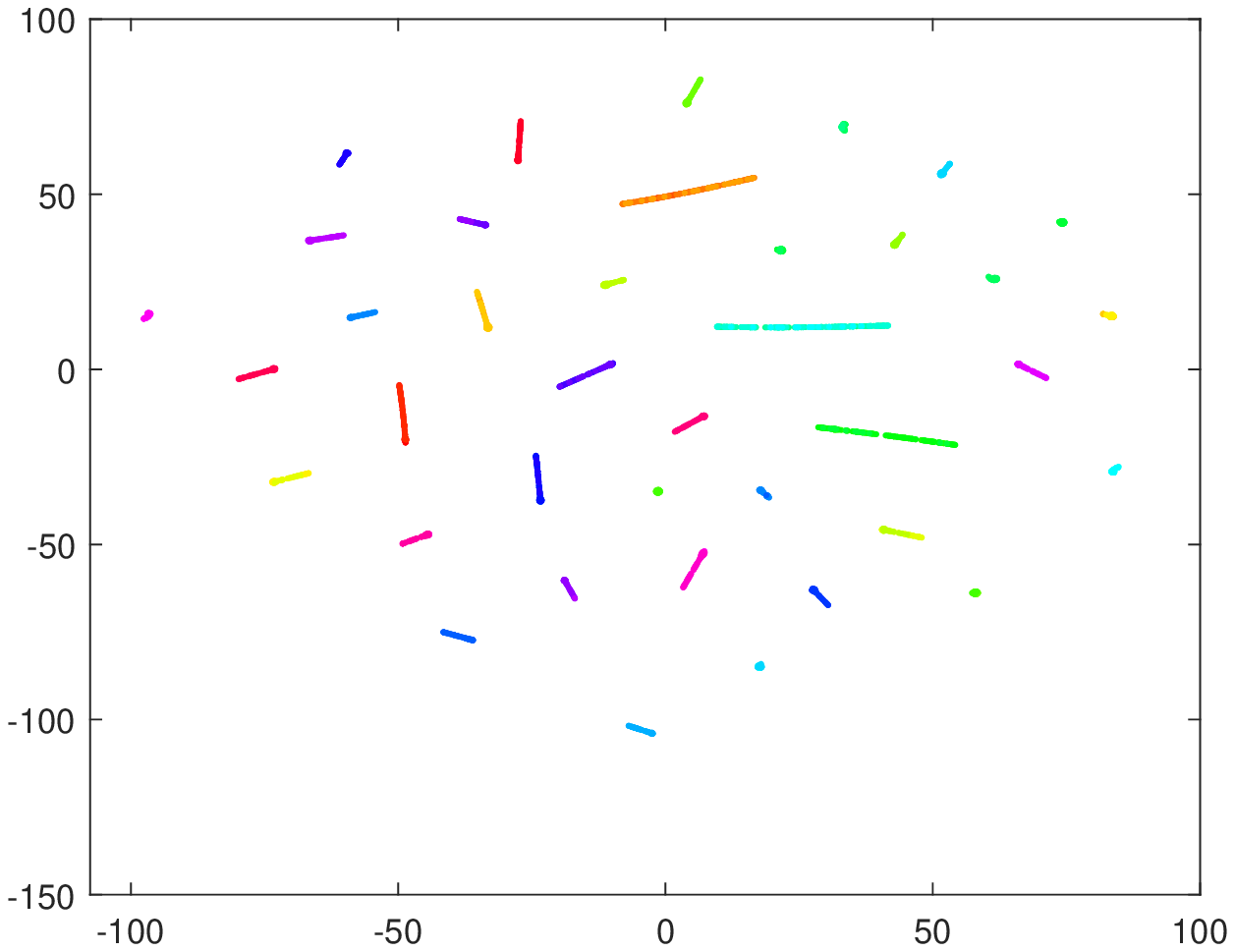}
}
\subfigure[Rcut+Alg.2]{
\centering
\label{tsned}
\includegraphics[width=3.65cm,height=3cm]{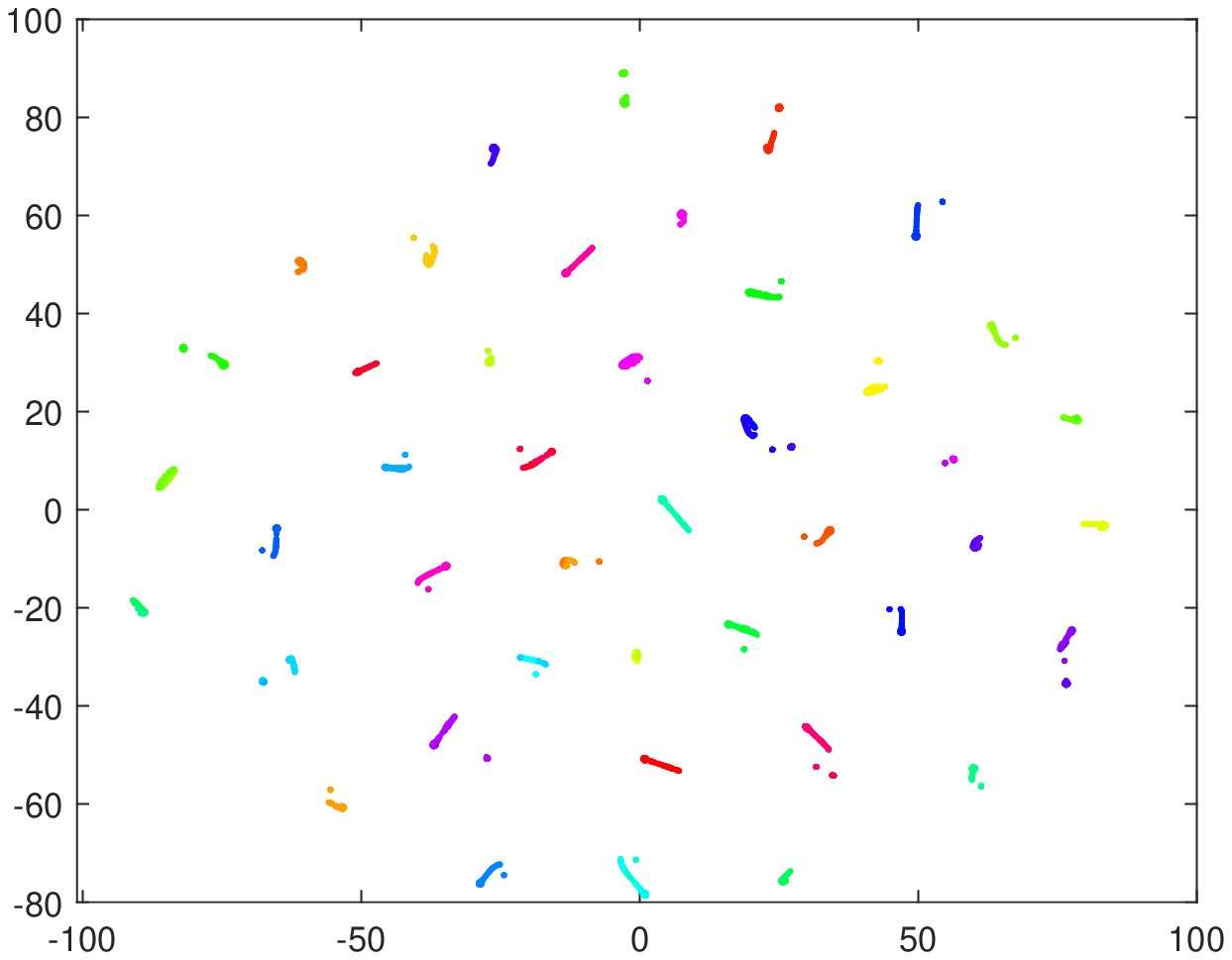}
}
\subfigure[DNSC]{
\centering
\includegraphics[width=3.65cm,height=3cm]{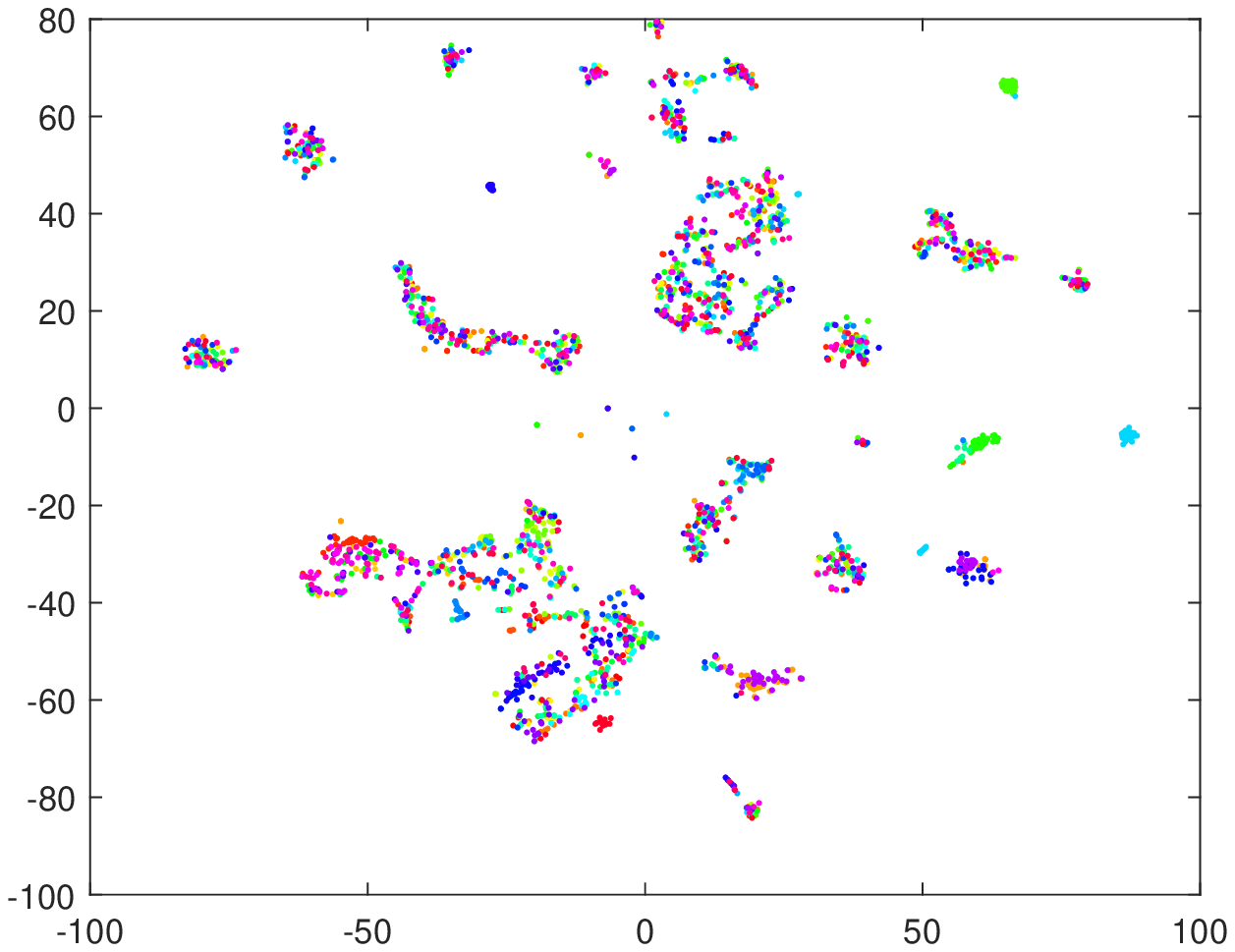}
}
\subfigure[DNSC+Alg.1]{
\centering
\label{tsnec}
\includegraphics[width=3.65cm,height=3cm]{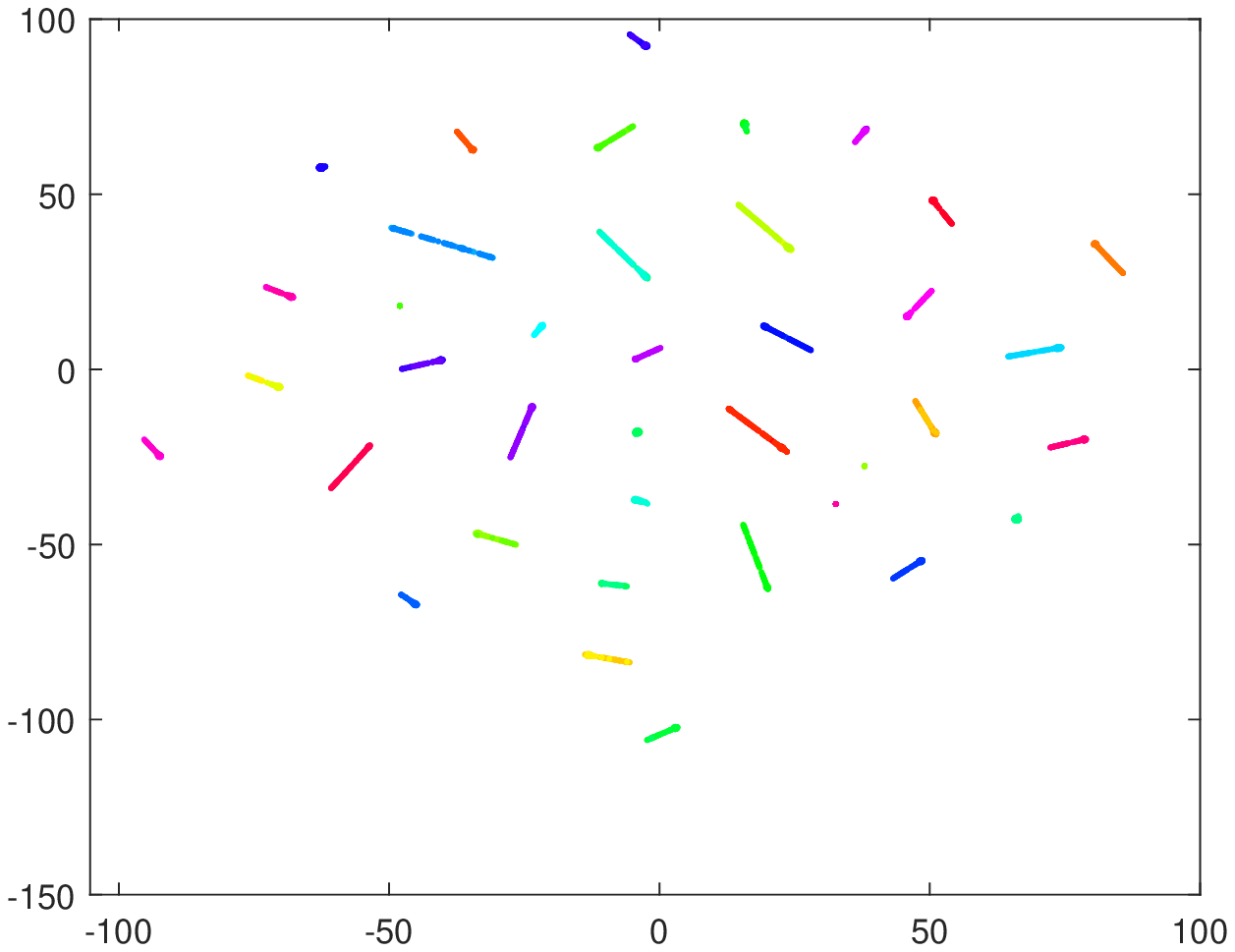}
}
\subfigure[DNSC+Alg.2]{
\centering
\label{tsned}
\includegraphics[width=3.65cm,height=3cm]{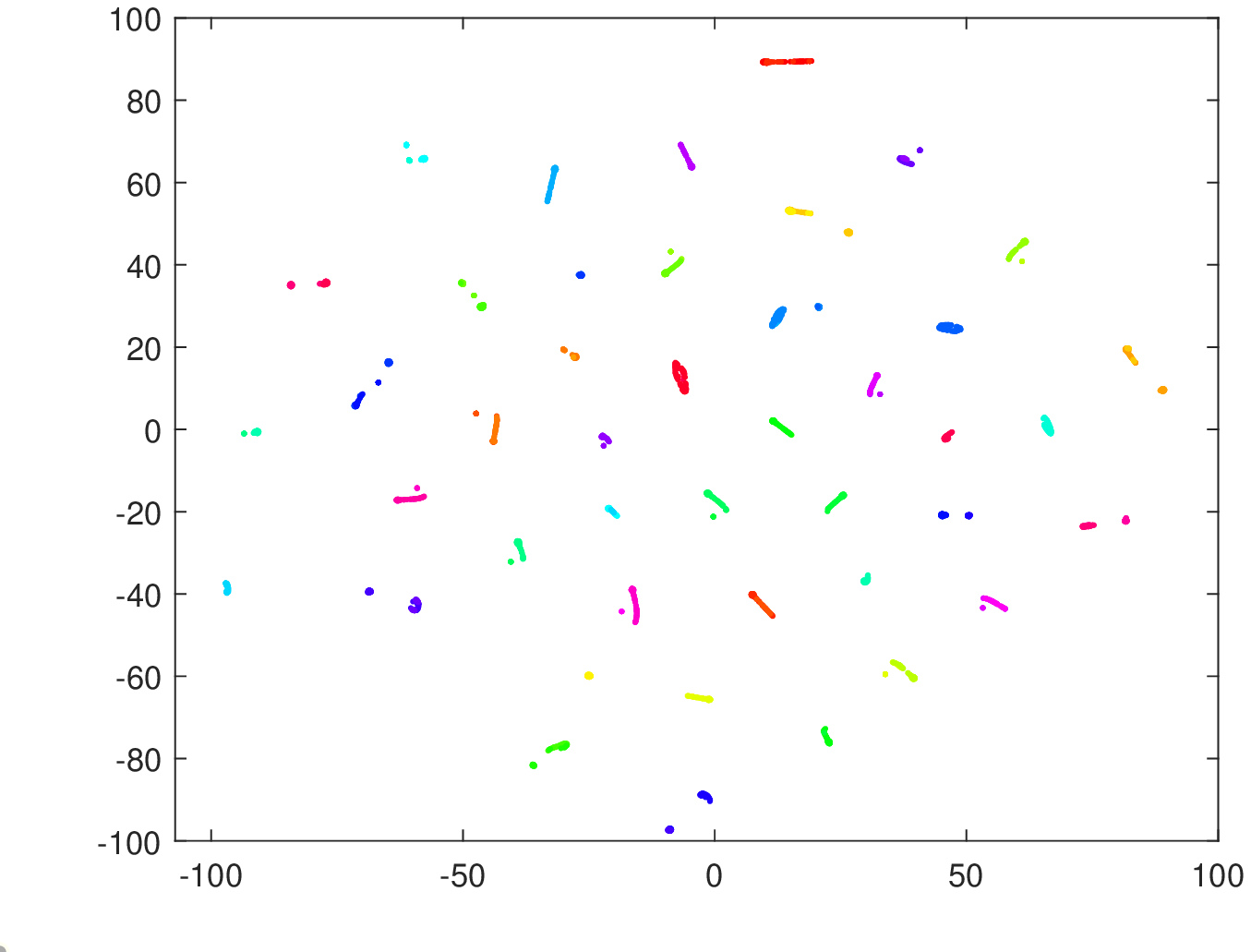}
}
\subfigure[MKKM]{
\centering
\includegraphics[width=3.65cm,height=3cm]{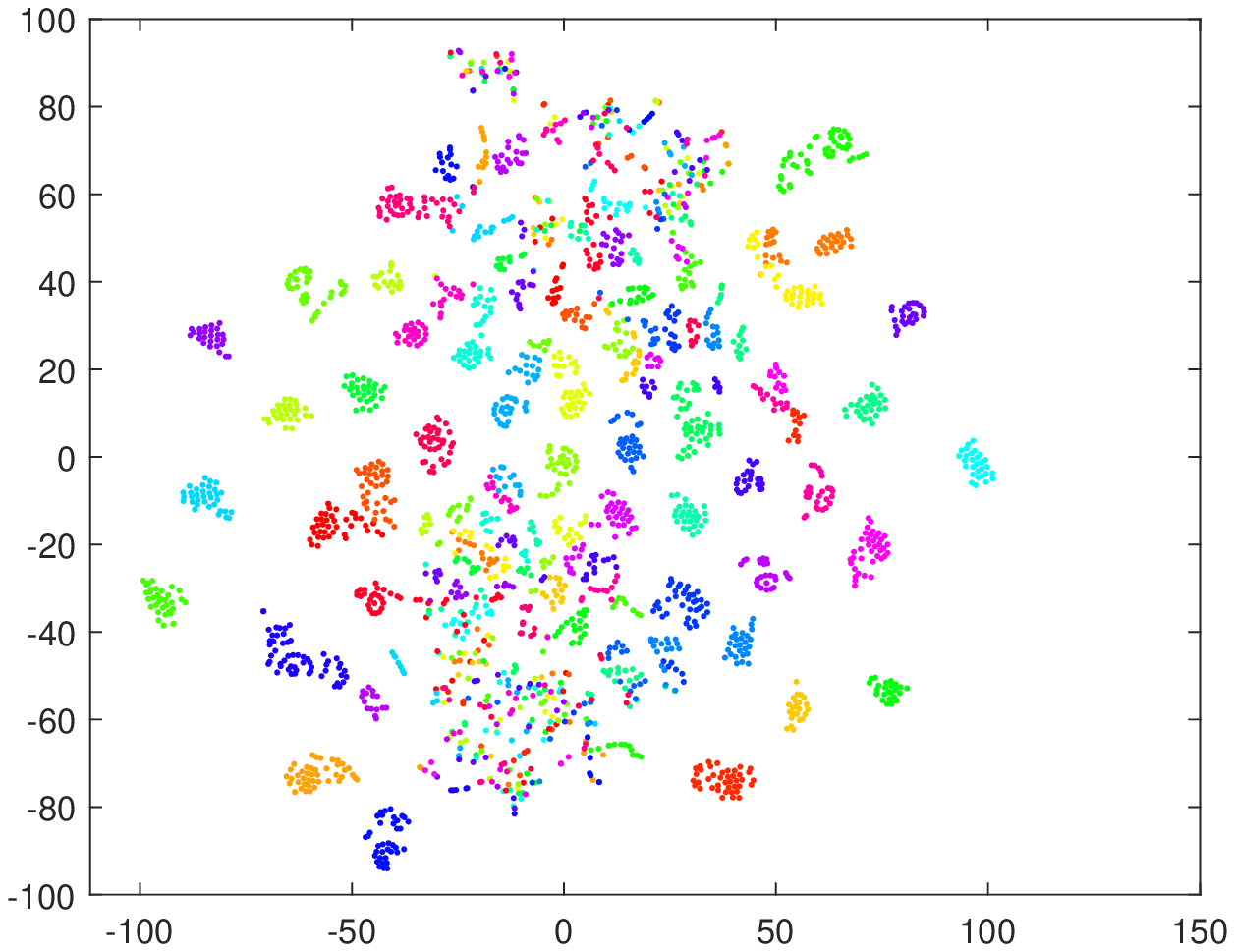}
}
\subfigure[MKKM+Alg.1]{
\centering
\label{tsnec}
\includegraphics[width=3.65cm,height=3cm]{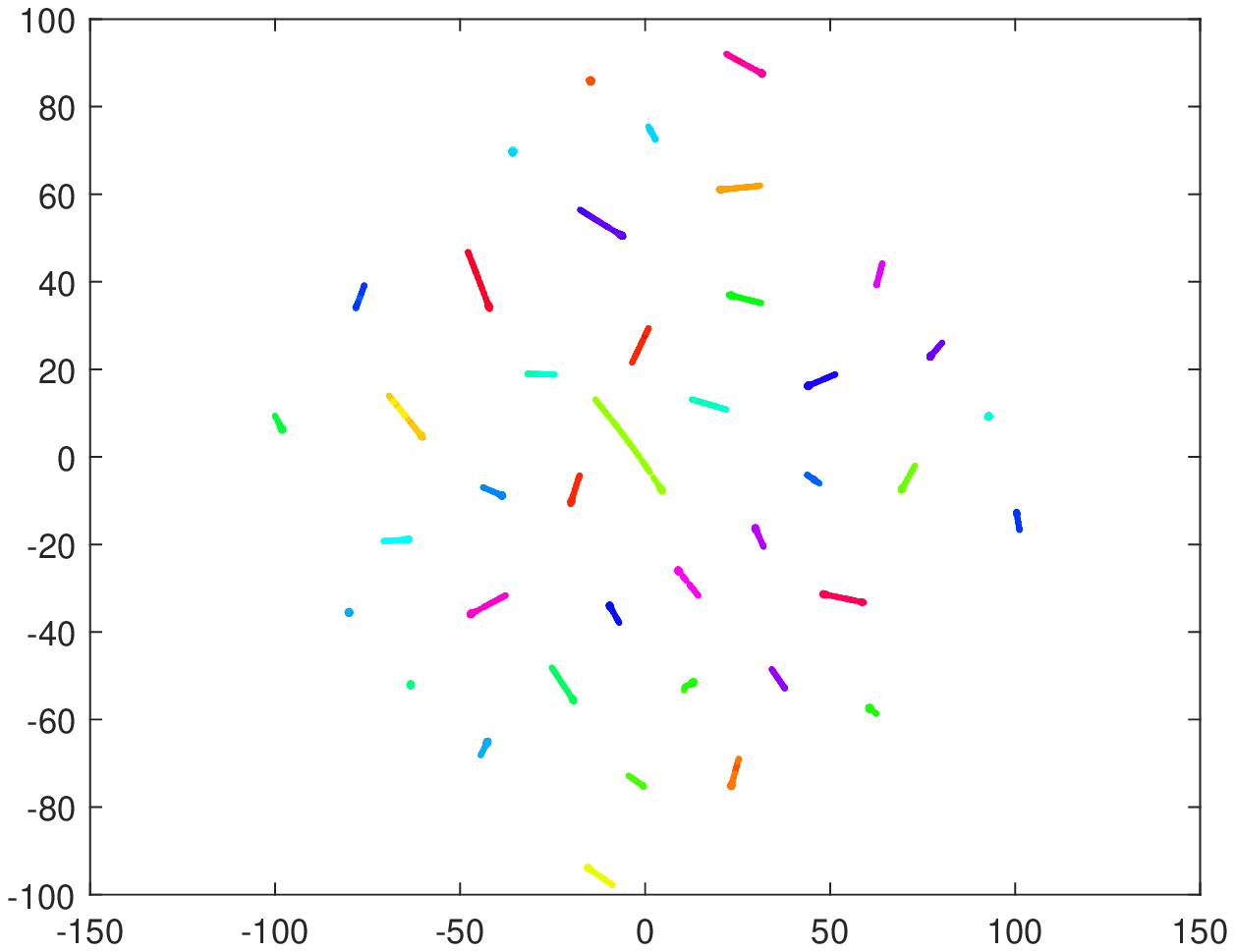}
}
\subfigure[MKKM+Alg.2]{
\centering
\label{tsned}
\includegraphics[width=3.65cm,height=3cm]{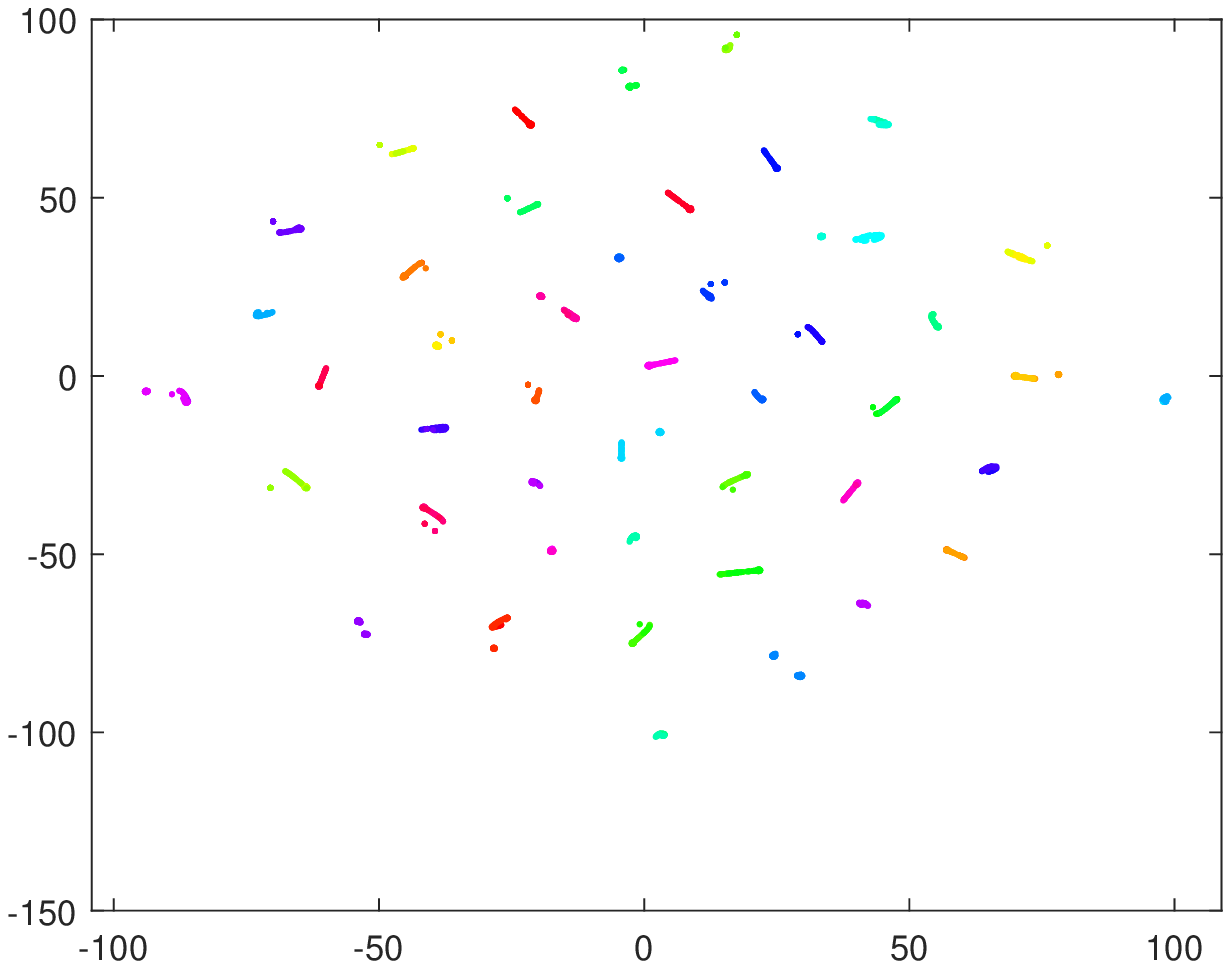}
}
\caption{Example \ref{Sec5.1}: Clustering visualization of the spectral embedding from different approaches on the {\tt Extended YaleB} database.
}
\label{tsne}
\end{figure*}

The results are plotted in Fig. \ref{tsne}, in which each point corresponds to an image and the color corresponds to the ground-truth clusters.
It is obvious to see from Fig. \ref{tsne} that both Algorithm \ref{Alg1} and Algorithm \ref{Alg3} improve the clustering performances of the corresponding algorithms significantly.

\begin{figure}[h!]
\centering
\includegraphics[width=9cm,height=4.5cm]{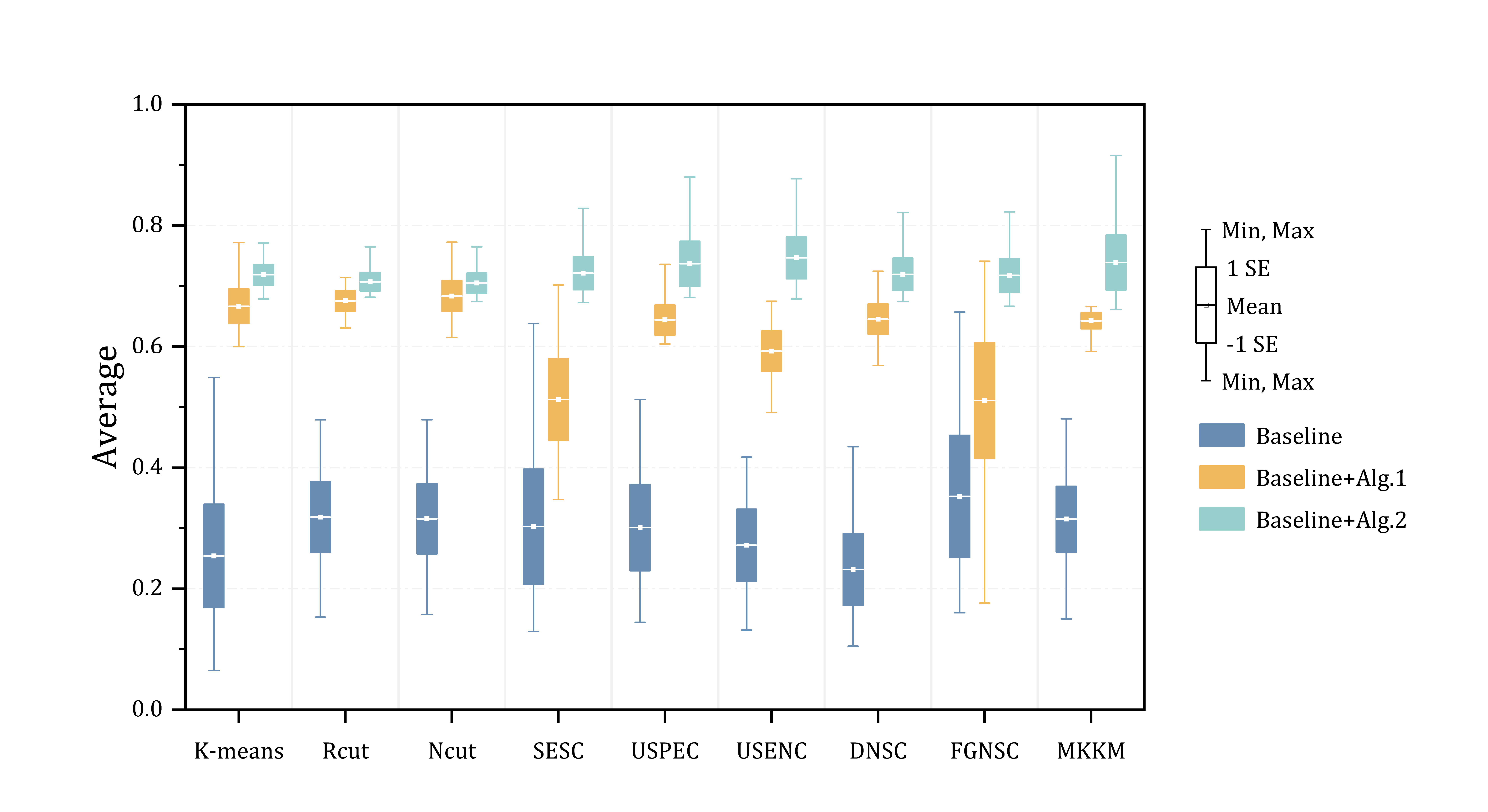}
\caption{Example \ref{Sec5.1}: Statistical experiments of nine ``baseline" algorithms and the corresponding eighteen restarted algorithms on five databases.}
\label{AVE}
\end{figure}

In the second experiment, we perform some statistical experiments to show the efficiency of our proposed algorithms
on improving the clustering performance. To do this, we plot in Fig. \ref{AVE} the values of Average
of the twenty-seven algorithms, i.e., nine ``baseline" algorithms ($K$-means \cite{79ACKA}, Rcut \cite{92RCPC}, Ncut \cite{00NCIS}, SESC \cite{22SESC}, USPEC and USENC \cite{20USPEC},
DNSC \cite{22DNSC}, FGNSC \cite{20FGNSC}, and MKKM \cite{22MKKM}), eighteen proposed algorithms including nine Baseline+Alg.1 and nine Baseline+Alg.2, on the five databases {\tt Yale}, {\tt wine}, {\tt FERET}, {\tt Extended YaleB} and {\tt AR}. Notice that the closer the Average value is to 1, the better, and the more concentrated of the box plot is, the more stable it is. It is easy to see that Algorithm \ref{Alg3} is better than Algorithm \ref{Alg1}, and the two proposed algorithms outperform the corresponding ``baseline" algorithms significantly in most cases.

\subsection{Effectiveness of the proposed restarting strategy for enhancing clustering performances}\label{Sec5.2}

To further demonstrate the superiority of our proposed two algorithms, we compare Algorithm \ref{Alg1} and Algorithm \ref{Alg3} with nine ``baseline" algorithms ($K$-means \cite{79ACKA}, Rcut \cite{92RCPC}, Ncut \cite{00NCIS}, SESC \cite{22SESC}, USPEC and USENC \cite{20USPEC},
DNSC \cite{22DNSC}, FGNSC \cite{20FGNSC}, and MKKM \cite{22MKKM}) on three large-scale databases {\tt USPS}, {\tt CMU-PIE} and {\tt 20Newsgroups}.

The aim of this experiment is to show that Algorithm \ref{Alg1} and Algorithm \ref{Alg3} can (significantly) enhance the performances of some state-of-the-art algorithms for spectral clustering, with relatively little additional costs. To do this, we run Algorithm \ref{Alg1} and Algorithm \ref{Alg3} by using the clustering results obtained from the nine ``baseline" algorithms as the initial guesses. Hence, the CPU time of our algorithms includes both that for run the ``baseline" algorithms and Algorithm \ref{Alg1} or Algorithm \ref{Alg3}. For instance, SESC+Alg.1 implies that we run SESC first, and then run Algorithm \ref{Alg1} with the classification label $Y$ got from SESC as the initial guess. The CPU time of SESC+Alg.1 includes both for running SESC and Algorithm \ref{Alg1}. The numerical results are listed in Tables \ref{resUSPS}--\ref{resNG}, and the best results according to Average are highlighted in bold.


\begin{table*}[h!]
\footnotesize{
\begin{center}
\caption{\it Example 5.2: Performances and CPU time (in seconds) of nine ``baseline" algorithms and eighteen corresponding restarting algorithms on the {\tt USPS} database.
 }\label{resUSPS}
\resizebox{\linewidth}{!}{
\begin{tabular}{c c c c c c c c c c}
\toprule
Method &ACC &NMI &Purity &ARI &precision &Recall &F-score &Average &CPU\\
\midrule
K-means &0.469	&0.463	&0.486	&0.315	&0.381	&0.388	&0.384	&0.413	&1.6\\
K-means+Alg.1 &0.832	&0.831	&0.832	&0.722	&0.743	&0.757	&0.750	&\bf{0.781}	&2.5\\
K-means+Alg.2 &0.825	&0.827	&0.827	&0.716	&0.725	&0.767	&0.745	&0.776	&12.2\\
\hline
Rcut  &0.698	&0.706	&0.720	&0.593	&0.594	&0.685	&0.636	&0.661	&3.4\\
Rcut+Alg.1 &0.767	&0.795	&0.777	&0.667	&0.657	&0.757	&0.703	&0.732	&4.5\\
Rcut+Alg.2 &0.887	&0.861	&0.887	&0.789	&0.801	&0.820	&0.810	&\bf{0.836}	&13.7\\
\hline
Ncut &0.621	&0.686	&0.710	&0.537	&0.534	&0.654	&0.588	&0.618	&3.0\\
Ncut+Alg.1 &0.671	&0.749	&0.707	&0.577	&0.567	&0.694	&0.624	&0.656	&4.3\\
Ncut+Alg.2 &0.816	&0.824	&0.826	&0.719	&0.734	&0.762	&0.747	&\bf{0.775}	&14.8\\
\hline
SESC &0.774	&0.738	&0.780	&0.652	&0.671	&0.705	&0.688	&0.715	&42.9\\
SESC+Alg.1 &0.794	&0.803	&0.796	&0.671	&0.688	&0.723	&0.705	&0.740	&44.1\\
SESC+Alg.2 &0.889	&0.858	&0.889	&0.786	&0.804	&0.812	&0.808	&\bf{0.835}	&51.3\\
\hline
USPEC &0.659	&0.676	&0.705	&0.551	&0.567	&0.633	&0.598	&0.627	&2.5\\
USPEC+Alg.1 &0.787	&0.812	&0.804	&0.687	&0.683	&0.762	&0.720	&\bf{0.751}	&3.7\\
USPEC+Alg.2 &0.764	&0.804	&0.777	&0.675	&0.687	&0.733	&0.709	&0.735	&12.8\\
\hline
USENC &0.767	&0.741	&0.767	&0.668	&0.697	&0.706	&0.702	&0.721	&34.7\\
USENC+Alg.1 &0.878	&0.859	&0.878	&0.779	&0.796	&0.806	&0.801	&0.828	&37.2\\
USENC+Alg.2 &0.945	&0.913	&0.945	&0.883	&0.889	&0.901	&0.895	&\bf{0.910}	&45.7\\
\hline
DNSC   &0.616	&0.655	&0.660	&0.521	&0.542	&0.606	&0.572	&0.596	&2.1\\
DNSC+Alg.1 &0.762	&0.819	&0.797	&0.697	&0.691	&0.772	&0.729	&0.753	&3.3\\
DNSC+Alg.2 &0.858	&0.856	&0.865	&0.777	&0.786	&0.815	&0.800	&\bf{0.823}	&10.8\\
\hline
FGNSC  &0.553	&0.484	&0.599	&0.359	&0.405	&0.450	&0.426	&0.468	&3999.2\\
FGNSC+Alg.1 &0.704	&0.783	&0.737	&0.627	&0.633	&0.703	&0.666	&0.693	&3999.6\\
FGNSC+Alg.2 &0.813	&0.822	&0.821	&0.712	&0.730	&0.754	&0.742	&\bf{0.771}	&4007.5\\
\hline
MKKM &0.477	&0.437	&0.508	&0.298	&0.365	&0.372	&0.368	&0.404	&60.4\\
MKKM+Alg.1 &0.894	&0.874	&0.894	&0.804	&0.816	&0.832	&0.824	&\bf{0.848}	&61.7\\
MKKM+Alg.2 &0.868	&0.846	&0.868	&0.758	&0.769	&0.798	&0.783	&0.813	&70.1\\
\bottomrule
\end{tabular}}
\end{center}}
\end{table*}

\begin{table*}[h!]
\footnotesize{
\begin{center}
\caption{\it Example 5.2: Performances and CPU time (in seconds) of nine ``baseline" algorithms and eighteen corresponding restarting algorithms on the {\tt CMU-PIE} database.
}\label{resPIE}
\resizebox{\linewidth}{!}{
\begin{tabular}{c c c c c c c c c c}
\toprule
Method &ACC &NMI &Purity &ARI &precision &Recall &F-score &Average &CPU\\
\midrule
K-means &0.088	&0.205	&0.113	&0.027	&0.038	&0.051	&0.043	&0.081	&3.1\\
K-means+Alg.1&0.623	&0.846	&0.665	&0.591	&0.519	&0.703	&0.598	&0.649	&5.0\\
K-means+Alg.2 &0.715	&0.874	&0.735	&0.646	&0.633	&0.670	&0.651	&\bf{0.703}	&111.2\\
\hline
Rcut  &0.234	&0.473	&0.263	&0.120	&0.111	&0.176	&0.136	&0.216	&8.5\\
Rcut+Alg.1 &0.576	&0.832	&0.639	&0.533	&0.442	&0.699	&0.541	&0.609	&10.6\\
Rcut+Alg.2 &0.741	&0.880	&0.756	&0.664	&0.650	&0.689	&0.669	&\bf{0.721}	&119.4\\
\hline
Ncut  &0.222	&0.465	&0.261	&0.116	&0.105	&0.178	&0.132	&0.211	&7.2\\
Ncut+Alg.1 &0.548	&0.822	&0.609	&0.510	&0.414	&0.697	&0.519	&0.588	&9.4\\
Ncut+Alg.2 &0.706	&0.870	&0.728	&0.634	&0.623	&0.658	&0.640	&\bf{0.694}	&115.9\\
\hline
SESC  &0.250	&0.435	&0.292	&0.060	&0.049	&0.253	&0.083	&0.203	&229.5\\
SESC+Alg.1 &0.459	&0.727	&0.540	&0.218	&0.142	&0.726	&0.237	&0.436	&231.7\\
SESC+Alg.2 &0.732	&0.878	&0.750	&0.658	&0.643	&0.684	&0.663	&\bf{0.715}	&330.6\\
\hline
USPEC  &0.108	&0.244	&0.125	&0.044	&0.055	&0.063	&0.059	&0.100	&5.5\\
USPEC+Alg.1  &0.679	&0.865	&0.712	&0.625	&0.587	&0.682	&0.631	&0.683	&7.2\\
USPEC+Alg.2 &0.722	&0.876	&0.739	&0.648	&0.635	&0.672	&0.653	&\bf{0.706}	&106.8\\
\hline
USENC  &0.114	&0.227	&0.124	&0.050	&0.059	&0.073	&0.066	&0.102	&72.1\\
USENC+Alg.1 &0.657	&0.859	&0.696	&0.612	&0.560	&0.690	&0.618	&0.670	&74.0\\
USENC+Alg.2 &0.714	&0.871	&0.732	&0.638	&0.626	&0.662	&0.643	&\bf{0.698}	&173.9\\
\hline
DNSC  &0.102	&0.210	&0.113	&0.037	&0.048	&0.056	&0.052	&0.088	&4.2\\
DNSC+Alg.1 &0.668	&0.863	&0.703	&0.616	&0.577	&0.674	&0.622	&0.675	&5.8\\
DNSC+Alg.2 &0.717	&0.874	&0.736	&0.644	&0.633	&0.666	&0.649	&\bf{0.703}	&111.3\\
\hline
FGNSC  &0.312	&0.551	&0.338	&0.182	&0.159	&0.259	&0.197	&0.285	&5155.9\\
FGNSC+Alg.1 &0.600	&0.834	&0.654	&0.526	&0.431	&0.704	&0.535	&0.612	&5157.4\\
FGNSC+Alg.2 &0.700	&0.871	&0.724	&0.633	&0.619	&0.659	&0.638	&\bf{0.692}	&5267.6\\
\hline
MKKM  &0.301	&0.455	&0.321	&0.148	&0.141	&0.192	&0.162	&0.246	&186.6\\
MKKM+Alg.1 &0.631	&0.852	&0.684	&0.573	&0.503	&0.687	&0.581	&0.644	&188.3\\
MKKM+Alg.2 &0.721	&0.875	&0.739	&0.648	&0.635	&0.672	&0.653	&\bf{0.706}	&288.1\\
\bottomrule
\end{tabular}}
\end{center}}
\end{table*}

\begin{table*}[h!]
\footnotesize{
\begin{center}
\caption{\it Example 5.2: Performances and CPU time (in seconds) of nine ``baseline" algorithms and eighteen corresponding restarting algorithms on the {\tt 20Newsgroups} database, where ``--" means the algorithm fail to work within 12 hours.
}\label{resNG}
\resizebox{\linewidth}{!}{
\begin{tabular}{c c c c c c c c c c}
\toprule
Method &ACC &NMI &Purity &ARI &precision &Recall &F-score &Average &CPU\\
\midrule
K-means  &0.261	&0.249	&0.287	&0.088	&0.119	&0.184	&0.144	&0.190 &4.9\\
K-means+Alg.1 &0.572	&0.767	&0.631	&0.521	&0.452	&0.703	&0.550	&0.599	&9.5\\
K-means+Alg.2 &0.736	&0.833	&0.753	&0.652	&0.656	&0.683	&0.670	&\bf{0.712}	&58.9\\
\hline
Rcut  &0.391	&0.390	&0.406	&0.218	&0.205	&0.395	&0.270	&0.325&23.1\\
Rcut+Alg.1 &0.535	&0.714	&0.562	&0.481	&0.391	&0.753	&0.515	&0.565	&26.0\\
Rcut+Alg.2 &0.717	&0.826	&0.739	&0.634	&0.639	&0.667	&0.653	&\bf{0.696}	&76.7\\
\hline
Ncut  &0.386	&0.409	&0.412	&0.235	&0.226	&0.379	&0.283	&0.333	&22.1\\
Ncut+Alg.1 &0.552	&0.752	&0.602	&0.518	&0.437	&0.734	&0.548	&0.592	&25.1\\
Ncut+Alg.2 &0.727	&0.831	&0.747	&0.639	&0.647	&0.670	&0.658	&\bf{0.703}	&77.0\\
\hline
SESC  &0.065	&0.006	&0.066	&0.001	&0.051	&0.385	&0.090	&0.095	&3026.3\\
SESC+Alg.1 &0.287	&0.438	&0.359	&0.109	&0.107	&0.809	&0.189	&0.328	&3031.2\\
SESC+Alg.2 &0.695	&0.820	&0.721	&0.615	&0.620	&0.651	&0.635	&\bf{0.680}	&3078.1\\
\hline
USPEC  &0.254	&0.223	&0.274	&0.071	&0.100	&0.201	&0.133	&0.179	&172.3\\
USPEC+Alg.1 &0.500	&0.719	&0.570	&0.450	&0.365	&0.734	&0.487	&0.546	&175.6\\
USPEC+Alg.2  &0.746	&0.830	&0.764	&0.650	&0.654	&0.683	&0.668	&\bf{0.714}	&222.9\\
\hline
USENC &0.293	&0.284	&0.323	&0.096	&0.117	&0.238	&0.157	&0.215 &3303.0\\
USENC+Alg.1 &0.525	&0.729	&0.600	&0.440	&0.356	&0.723	&0.477	&0.550	&3306.7\\
USENC+Alg.2 &0.714	&0.828	&0.743	&0.638	&0.644	&0.670	&0.657	&\bf{0.699}	&3354.6\\
\hline
DNSC &0.217	&0.192	&0.240	&0.058	&0.093	&0.160	&0.118	&0.154	&132.1\\
DNSC+Alg.1 &0.529	&0.752	&0.607	&0.512	&0.430	&0.736	&0.543	&0.587	&135.0\\
DNSC+Alg.2 &0.728	&0.823	&0.740	&0.629	&0.638	&0.659	&0.648	&\bf{0.695}	&189.0\\
\hline
FGNSC  &--&--&--&--&--&--&--&--&-- \\
FGNSC+Alg.1 &--&--&--&--&--&--&--&--&-- \\
FGNSC+Alg.2 &--&--&--&--&--&--&--&--&-- \\
\hline
MKKM &0.237	&0.206	&0.259	&0.081	&0.119	&0.150	&0.133	&0.169 &259.7\\
MKKM+Alg.1 &0.614	&0.789	&0.669	&0.560	&0.523	&0.663	&0.585	&0.629 	&262.8\\
MKKM+Alg.2 &0.708	&0.822	&0.731	&0.621	&0.627	&0.656	&0.641	&\bf{0.686} 	&312.1\\
\bottomrule
\end{tabular}}
\end{center}}
\end{table*}

First, we observe from Tables \ref{resUSPS}--\ref{resNG} that, for these large-scale problems, the proposed algorithms improve all the seven clustering evaluation criteria, and  Algorithm \ref{Alg1} outperforms the ``baseline" algorithm while Algorithm \ref{Alg3} is better than Algorithm \ref{Alg1}. Indeed, our algorithms can even improve these criteria for 3 to 10 times. For example, for the {\tt 20Newsgroups} database (see Table \ref{resNG}), the ACC value of SESC is 0.065, while those of SESC+Alg.1 and SESC+Alg.2 are 0.287 and 0.695, respectively. These demonstrate the efficiency of the proposed restarting strategy.

\begin{figure*}[h!]
\centering
\subfigure[$K$-means+Alg.1]{
\centering
\includegraphics[width=3.7cm,height=3cm]{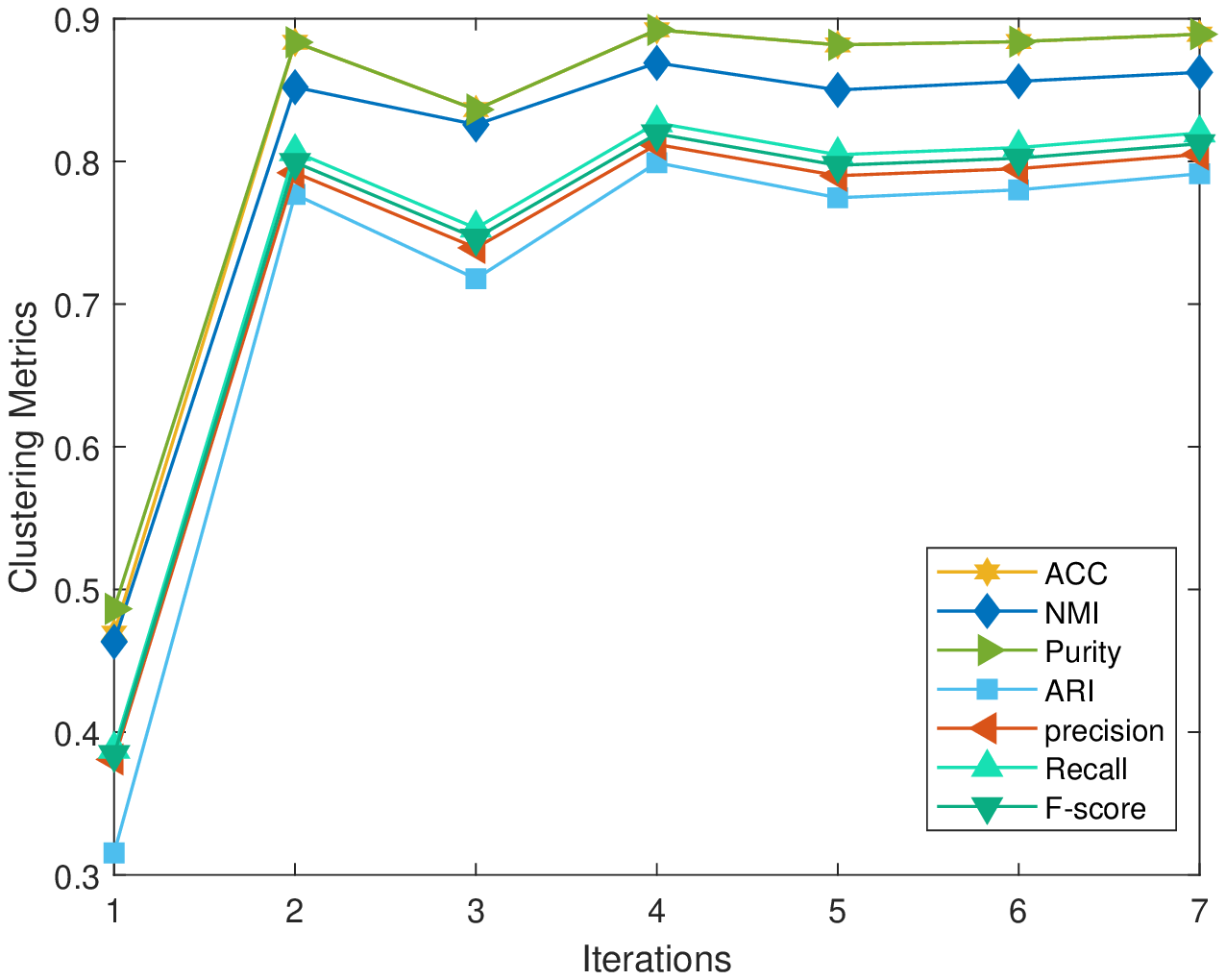}
}
\subfigure[Rcut+Alg.1]{
\centering
\includegraphics[width=3.7cm,height=3cm]{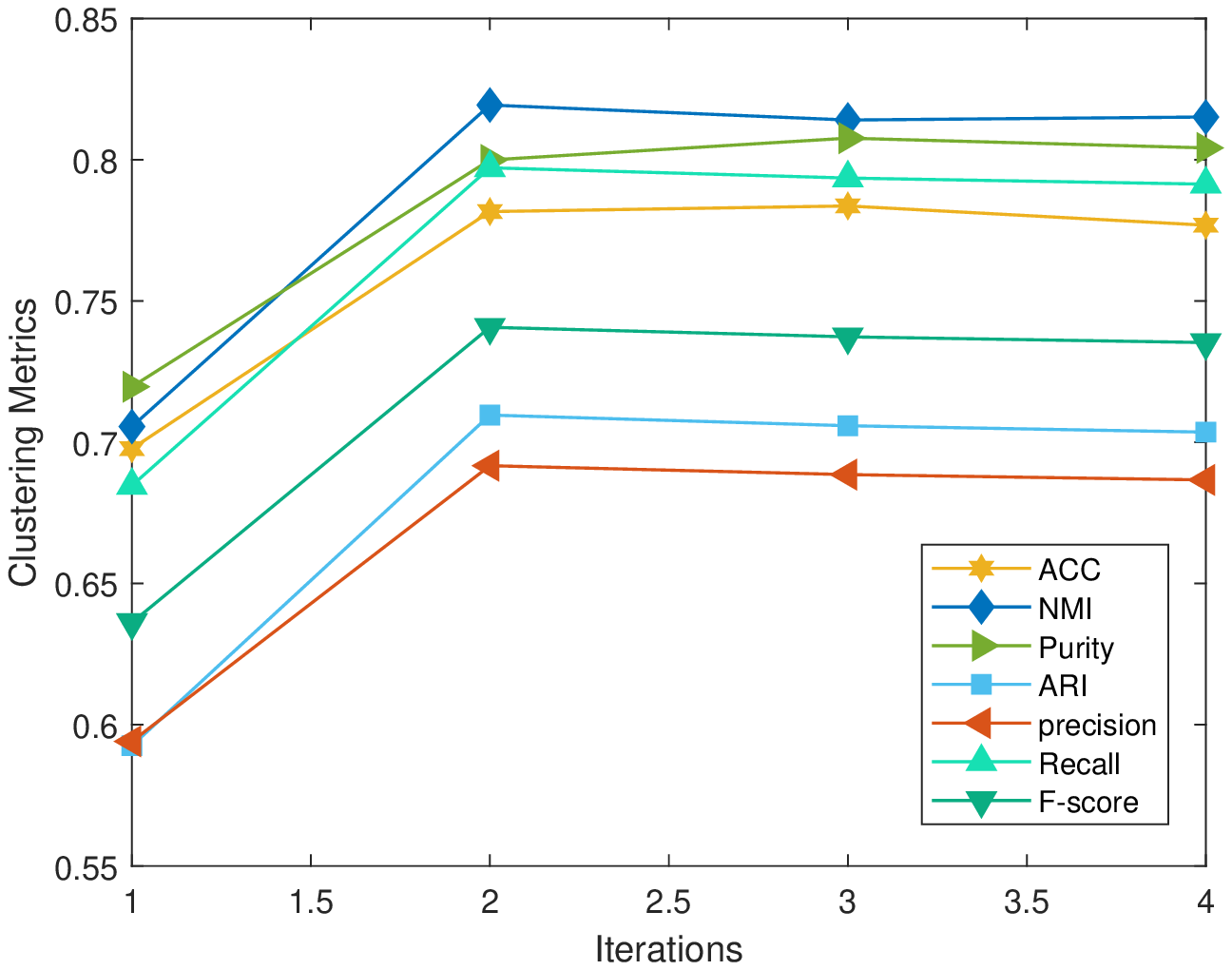}
}
\subfigure[Ncut+Alg.1]{
\centering
\includegraphics[width=3.7cm,height=3cm]{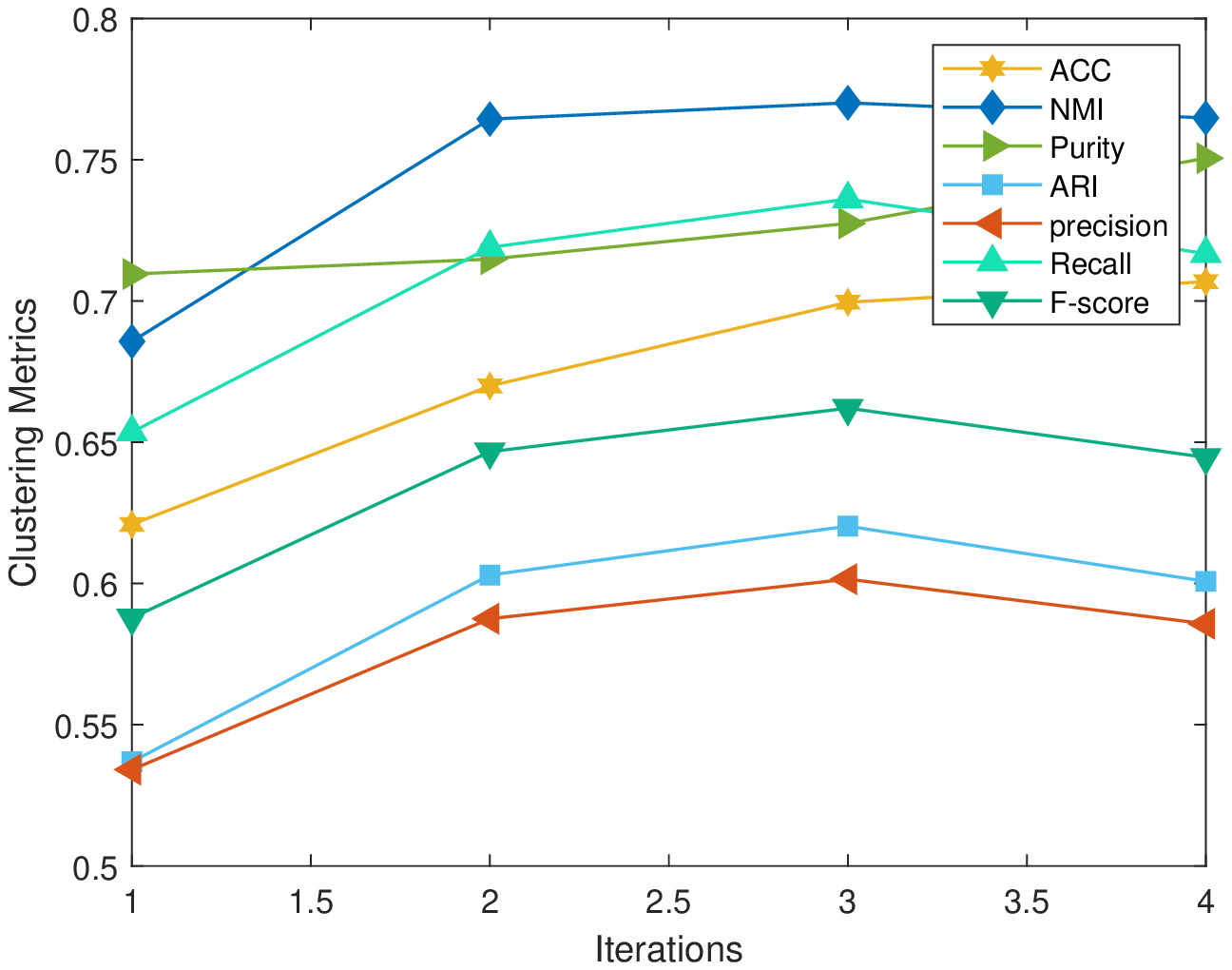}
}
\subfigure[SESC+Alg.1]{
\centering
\includegraphics[width=3.7cm,height=3cm]{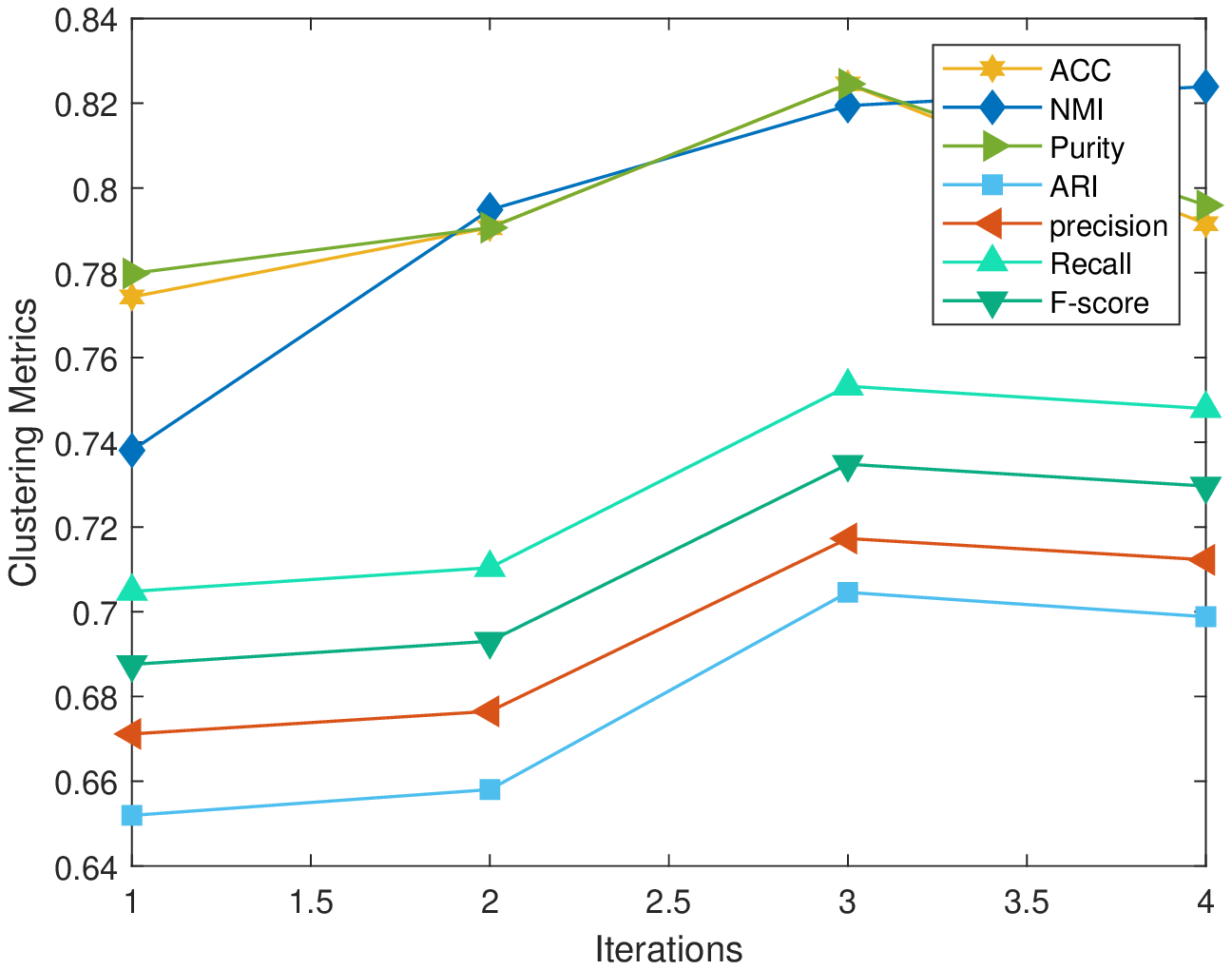}
}
\subfigure[USPEC+Alg.1]{
\centering
\includegraphics[width=3.7cm,height=3cm]{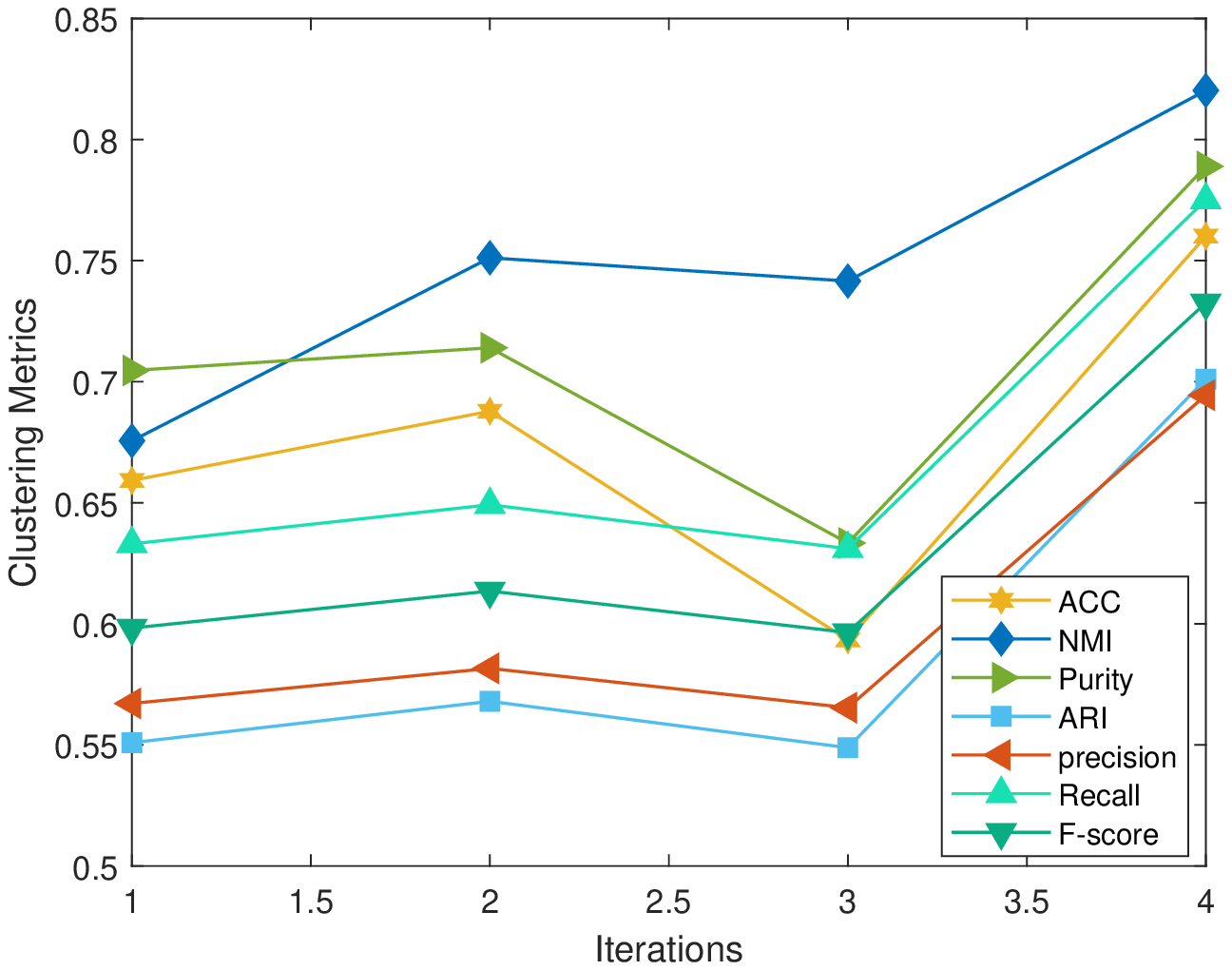}
}
\subfigure[USENC+Alg.1]{
\centering
\includegraphics[width=3.7cm,height=3cm]{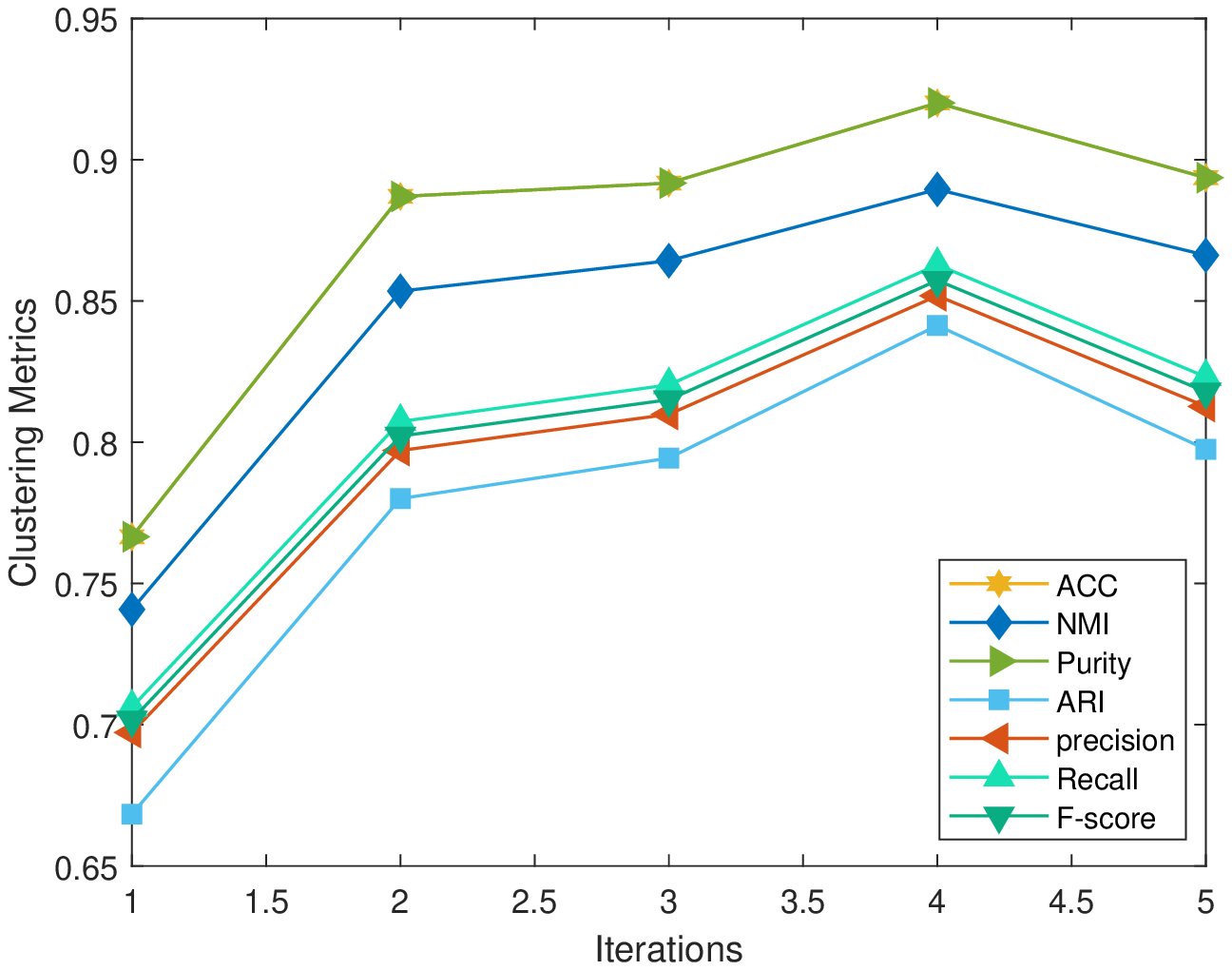}
}
\subfigure[DNSC+Alg.1]{
\centering
\includegraphics[width=3.7cm,height=3cm]{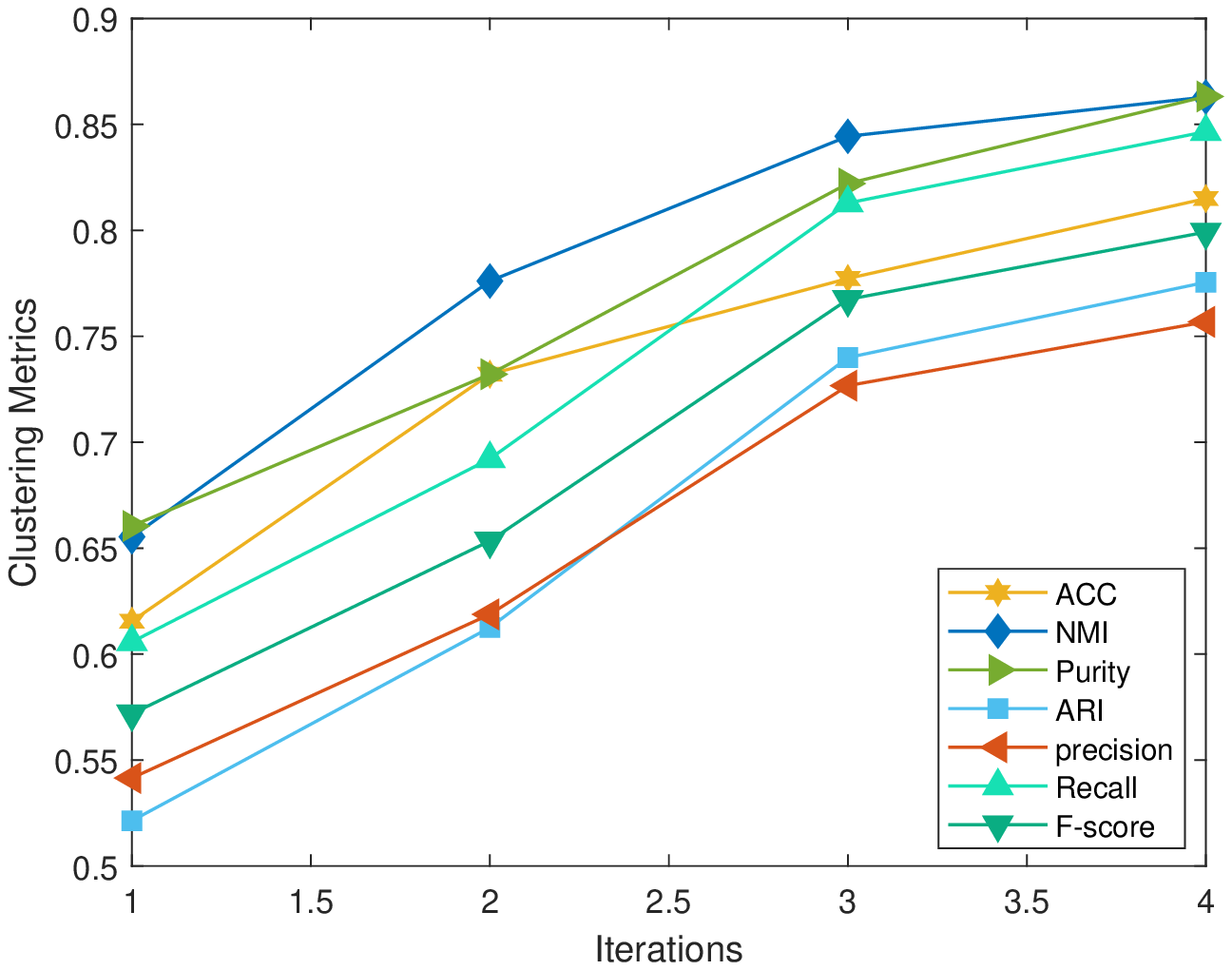}
}
\subfigure[FGNSC+Alg.1]{
\centering
\includegraphics[width=3.7cm,height=3cm]{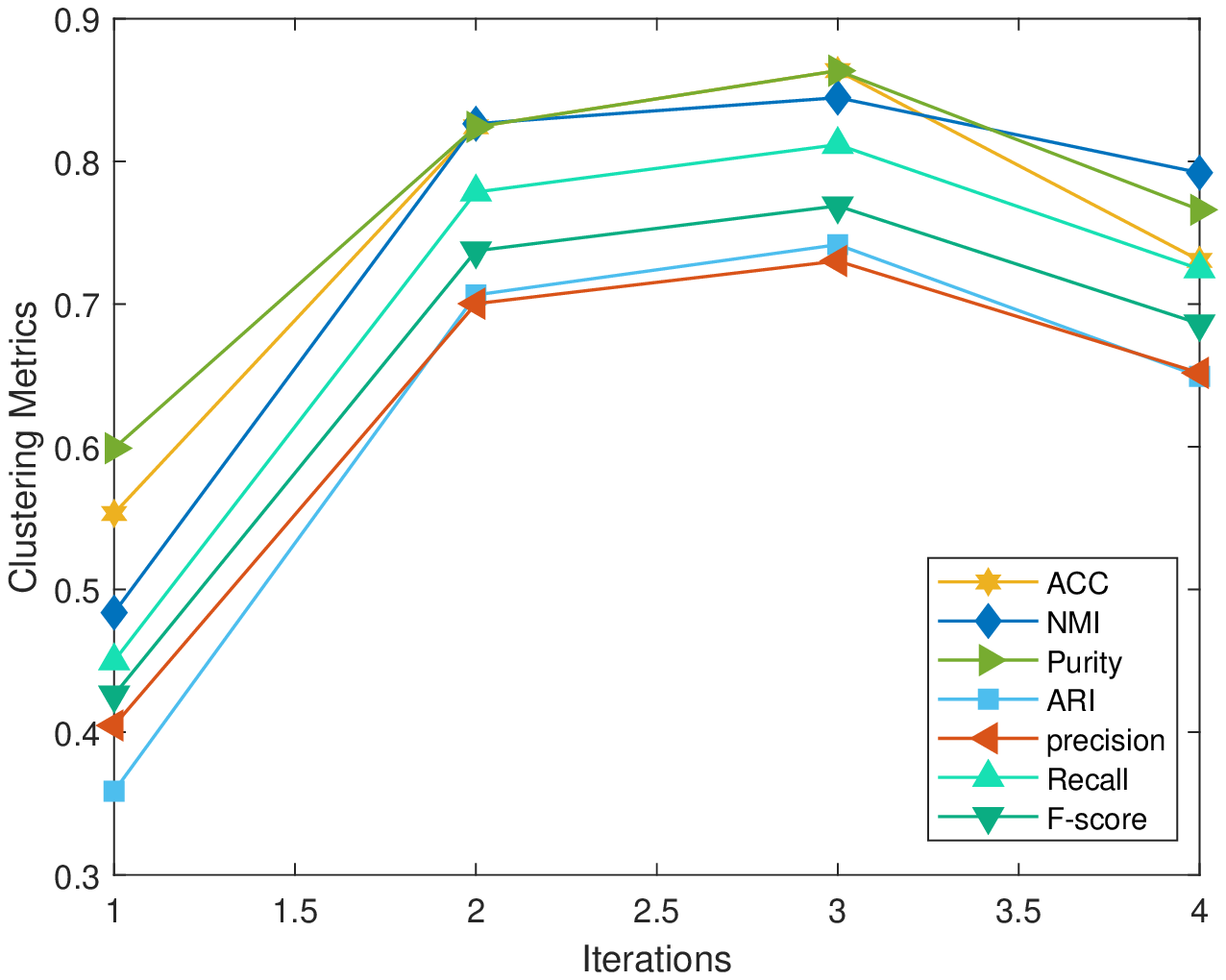}
}
\subfigure[MKKM+Alg.1]{
\centering
\includegraphics[width=3.7cm,height=3cm]{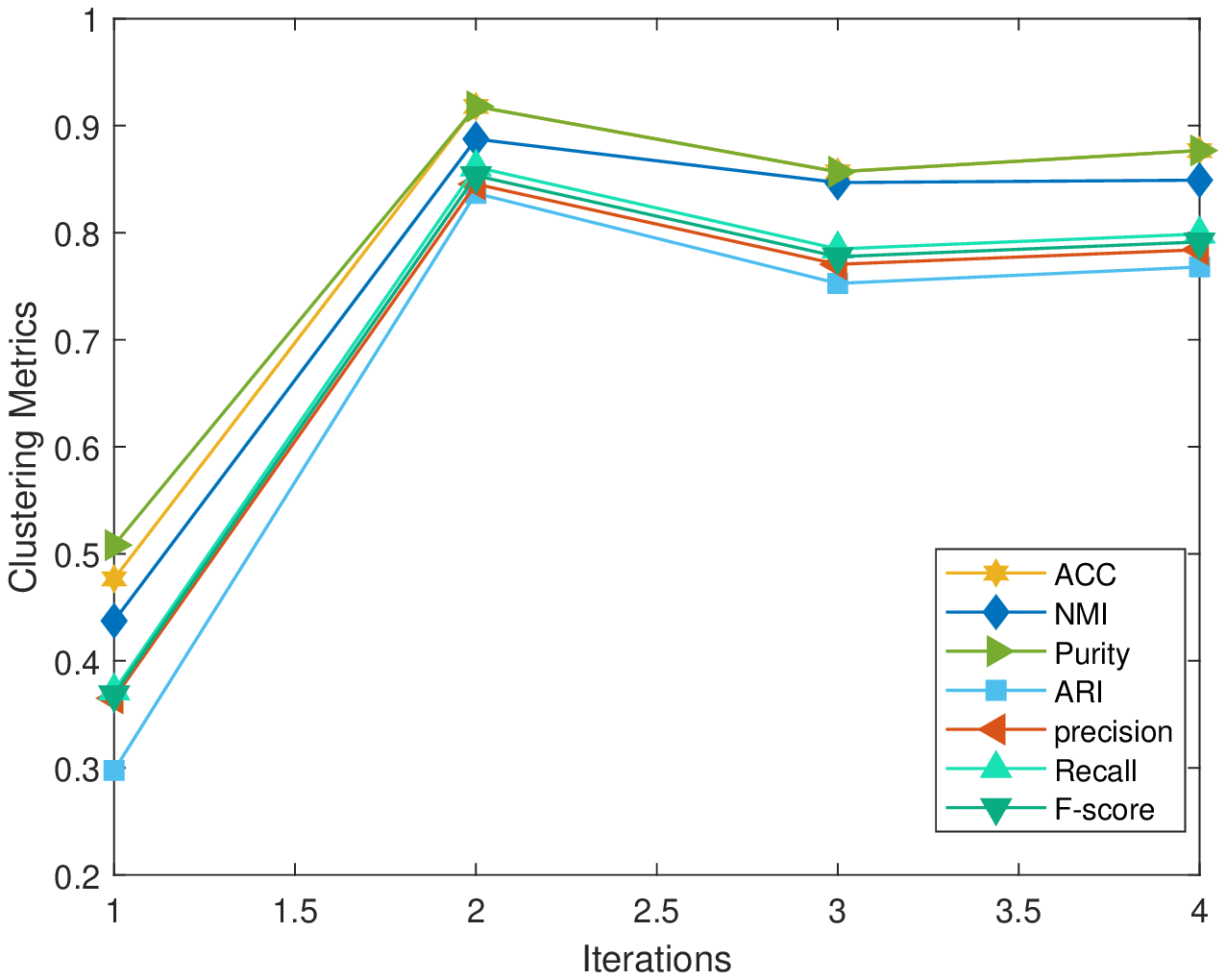}
}
\caption{Example 5.2: Values of the seven clustering evaluation criteria form Algorithm \ref{Alg1} during cycles, the {\tt USPS} database. Here the values of the first cycle are from the ``baseline" algorithms.}
\label{resK}
\end{figure*}

\begin{figure*}[htbp]
\centering
\subfigure[$K$-means+Alg.2]{
\centering
\includegraphics[width=3.7cm,height=3cm]{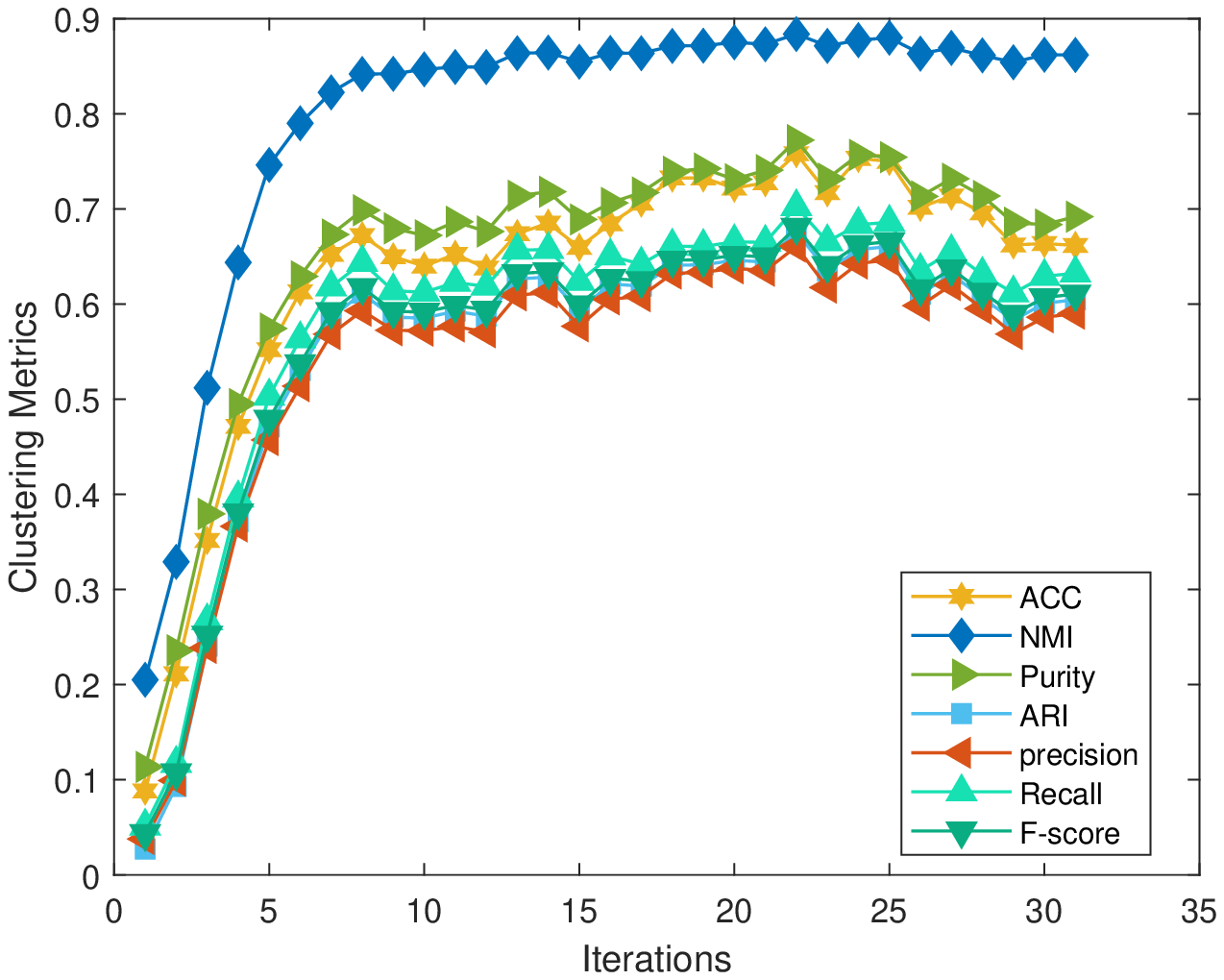}
}
\subfigure[Rcut+Alg.2]{
\centering
\includegraphics[width=3.7cm,height=3cm]{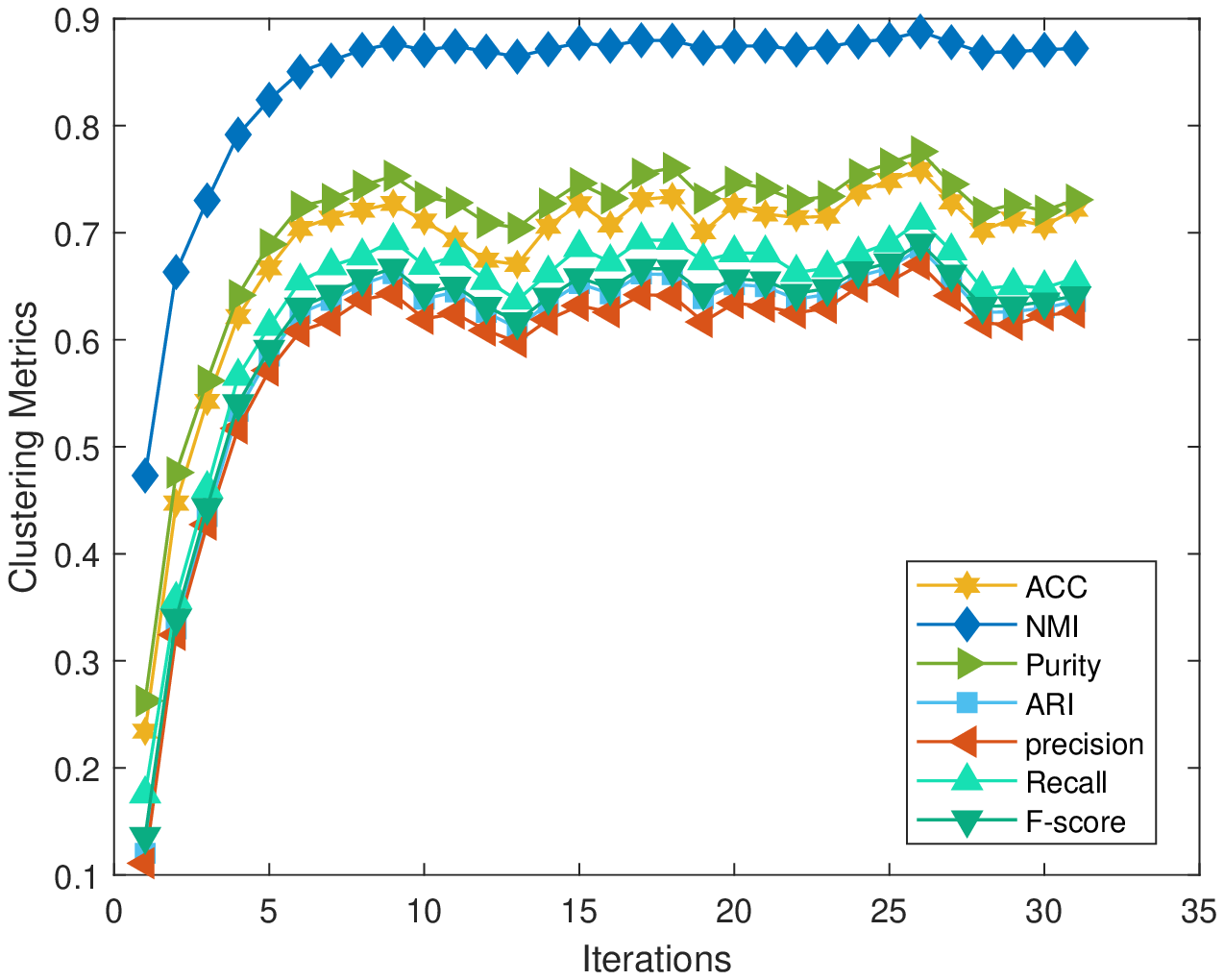}
}
\subfigure[Ncut+Alg.2]{
\centering
\includegraphics[width=3.7cm,height=3cm]{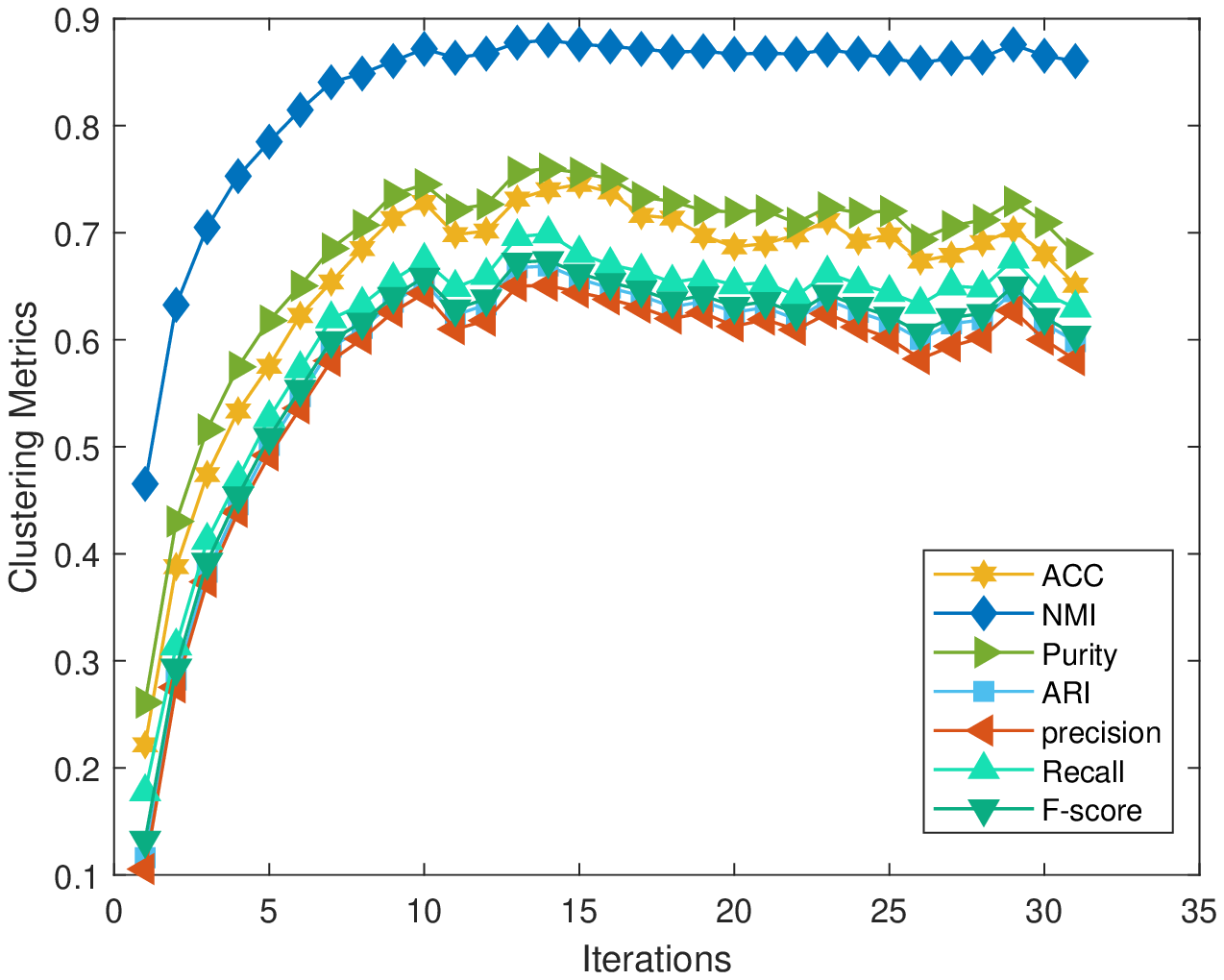}
}
\subfigure[SESC+Alg.2]{
\centering
\includegraphics[width=3.7cm,height=3cm]{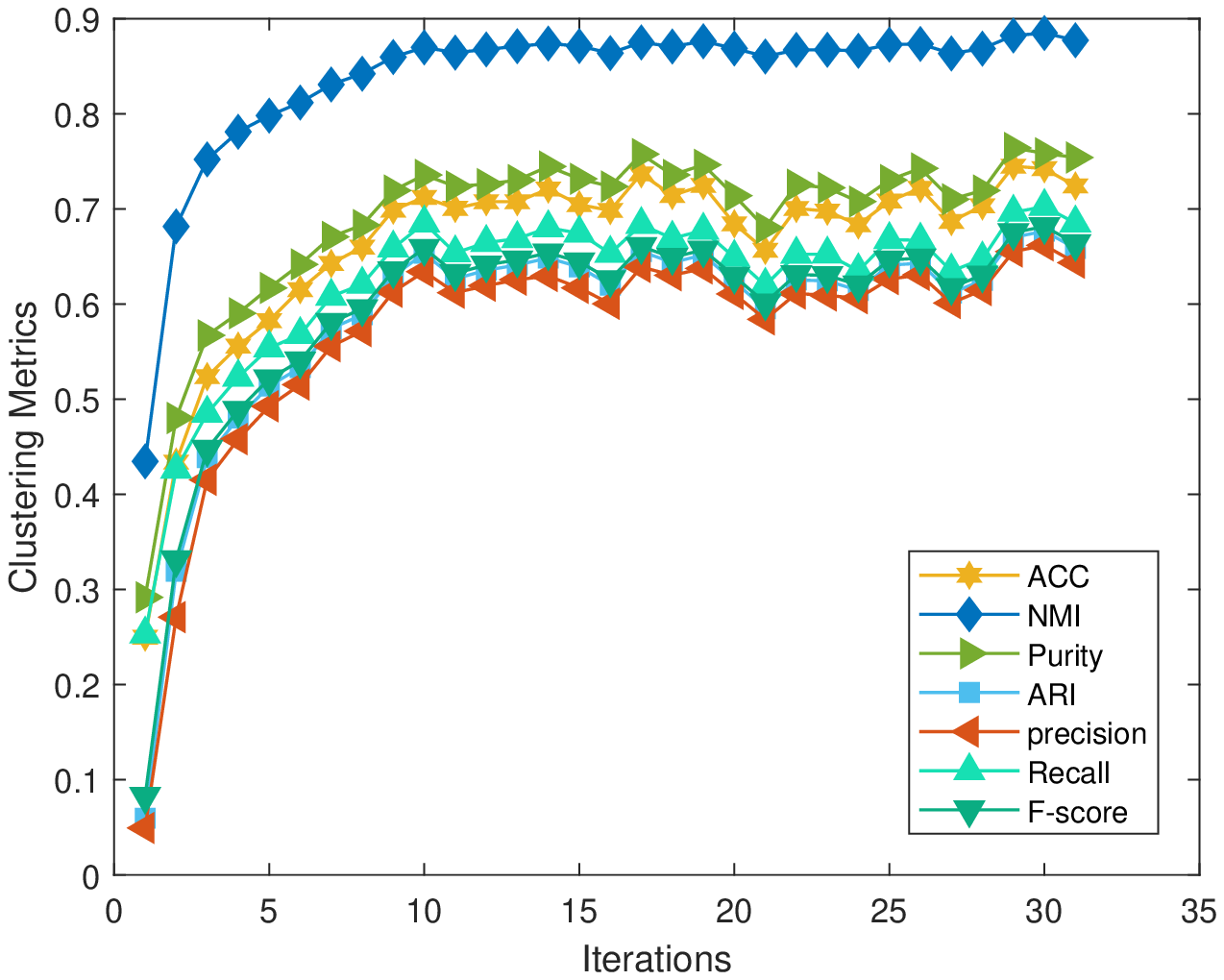}
}
\subfigure[USPEC+Alg.2]{
\centering
\includegraphics[width=3.7cm,height=3cm]{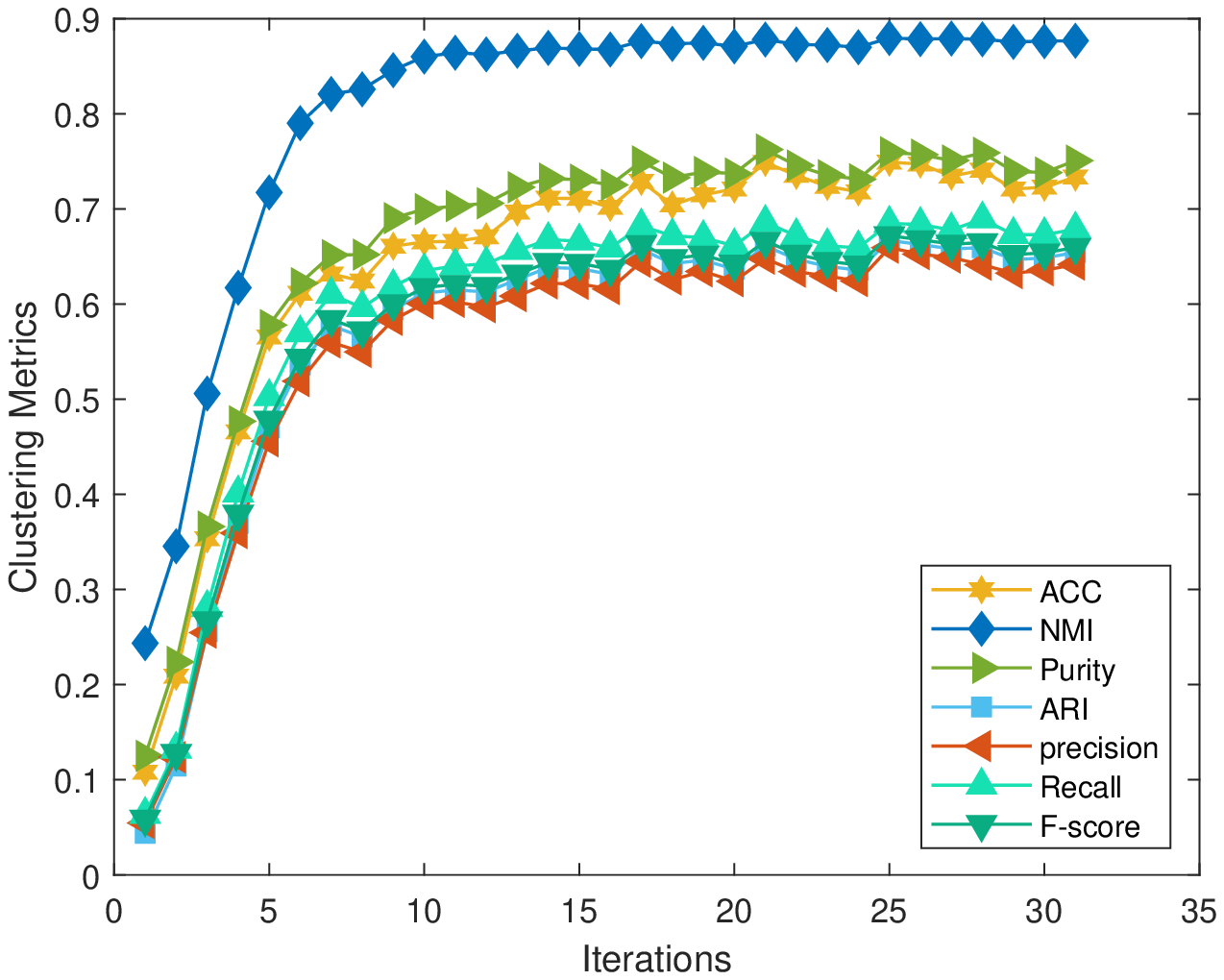}
}
\subfigure[USENC+Alg.2]{
\centering
\includegraphics[width=3.7cm,height=3cm]{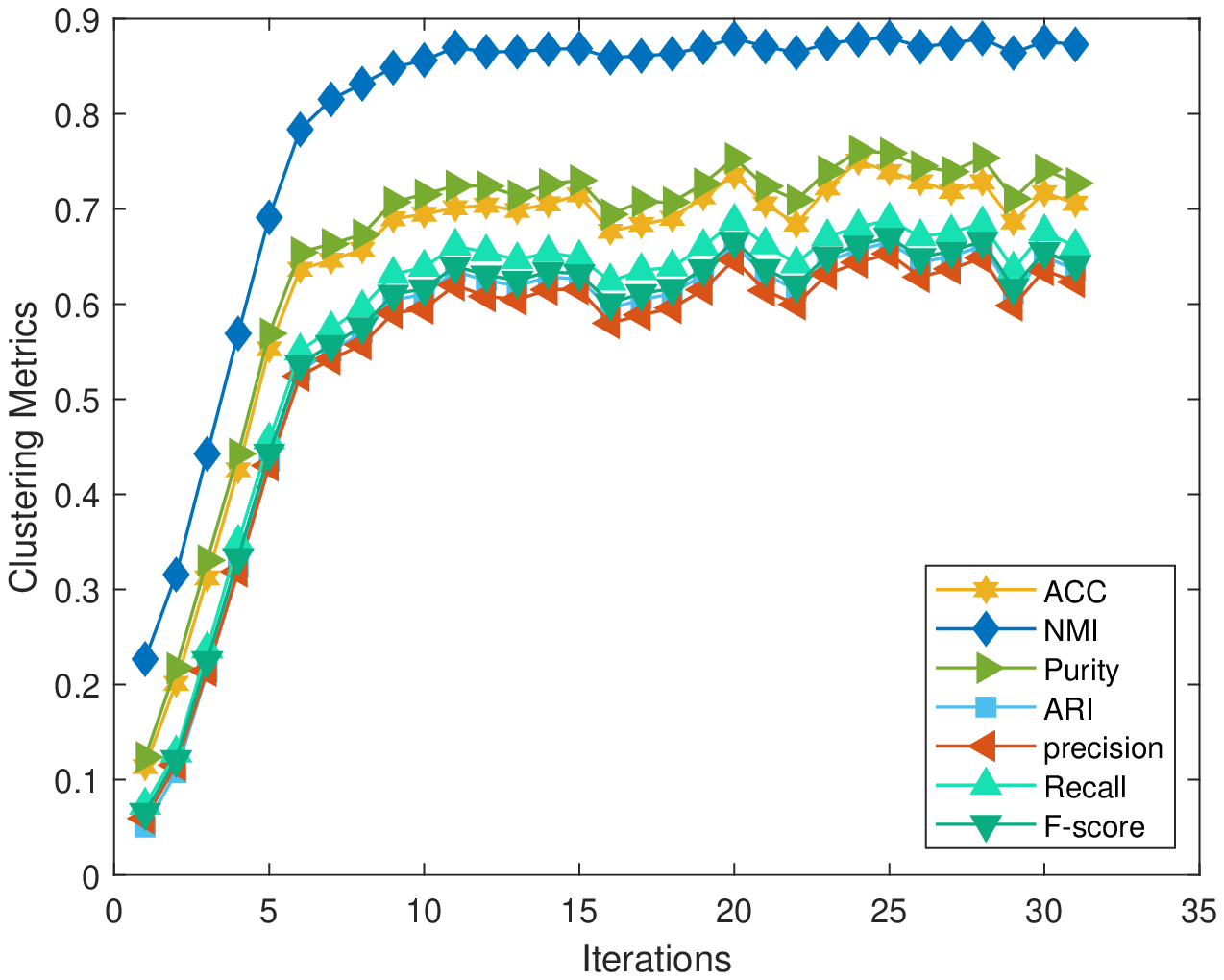}
}
\subfigure[DNSC+Alg.2]{
\centering
\includegraphics[width=3.7cm,height=3cm]{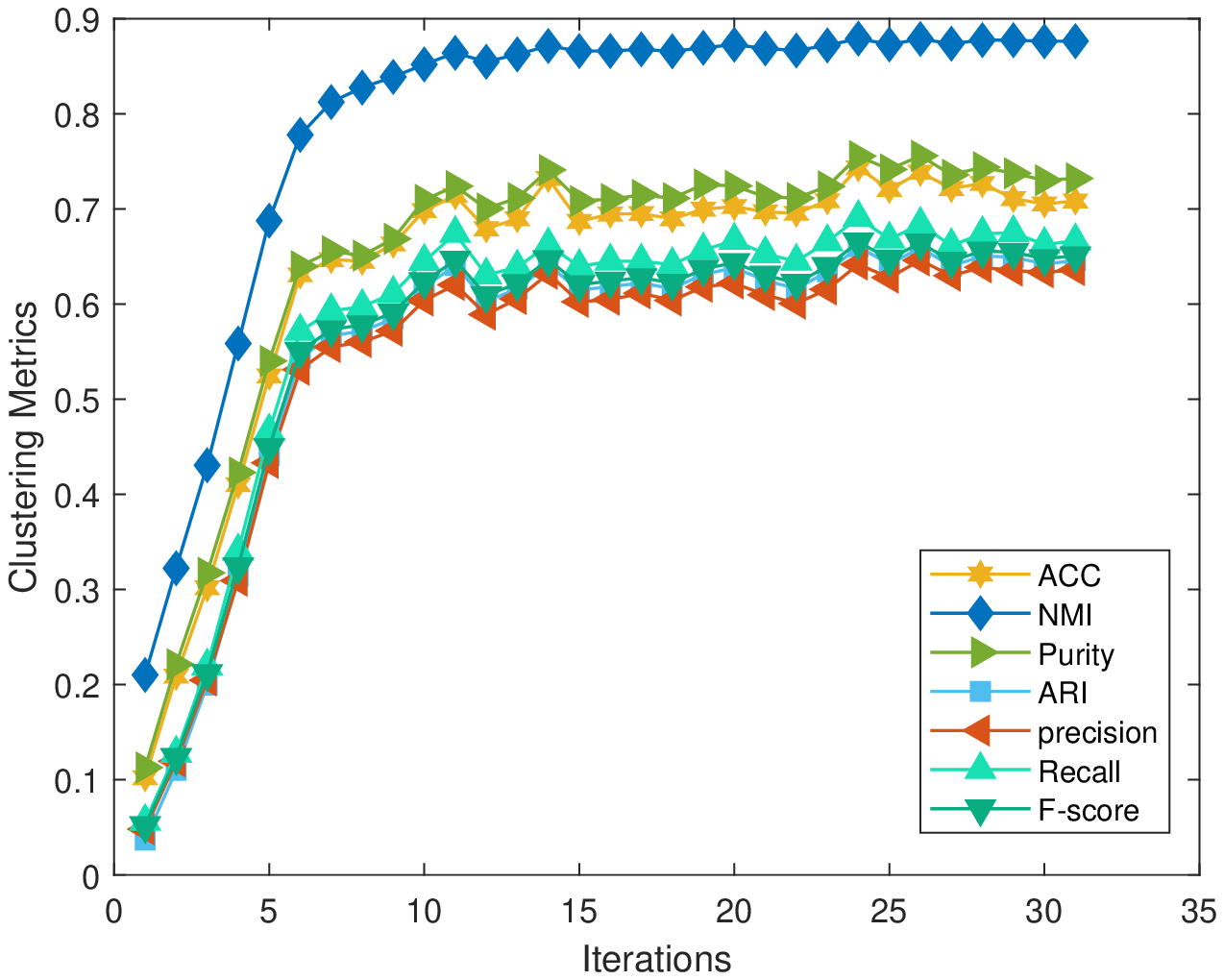}
}
\subfigure[FGNSC+Alg.2]{
\centering
\includegraphics[width=3.7cm,height=3cm]{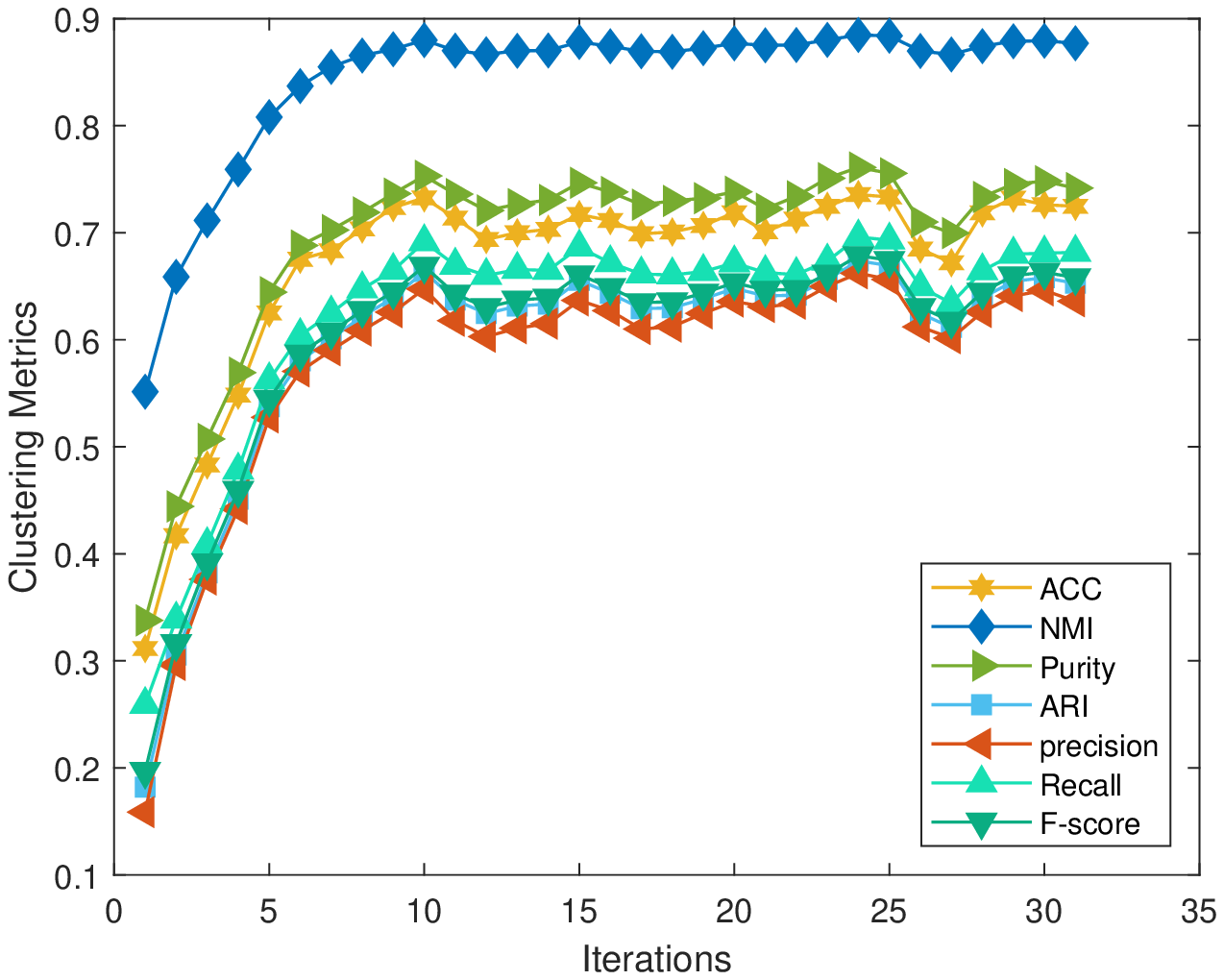}
}
\subfigure[MKKM+Alg.2]{
\centering
\includegraphics[width=3.7cm,height=3cm]{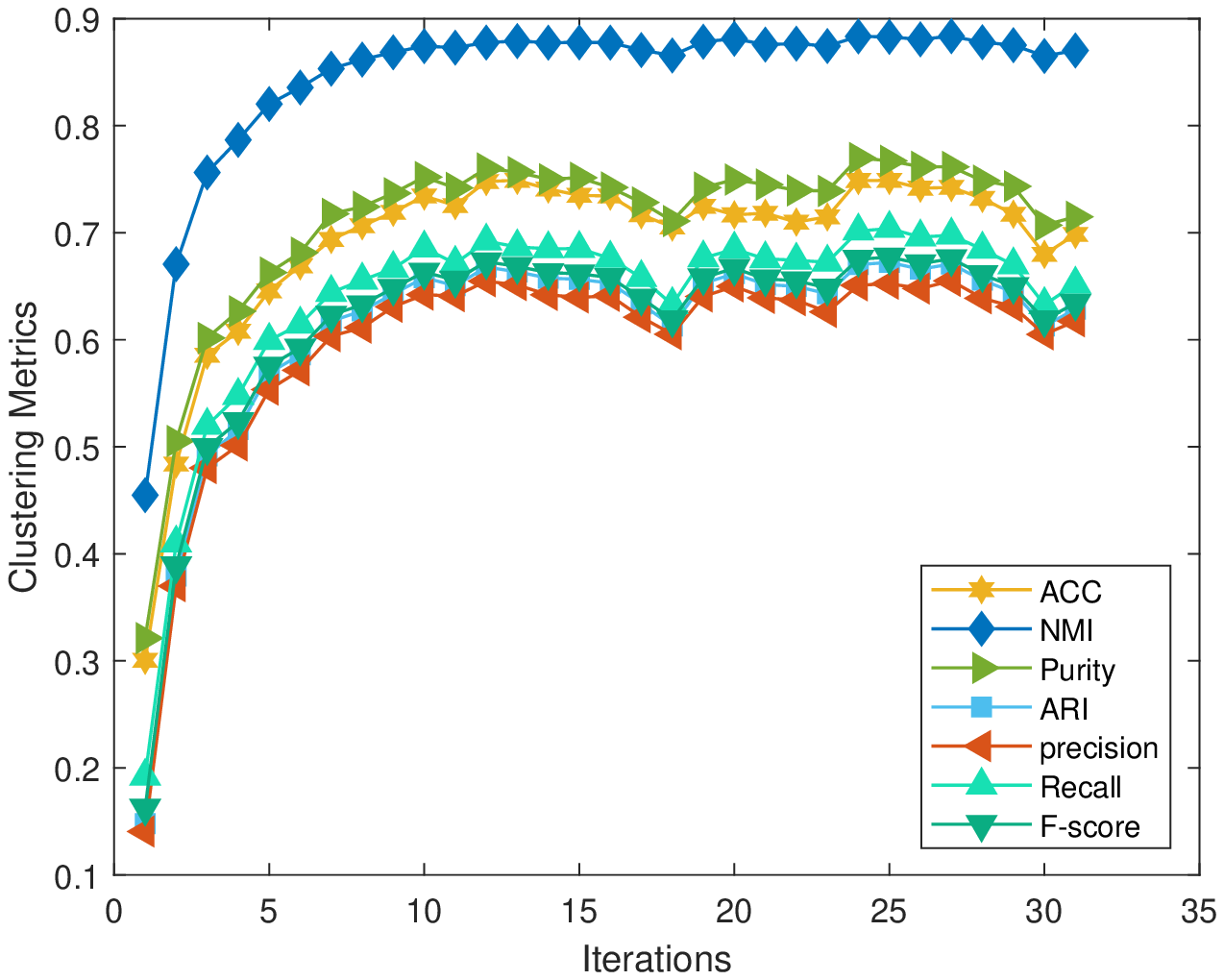}
}
\caption{Example 5.2: Values of the seven clustering evaluation criteria from Algorithm \ref{Alg3} during cycles, the {\tt CMU-PIE} database. Here the values of the first cycle are from the ``baseline" algorithms.}
\label{resR}
\end{figure*}

To show this more precisely, we plot in Fig. \ref{resK} and Fig. \ref{resR} values of the seven clustering evaluation criteria form Algorithm \ref{Alg1} and Algorithm \ref{Alg3} during cycles, respectively, where the values of the first cycle are from the ``baseline" algorithms. It is obvious to see that Algorithm \ref{Alg1} and Algorithm \ref{Alg3} improve the clustering performances of the ``baseline" algorithms gradually during cycles. As a comparison, Algorithm \ref{Alg1} is easier to meet the stopping conditions, while  Algorithm \ref{Alg3} is relatively more stable. Indeed, unlike some existing methods, as we do not fuse the similarity matrix and indicator matrix into a unified objective function, which may bring us some difficulties in the convergence or the algorithms may even diverge. However, the proposed algorithms can substantially improve the clustering performances of some popular algorithms.

\begin{figure*}[htbp]
\centering
\subfigure[USPS]{
\centering
\includegraphics[width=3.7cm,height=3cm]{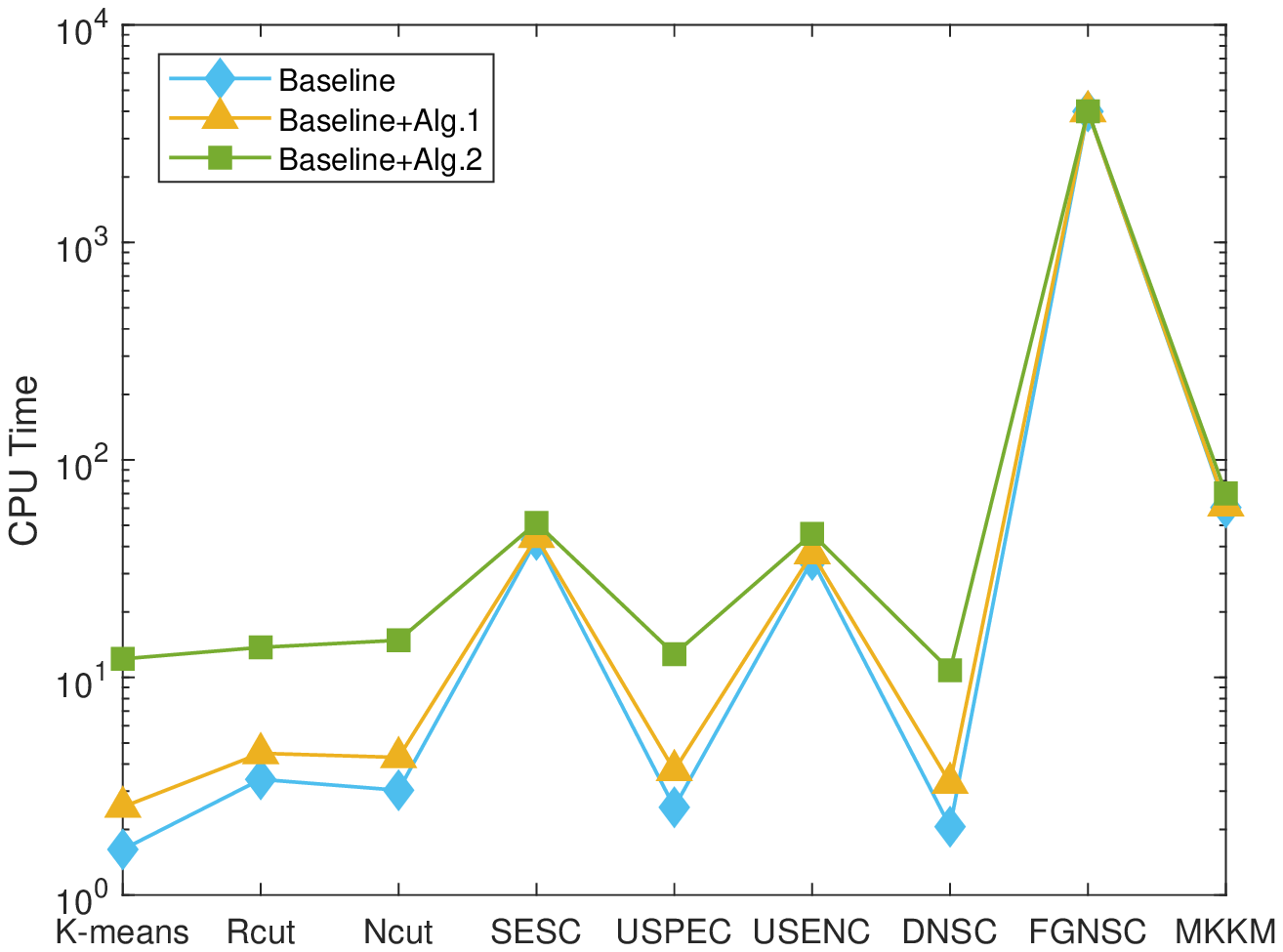}
}
\subfigure[CMU-PIE]{
\centering
\includegraphics[width=3.7cm,height=3cm]{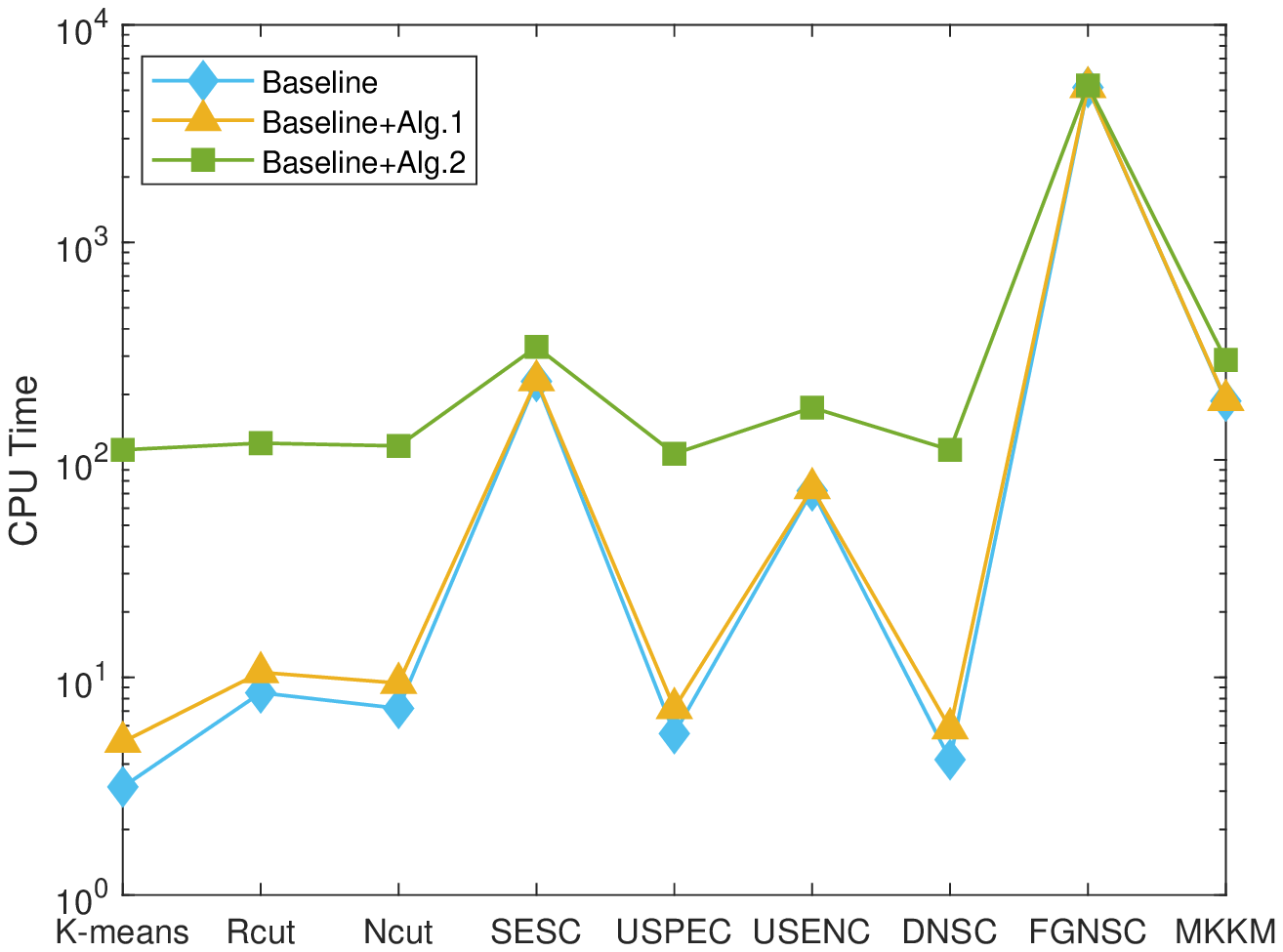}
}
\subfigure[20Newsgroups]{
\centering
\includegraphics[width=3.7cm,height=3cm]{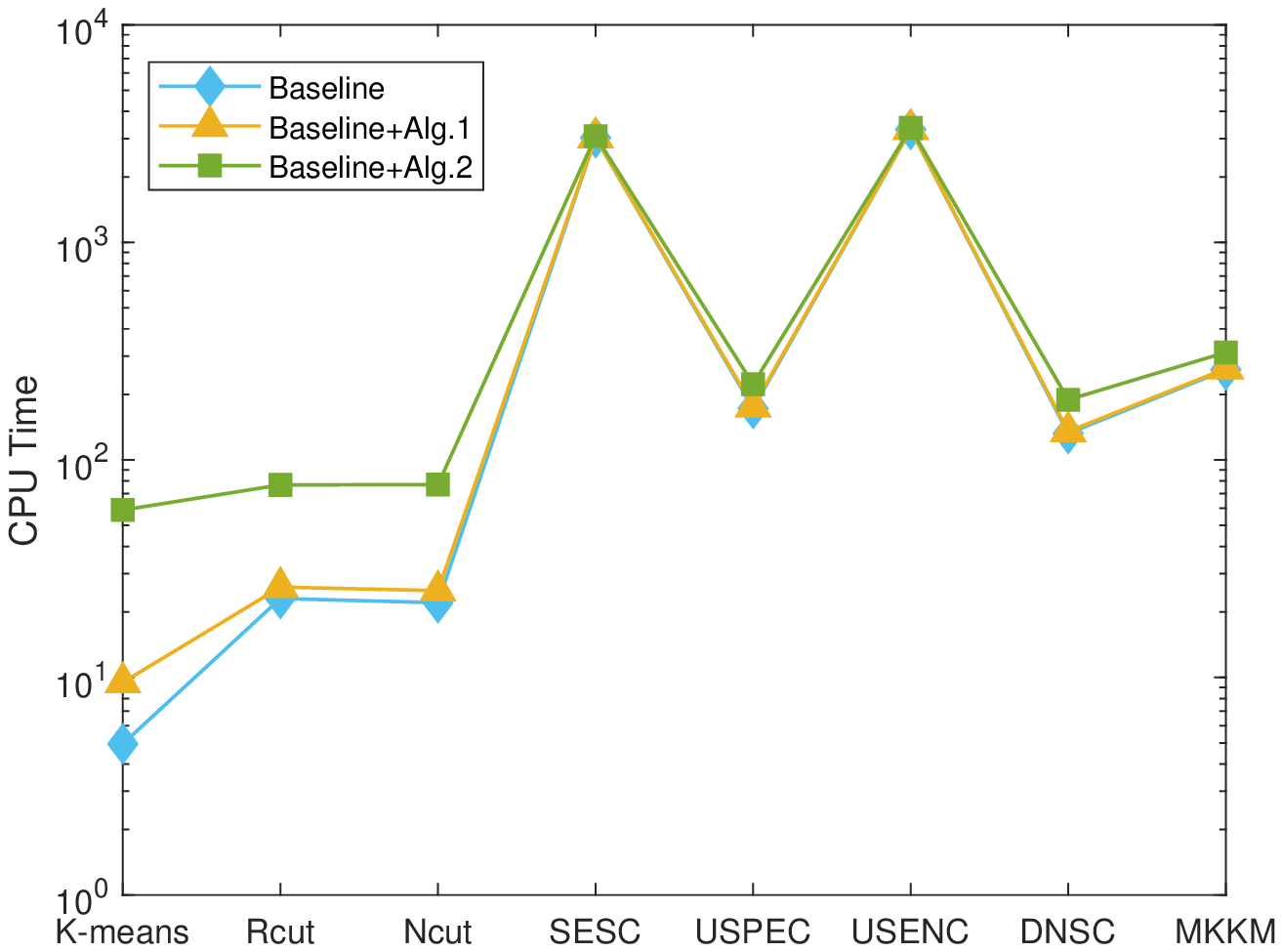}
}
\caption{Example 5.2: A comparison of the CPU time (in seconds) of the ``baseline" algorithms and the corresponding restarted algorithms on the three databases.}
\label{time}
\end{figure*}

Second, we see from Tables \ref{resUSPS}--\ref{resNG} that Algorithm \ref{Alg1} and Algorithm \ref{Alg3} improve the performances of the ``baseline" algorithms using very limited costs. Indeed, as
Baseline+Alg.1 and Baseline+Alg.2 make use of the results from the ``baseline" algorithms as the initial guesses, we need more running time in the proposed algorithms.
Notice that for the {\tt 20Newsgroups} database, the ``baseline" algorithm {\tt FGNSC} failed to work within 12 hours.
To show this more precisely, we plot in Fig. \ref{time} a comparison of the CPU time (in seconds) of the ``baseline" algorithms and the corresponding restarted algorithms on the three databases {\tt USPS}, {\tt CMU-PIE} and {\tt 20Newsgroups}. It is seen that the CPU timings of the ``baseline" algorithm, Baseline+Alg.1 and Baseline+Alg.2 are comparable in some cases, and Baseline+Alg.2 is slower than Baseline+Alg.1 since five variables need be updated in each cycle. Indeed, we find that the main workload of Baseline+Alg.2 is to update $M$ by using the generalized power iteration method \cite{17GPISM}.


\subsection{A comparison with the restarting and updating algorithms}\label{Sec5.3}

Some existing {\it updating} methods often suffer from heavy overhead for updating the similarity matrices for big data problems. This is prohibitive and even is infeasible in practice. To overcome this difficulty, we run the proposed {\it restarting} algorithms with initial guesses chosen randomly, denoted by RINV+Alg.1 and RINV+Alg.2. More precisely, we run Algorithm \ref{Alg1} and Algorithm \ref{Alg3} with the initial classification label $Y^{(0)}$ chosen randomly.
In this example, we try to illustrate the superiority of our {\it restarting} algorithms over some state-of-the-art {\it updating} algorithms in terms of clustering performances and running time.

We compare the two proposed algorithms with BDRSC \cite{19BDRSC} and  AwSCGLD \cite{22SCGLD}, UoMvSC \cite{22UMVSC} and MVCMK \cite{19MVCMK} on five large-scale databases {\tt USPS}, {\tt CMU-PIE}, {\tt 20Newsgroups}, {\tt MNIST} and {\tt YoutubeFace}. We mention that the last three approaches are multi-view algorithms that can be also applied to single-view problems. The numerical results are listed in Table \ref{resALL}, and the best results according to Average are highlighted in bold. It is seen from the table that RINV+Alg.1 and RINV+Alg.2 outperform the other four algorithms significantly, both in terms of clustering performances and CPU time, while RINV+Alg.1 is better than RINV+Alg.2. In other words, our algorithms performs very well even for initial guesses chosen randomly, and the proposed restarting strategy is promising for large-scale clustering problems.

\begin{figure*}[h!]
\centering
\subfigure[BDRSC]{
\begin{minipage}[c]{0.45\textwidth}
\centering
\includegraphics[width=4.5cm,height=3cm]{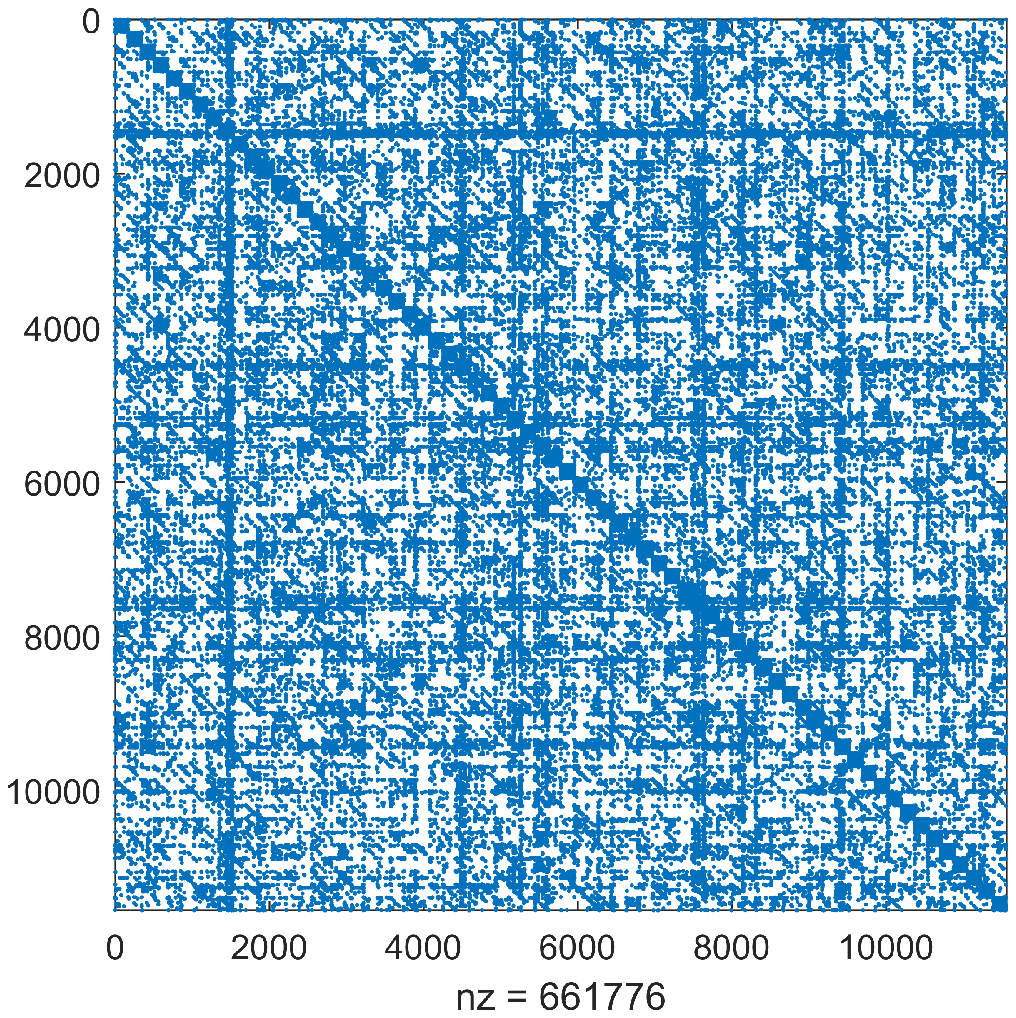}
\end{minipage}
}
\subfigure[UoMvSC]{
\begin{minipage}[c]{0.45\textwidth}
\centering
\includegraphics[width=4.5cm,height=3cm]{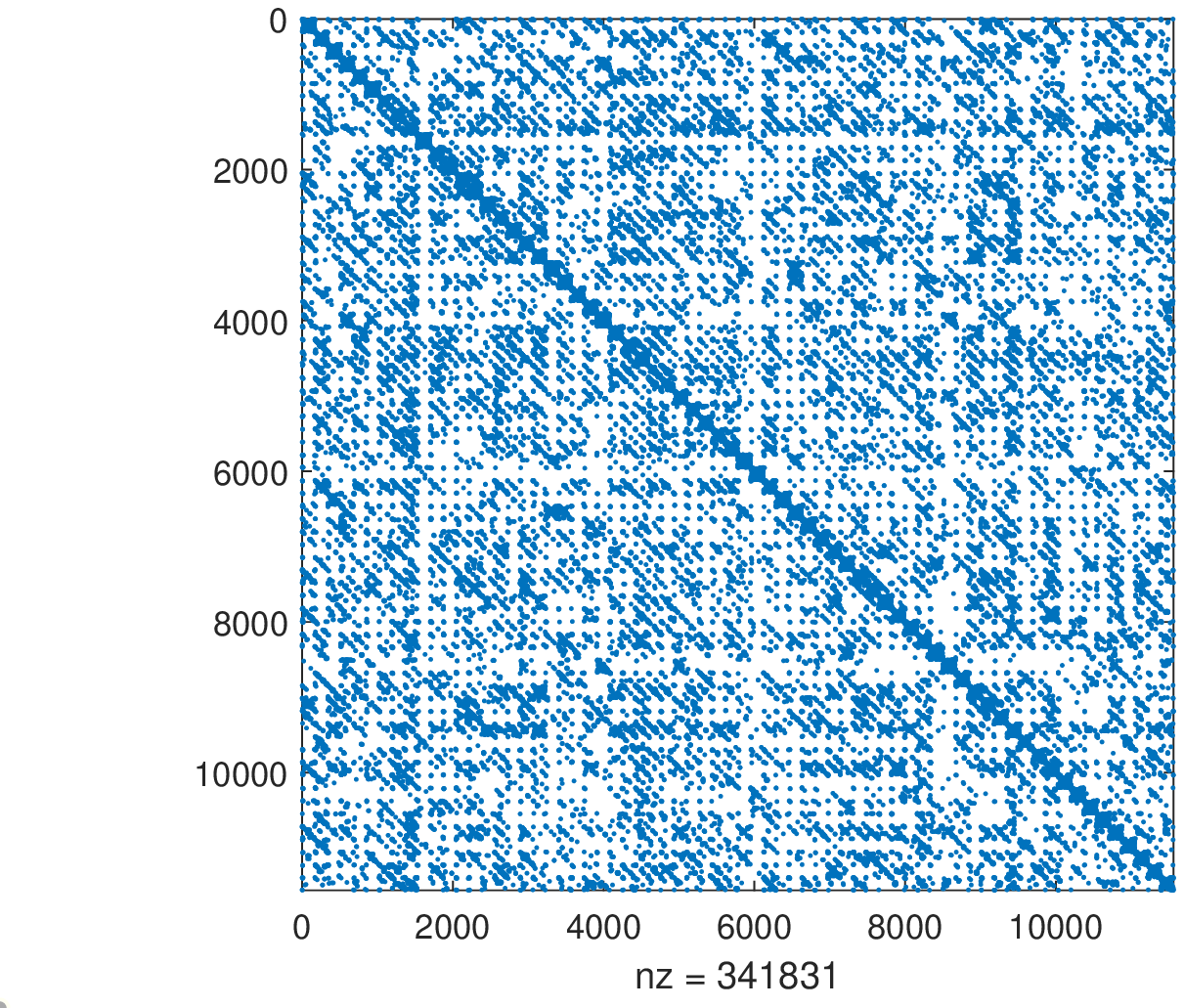}
\end{minipage}
}
\subfigure[RINV+Alg.1]{
\begin{minipage}[c]{0.45\textwidth}
\centering
\includegraphics[width=4.5cm,height=3cm]{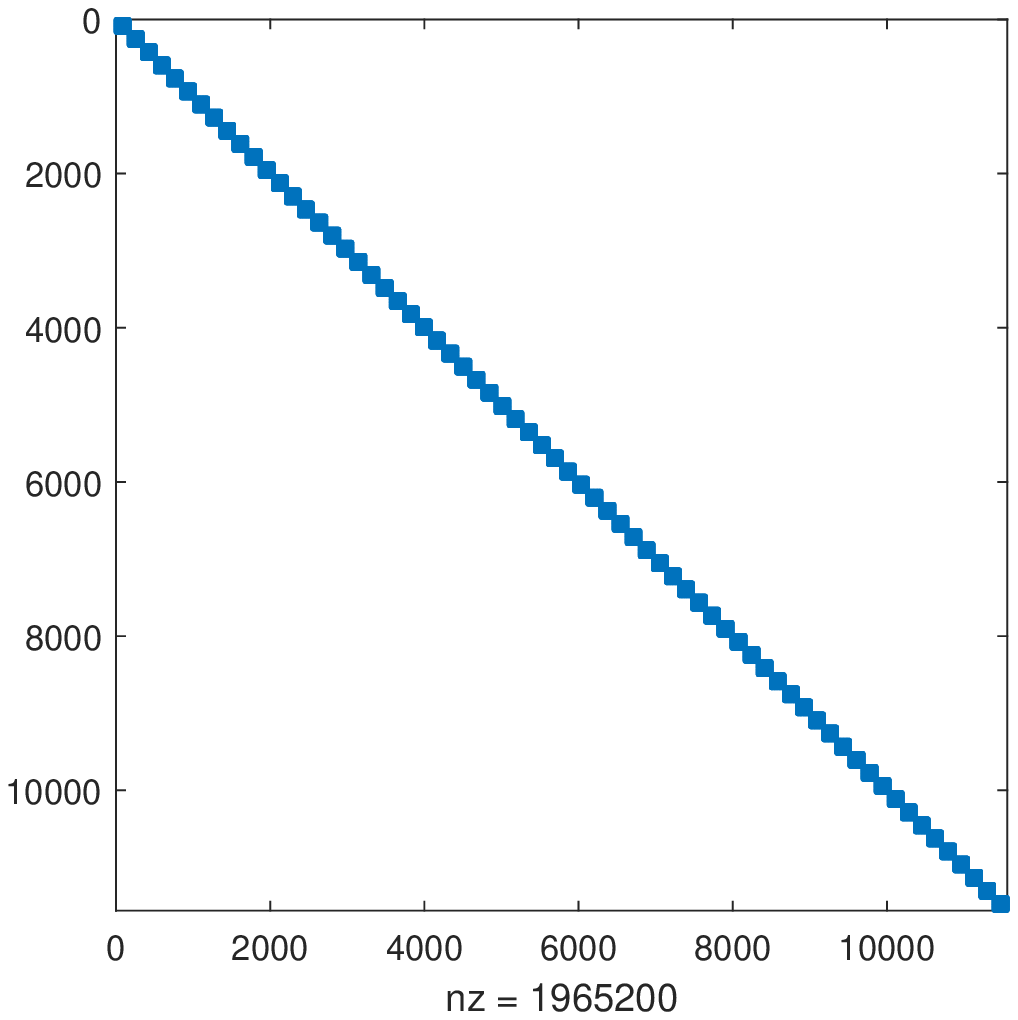}
\end{minipage}
}
\subfigure[RINV+Alg.2]{
\begin{minipage}[c]{0.45\textwidth}
\centering
\includegraphics[width=4.5cm,height=3cm]{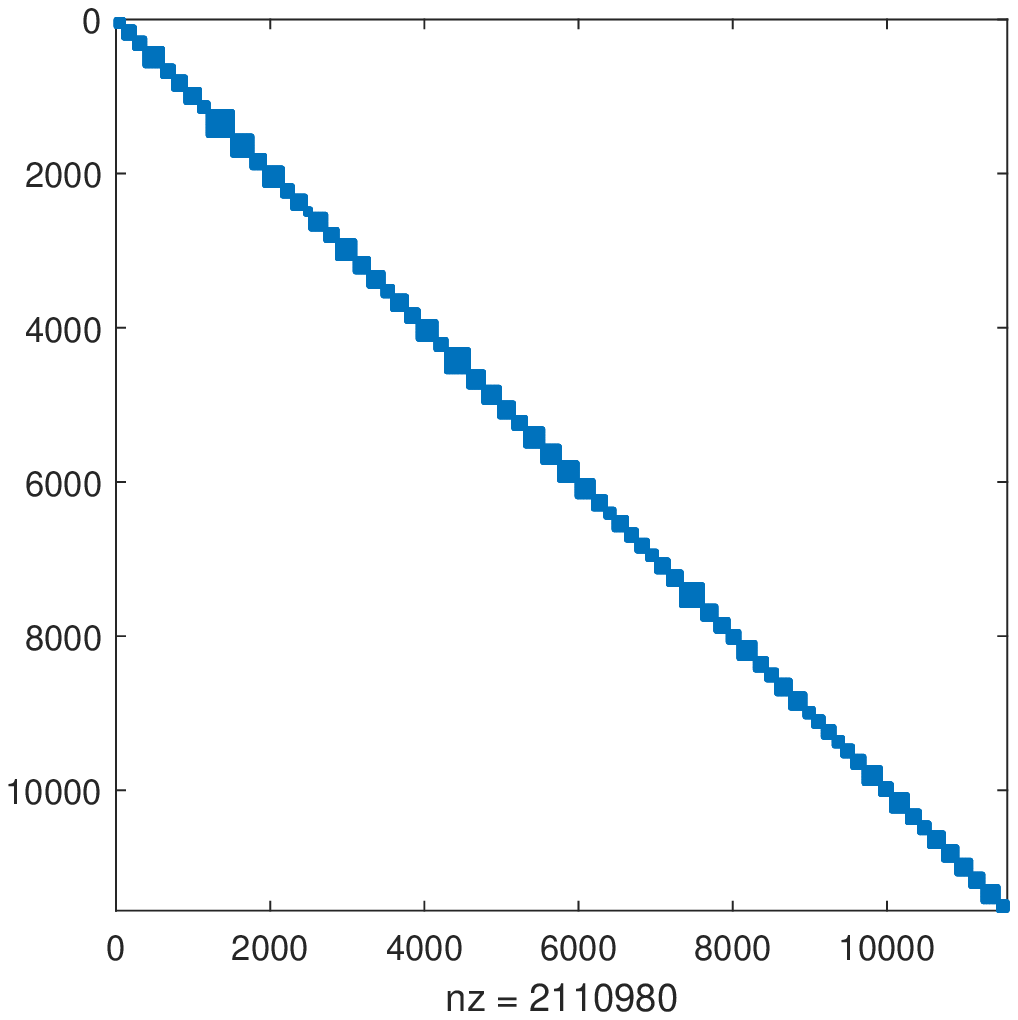}
\end{minipage}
}
\caption{Example 5.3: Sparse structures of the similarity matrices in the last cycle of the BDRSC, UoMvSC, RINV+Alg.1 and RINV+Alg.2, the {\tt CMU-PIE} database.}
\label{BDR}
\end{figure*}

\begin{table*}[h!]
\footnotesize{
\begin{center}
\caption{\it Example 5.3: Performances and CPU time (in seconds) on large-scale databases, where ``--" means the algorithm fail to work within 12 hours and ``O.M." stands for the algorithm suffers from the difficulty if ``out of memory".}\label{resALL}
\resizebox{\linewidth}{!}{
\begin{tabular}{c c c c c c c c c c c}
\toprule
database &Method &ACC &NMI &Purity &ARI &precision &Recall &F-score &Average &CPU\\
\midrule
\multirow{6}{*}{USPS}
&BDRSC  &0.402	&0.511	&0.494	&0.337	&0.343	&0.535	&0.418	&0.434	&17526.1\\
&AwSCGLD &0.613	&0.683	&0.663	&0.536	&0.526	&0.667	&0.588	&0.611	&1948.7\\
&UoMvSC &0.659	&0.696	&0.694	&0.541	&0.556	&0.628	&0.590	&0.623	&365.9\\
&MVCMK &-- &-- &-- &-- &-- &-- &-- &--&--\\
&RINV+Alg.1 &1.000	&1.000	&1.000	&1.000	&1.000	&1.000	&1.000	&\bf{1.000}	&1.0\\
&RINV+Alg.2 &0.771	&0.788	&0.778	&0.655	&0.670	&0.712	&0.690	&0.723 &10.8\\
\hline
\multirow{6}{*}{CMU-PIE}
&BDRSC &0.532	&0.721	&0.575	&0.407	&0.349	&0.520	&0.417	&0.503&16807.8\\
&AwSCGLD &-- &-- &-- &-- &-- &-- &-- &--&--\\
&UoMvSC &0.278	&0.530	&0.325	&0.157	&0.141	&0.221	&0.172	&0.260	&514.4\\
&MVCMK &-- &-- &-- &-- &-- &-- &-- &--&--\\
&RINV+Alg.1 &1.000	&1.000	&1.000	&1.000	&1.000	&1.000	&1.000	&\bf{1.000}	&3.1\\
&RINV+Alg.2 &0.721	&0.876	&0.739	&0.649	&0.635	&0.675	&0.654	&0.707&108.0\\
\hline
\multirow{6}{*}{20Newsgroups}
&BDRSC  &-- &-- &-- &-- &-- &-- &-- &--&--\\
&AwSCGLD &-- &-- &-- &-- &-- &-- &-- &--&--\\
&UoMvSC &0.087	&0.083	&0.098	&0.001	&0.051	&0.817	&0.096	&0.176	&1499.0\\
&MVCMK &-- &-- &-- &-- &-- &-- &-- &--&--\\
&RINV+Alg.1 &0.754	&0.847	&0.761	&0.677	&0.697	&0.690	&0.693	&\bf{0.731}	&40.4\\
&RINV+Alg.2 &0.705	&0.818	&0.728	&0.621	&0.624	&0.659	&0.641	&0.685	&56.2\\
\hline
\multirow{6}{*}{MNIST}
&BDRSC  &-- &-- &-- &-- &-- &-- &-- &--&--\\
&AwSCGLD &-- &-- &-- &-- &-- &-- &-- &--&--\\
&UoMvSC &O.M.&O.M.&O.M.&O.M.&O.M.&O.M.&O.M.	&O.M. &O.M.\\
&MVCMK &-- &-- &-- &-- &-- &-- &-- &--&--\\
&RINV+Alg.1 &0.947	&0.923	&0.947	&0.894	&0.906	&0.903	&0.904	&\bf{0.918}	&44.3\\
&RINV+Alg.2 &0.832	&0.833	&0.837	&0.737	&0.750	&0.778	&0.764	&0.790	&303.3\\
\hline
\multirow{6}{*}{YoutubeFace}
&BDRSC  &-- &-- &-- &-- &-- &-- &-- &--&--\\
&AwSCGLD &-- &-- &-- &-- &-- &-- &-- &--&--\\
&UoMvSC &O.M.&O.M.&O.M.&O.M.&O.M.&O.M.&O.M.	&O.M. &O.M.\\
&MVCMK &-- &-- &-- &-- &-- &-- &-- &--&--\\
&RINV+Alg.1 &0.720	&0.864	&0.788	&0.662	&0.710	&0.636	&0.671	&\bf{0.722}&187.5\\
&RINV+Alg.2 &0.675	&0.848	&0.736	&0.611	&0.639	&0.606	&0.622	&0.677 &1187.7\\
\bottomrule
\end{tabular}}
\end{center}}
\end{table*}
To demonstrate the effectiveness of block diagonal representation in {\it restarting} algorithms, for the {\tt CMU-PIE} database, we plot in Fig. \ref{BDR} the sparse structures of the similarity matrices
in the last cycle of the BDRSC \cite{19BDRSC}, UoMvSC \cite{22UMVSC}, RINV+Alg.1 and RINV+Alg.2. Here BDRSC and UoMvSC pursue a soft block diagonal regular and try to make the similarity matrix
to be block diagonal \cite{19BDRSC,22UMVSC}. Notice that there is no block diagonal structure in the other two algorithms. We observe that for this problem, there is no block diagonal structure in the similarity matrices arising from BDRSC and UoMvSC.
Indeed, a block diagonal similarity matrix will bring us a block diagonal Laplacian matrix, and we only need to perform eigendecomposition on block matrices with much smaller size, and it is suitable to parallel computing.

\section{Conclusion}\label{sec6}

We propose a unified {\it restarting} framework with self-guiding for large-scale spectral clustering. The idea stems from keeping the last clustering information to improve the clustering performance gradually during cycles. Moreover, to further reduce the workload for large-scale problems, we introduce a block diagonal representation to approximate the similarity matrix. Theoretical results are established to show the rationality of using low-rank approximations in the proposed methods. Comprehensive numerical experiments verified the efficiency of the new algorithms and the effectiveness of the proposed strategy. Future works include considering convergence properties of the proposed restarting strategy in some sense, and applications of the proposed framework to multi-view clustering. These are interesting topics deserve further investigation.

\section*{Acknowledgments}
This work was supported by the Fundamental Research Funds for the Central Universities (No.2022XSCX10).
The authors would like to thank Dr. Feng Bo for helpful discussions.

\end{document}